\documentclass[a4paper]{article}
\pdfoutput=1

\usepackage{authblk}
\usepackage[margin=0.8in,a4paper]{geometry}
\usepackage[round,comma]{natbib}
\usepackage{eurosym}
\usepackage[utf8]{inputenc} 
\usepackage[T1]{fontenc}    

\usepackage{graphicx}
\usepackage{amssymb}
\usepackage{amsmath}
\usepackage{eurosym}
\usepackage{bbm}
\usepackage{float}
\usepackage{subfigure}
\usepackage{url}  

\newcommand{\RR}{\mathbb{R}}

\newcommand{\fig}[1]{Figure~\protect\ref{#1}}

\newcommand{\reach}{\textit{reach}}
\newcommand{\grain}{\textit{grain}}
\newcommand{\coherence}{\textit{coherence}}

\DeclareMathOperator{\sgn}{sgn}

\title{Estimation of perceptual scales using ordinal embedding}

\author[1]{Siavash Haghiri \thanks{Corresponding author, email: siyavash.haghiri@gmail.com}}
\author[2]{Felix Wichmann} 
\author[1]{Ulrike von Luxburg}
\date{May 2019}

\affil[1]{Department of Computer Science, University of T{\"u}bingen, Germany}
\affil[2]{Neural Information Processing Group, University of T{\"u}bingen, Germany}

\begin{document}

\maketitle

\abstract{
In this paper we address the problem of measuring and analysing sensation, the subjective magnitude of one's experience. We do this in the context of the method of triads: the sensation of the stimulus is evaluated via relative judgments of the form: ``Is stimulus $S_i$ more similar to stimulus $S_j$ or to stimulus $S_k$?''. We propose to use ordinal embedding methods from machine learning to estimate the scaling function from the relative judgments. We review two relevant and well-known methods in psychophysics which are partially applicable in our setting: non-metric multi-dimensional scaling (NMDS) and the method of maximum likelihood difference scaling (MLDS). We perform an extensive set of simulations, considering various scaling functions, to demonstrate the performance of the ordinal embedding methods. We show that in contrast to existing approaches our ordinal embedding approach allows, first, to obtain reasonable scaling function from comparatively few relative judgments, second, the estimation of non-monotonous scaling functions, and, third,  multi-dimensional perceptual scales. In addition to the simulations, we analyse data from two real psychophysics experiments using ordinal embedding methods. Our results show that in the one-dimensional, monotonically increasing perceptual scale our ordinal embedding approach works as well as MLDS, while in higher dimensions, only our ordinal embedding methods can produce a desirable scaling function. To make our methods widely accessible, we provide an R-implementation and general rules of thumb on how to use ordinal embedding in the context of psychophysics. 
}
\maketitle

\section{Introduction}

The quantitative study of human behavior dates back to at least 1860 when the experimental physicist Gustav Theodor Fechner published \emph{Die Elemente der Psychophysik} \cite{Fechner_1860}. Since Fechner's seminal work the ``measurement of sensation magnitude''---nowadays typically referred to as ``psychophysical scaling''---has been one of the central aims of psychophysics \citep{Gescheider1988}\footnote{Other central aims are to measure detection and discrimination thresholds, or just-noticeable-differences (JNDs), reaction times (RT) and confidence ratings, see e.g. \citet{Wichmann_Jakel_2018}.}. Psychophysical scaling is formally defined as the problem of quantifying the magnitude of sensation induced by a physical stimulus~\citep{Marks_Gescheider_2002,krantz2007foundations}.

\begin{figure}
    \centering
    \includegraphics[width = .4\linewidth]{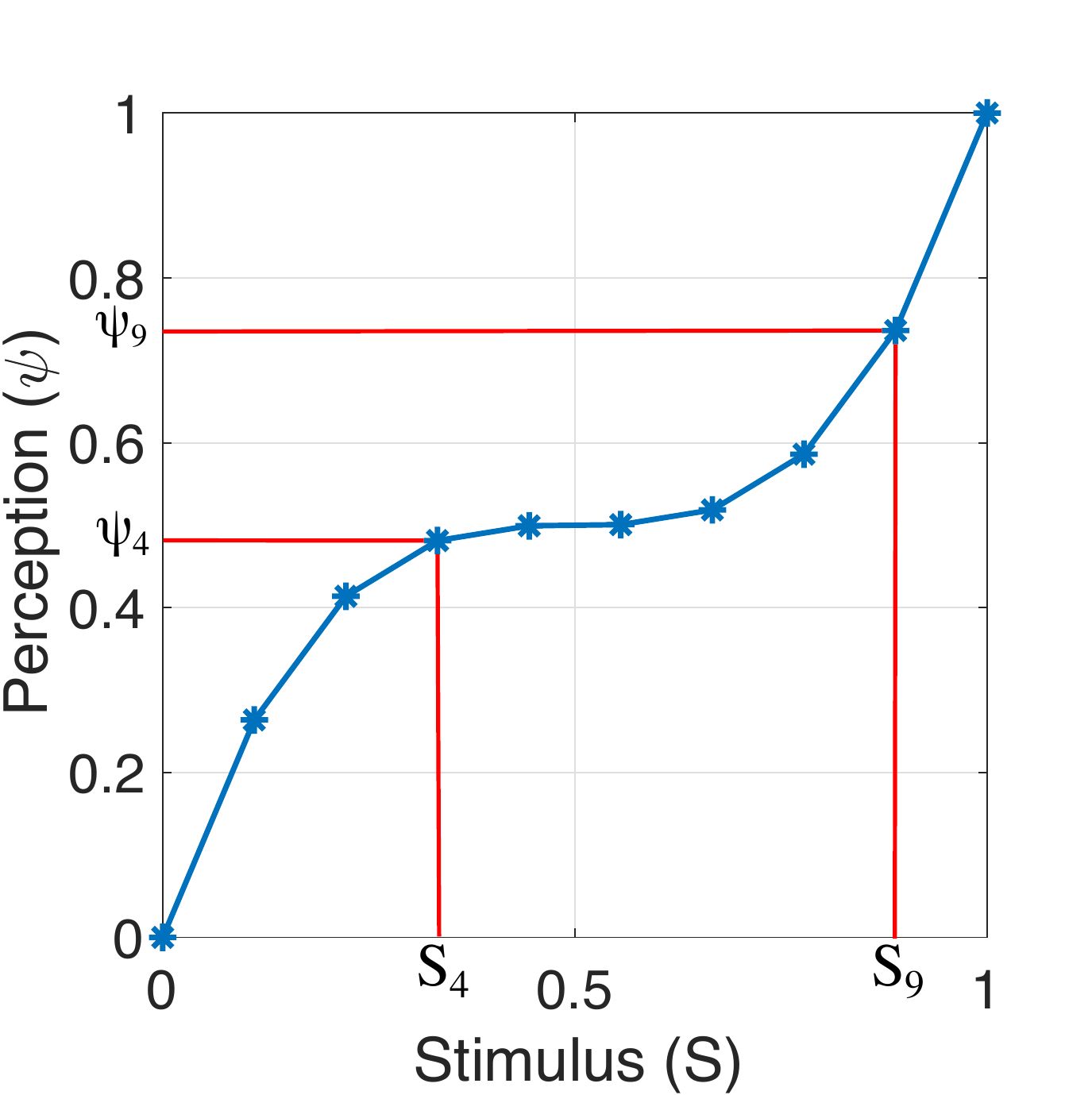}
    \caption{An example of a scaling function. The X-axis shows the physical stimulus values ($S$) with 10 discrete steps. Y-axis denotes the perceived value ($\psi$)}
    \label{fig:scalingExample}
\end{figure}

In the following we assume that there exists a physical quantity---the external stimulus---which we can objectively measure. The perception (or sensation, the subjective or internal experience) of the stimulus, however, is usually hard to measure and quantify. The (difference) scaling problem refers to experiments and methods designed to find the functional relation between the perceived (internal) magnitude and the (external) stimulus. An example of a scaling function is shown in Figure~\ref{fig:scalingExample}. In this figure, the physical stimulus $S$ and its perceived counterpart $\psi$ are denoted on the X and Y axes, respectively. Throughout the rest of the paper, we refer to this function as scaling function.

\subsection{Traditional scaling methods}

Early attempts to obtain the scaling function by Fechner were based on the concatenation of just-noticeable-difference (JND), the smallest amount of change in the stimulus level which is noticeable by a human observer. Fechner assumed that each JND in $S$ corresponds to one fixed-size unit of the perceptual scale $\psi$, and attempted to reconstruct the scaling function based on this assumption \citep{Fechner_1860,luce1958derivation}. Fechner thus tried to link {\em discriminability} and {\em subjective magnitude} in a simple way. However, the Fechnerian approach---albeit sometimes successful---has been vigorously criticised for both theoretical and empirical reasons and cannot serve as a generic method to obtain scaling functions, e.g.\ \citep{norris1898system,stevens1957psychophysical,Gescheider1988}. 
Thurstonian scaling is an alternative approach proposed to solve the scaling problem in the tradition of linking discriminability to subjective magnitude, incorporating an internally variable mapping from stimulus to sensation (internal ``noise'' in modern parlance)~\citep{thurstone1927law}. Thurstonian scaling is based on discrimination of stimuli pairs. The perceptual distance of two stimuli is determined by the probability that a human observer can discriminate them. However, like Fechner's JND-approach, Thurstonian scaling is criticised because discriminability is, at best, only {\em indirectly} and in yet to be understood ways related to sensory magnitude~\citep{krantz1972visual,stevens1961honor}.%

Another well-known approach to scaling, but this time not based on discriminability, is termed {\em direct} magnitude estimation~\citep{stevens1957psychophysical}. In this approach a human observer is asked to provide intensity values corresponding to physical stimuli in a way that ratios of given values represent the ratios of perception. However, Shepard pointed out that there might exist an unknown and undesirable~\textit{response transformation function} which the  direct magnitude estimation method neglects.
 \citep{shepard1981psychological}.

For a much more detailed and in-depth overview of the traditional psychophysical scaling methods see, e.g.~\citet{Gescheider1988}.

\begin{figure}
\centering
\renewcommand{\thesubfigure}{}

    \subfigure[Degree = 0]{
        \centering
        \includegraphics[width = .1\linewidth]{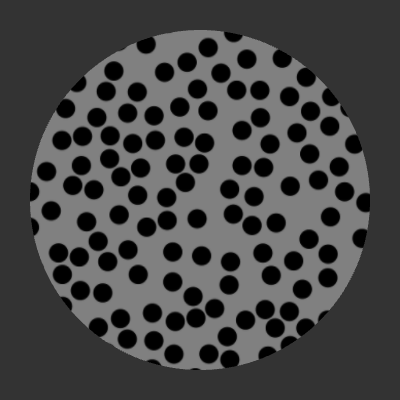}
    }
    \subfigure[Degree = 10]{
        \centering
        \includegraphics[width = .1\linewidth]{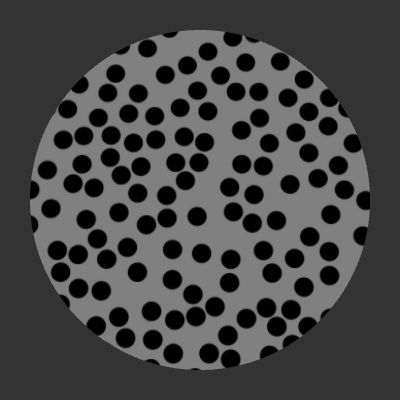}
    }
    \subfigure[Degree = 20]{
        \centering
        \includegraphics[width = .1\linewidth]{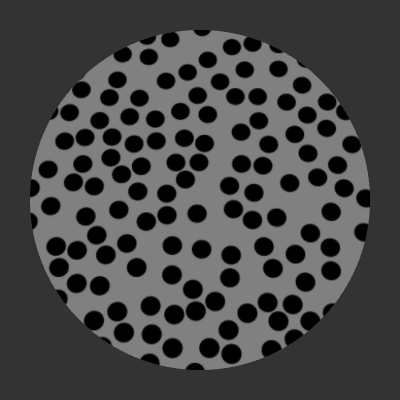}
    }
    \subfigure[Deg = 30]{
        \centering
        \includegraphics[width = .1\linewidth]{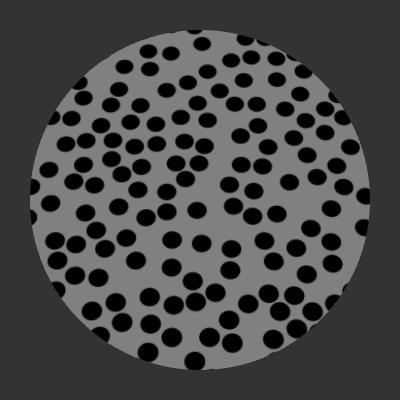}
    }
    \subfigure[Deg = 40]{
        \centering
        \includegraphics[width = .1\linewidth]{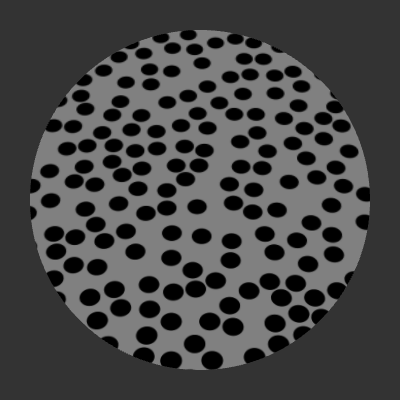}
    }
    \subfigure[Deg = 50]{
        \centering
        \includegraphics[width = .1\linewidth]{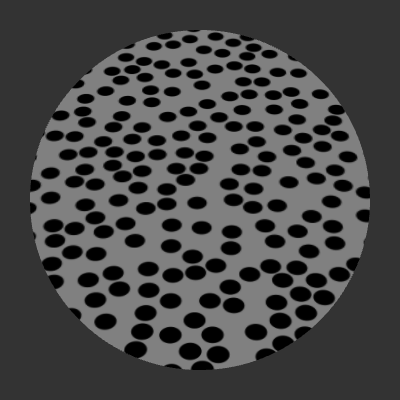}
    }
    \subfigure[Deg = 60]{
        \centering
        \includegraphics[width = .1\linewidth]{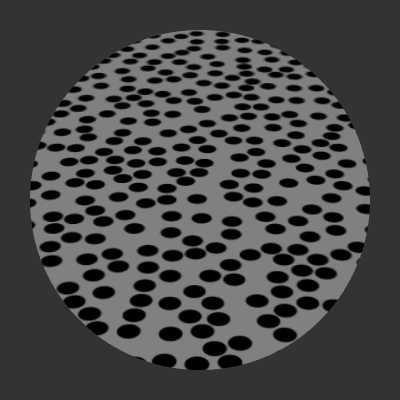}
    }
    \subfigure[Deg = 70]{
        \centering
        \includegraphics[width = .1\linewidth]{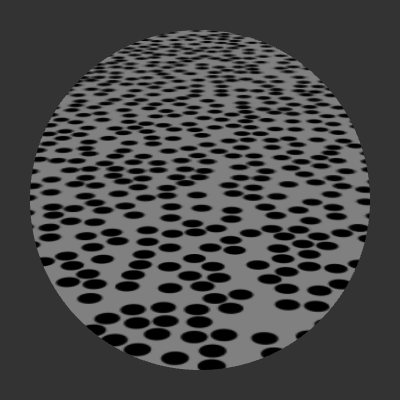}
    }\\
    \subfigure{
    \includegraphics[width = 0.3\linewidth]{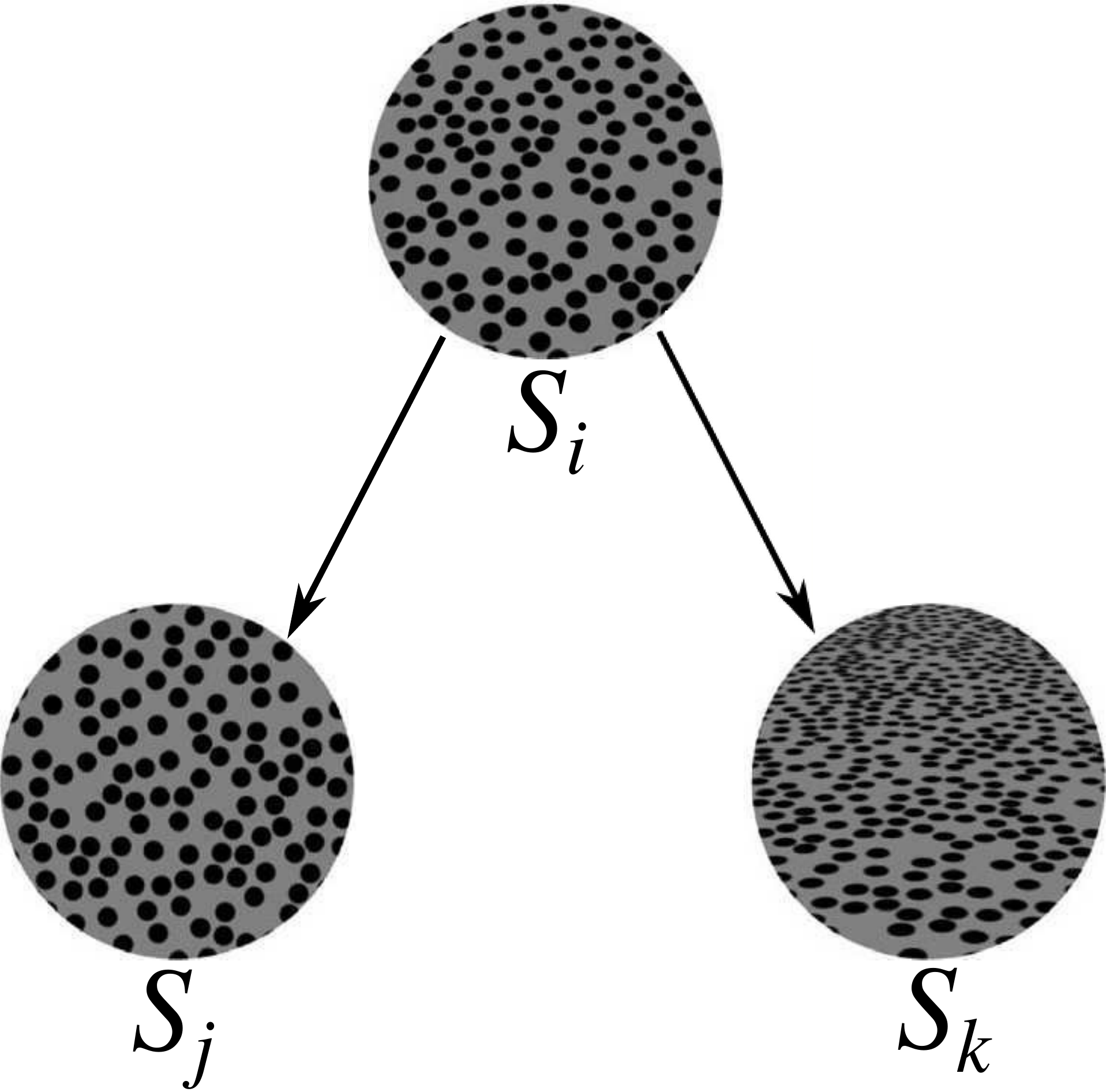}
    }\hspace{2cm}
    \subfigure{
            \centering
            \includegraphics[width = .3\linewidth]{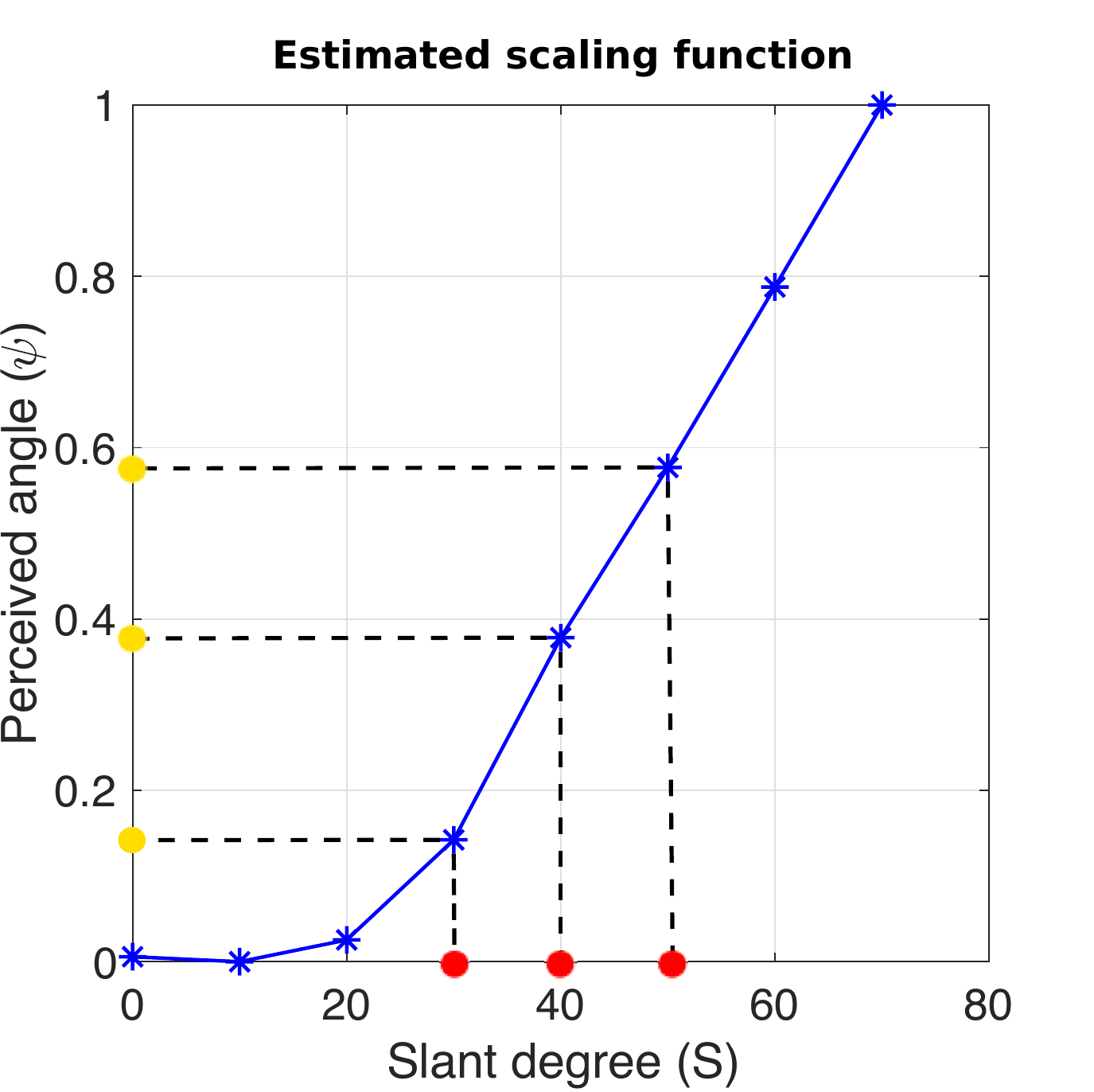}
    }
    \caption{\label{fig:slantExample} Top: Eight stimuli used in the slant-from-texture experiment~\protect\citep{Aguilar2017}. Bottom, left: An example of a triplet question used for the experiment. The triplet question is: ``Which of the bottom images, $S_j$ or $S_k$,  is more similar to the top image $S_i$?'' Bottom, right: The scaling function estimated by the comparison-based embedding method (t-STE). The red points on the X-axis correspond to three stimuli (S), while the yellow points on the Y-axis represents their perceived values ($\psi$). In Section 2 (Embedding methods), we describe in detail, how the position of yellow points corresponds to the ordinal embedding from the triplet questions.}

\end{figure}

\subsection{Scaling and the method of triads}
An alternative approach to data acquisition---neither based on JND-style discrimination nor on direct magnitude estimation---is based on triplet comparisons~\citep{Torgerson_1958}. This approach is often referred to as \emph{method of triads} in the psychophysics literature. Based on a fixed discretization of the physical stimulus, say $S_1, \ldots, S_n$, the method of triads %
asks participants to make comparisons of the form ''Is stimulus $S_i$ more similar to stimulus $S_j$ or to stimulus $S_k$?". In the computer science and machine learning literature such a question is called a \textbf{triplet question} (or,  interchangeably, a triplet comparison).

Rather than attempting accurate quantitative measurements of a particular phenomenon, triplet questions aim at qualitative (ordinal) observations. The obvious potential of such an approach is that the statements do not depend as much on the response transformation function of the observers, and that, e.g. the issue of scaling answers across many observers becomes easier. In addition, there exist studies in the machine learning literature that indicate the robustness of the triplet comparison approach~\citep{demiralp2014learning,li2016extracting}. The obvious challenge of the method of triads is how we can use the participants' answers to estimate the scaling function. More precisely, we need to estimate the magnitudes of perception $\hat{\psi_1}, \ldots \hat{\psi_n}$ in a way that is consistent with the answers to the queried triplet questions.

Let us give an example to clarify the procedure of scaling using the triplet questions. A psychophysical  ``slant-from-texture'' experiment is designed to find the functional relation of the perceived angle with the true angle of a tilted flat plane with a dotted texture \citep{Rosas_etal_2004b,Rosas_etal_2005,Rosas_etal_2007,Aguilar2017}. Figure~\ref{fig:slantExample}~(Top) shows the various stimuli used in the experiment by \citet{Rosas_etal_2004b} and \citet{Aguilar2017}. The bottom, left image of Figure~\ref{fig:slantExample} depicts an example of a triplet question designed for this task. The participant is asked ``which of the two bottom images, $S_j$ or $S_k$, is more similar to the top image $S_i$?'' Based on the answers to a set of such triplet questions, the goal is then to reconstruct the scaling function that describes the relation of perceived angle~$\psi$ and the slant degree~$S$. Figure~\ref{fig:slantExample}~(Bottom, right) shows the function that has been estimated with the t-STE method described below.

The approach of triplet comparisons---the method of triads---is not new to psychophysics; there has been a very long tradition in psychology to explore methods to estimate perceptual (difference) scales from clearly visible \emph{supra-threshold} differences in stimulus appearance \citep{Torgerson_1958,Coombs_etal_1970,Marks_Gescheider_2002}. The earlier approaches are based on inferring a similarity matrix using the triplet comparisons. Recently a more generic approach, called  ``maximum likelihood difference scaling (MLDS)'', has become popular in vision science~\citep{Maloney_Yang03,Knoblauch_Maloney_2010}. There have been reports that both naive, as well as seasoned observers, find the method of triads with supra-threshold stimuli intuitive and fast, requiring less training~\citep{Aguilar2017,Wichmann_etal_2017} than for the more traditional methods in psychophysics such as direct magnitude estimation or in particular methods based on JNDs.

Whilst clearly attractive, MLDS has some limitations, however: First, it makes a strong model assumption, namely that the scaling function is monotonic with respect to the stimulus. Second, the MLDS method can only be used to estimate one-dimensional scaling functions. Thus, it cannot deal with cases when perception is intrinsically multi-dimensional (e.g., color perception). Both issues are of potential relevance in a general psychophysical scaling setting. 

On the other hand, the evaluation of comparison-based data has been an active field of research in computer sciences and machine learning~\citep{Schultz_Joachims2003,AgaEtAl07,tamuz2011adaptively,Ailon11,Jamieson_Nowak2011,VanDerMaaten2012,KleLux14,TerLux14,ukkonen2015crowdsourced,Arias15,JaiJamNow16,haghiri17a}. The core question of these studies is to use the answers to triplet comparisons to find a Euclidean representation of the items (in our case, psychophysical stimuli). This problem is systematically studied in the machine learning literature under the name of \textbf{ordinal embedding}. A number of fast and accurate algorithms have been developed to solve the ordinal embedding problem~\citep{AgaEtAl07,VanDerMaaten2012,TerLux14}. As we will show in this paper, these algorithms may also be useful in psychophysics, vision science and the cognitive sciences in general.

This paper is organized as follows: In Section~2 (Embedding methods) we review two traditional psychophysical scaling methods, NMDS and MLDS, that are used to analyze data from triplet comparisons\footnote{NMDS is not directly based on triplet comparisons; instead it relies on the rank order of dissimilarities, which is, however, closely related to triplet comparisons.}. We then introduce the ordinal embedding problem of the machine learning literature and discuss its advantages in comparison to the traditional embedding methods of psychophysics. Section~3 (Simulations) is dedicated to extensive simulations comparing the performance of ordinal embedding to the applicable competitors in the psychophysics. In Section~4 (Experiments) we examine the ordinal embedding methods in two real psychophysics experiments. In Section~5 (How to apply ordinal embedding methods in psychophysics), we provide instructions and rules of thumb on how to use the comparison-based approach and the ordinal embedding algorithms in psychophysics experiments. In the last section, we conclude the paper by discussing the advantages of the ordinal embedding for scaling problem and mentioning the open problems.

\section{Embedding methods}
\label{sec:Embeddings}

\subsection{Non-metric multi-dimensional scaling (NMDS)}

Non-metric multi-dimensional scaling (NMDS) by Shepard and Kruskal is a well-established method to analyze dissimilarity data~\citep{shepard1962analysis, kruskal1964multidimensional,kruskal1964nonmetric}. It assumes that a complete matrix of dissimilarities (not necessarily metric distances) between pairs of items is given. We denote the dissimilarity of items $i$ and $j$ by $\delta_{ij}$. In the context of psychophysics, this matrix usually comes from a human (psychophysical) experiment. Shepard posed the problem of estimating a d-dimensional Euclidean representation of items, say $y_1,y_2,\ldots y_n \in \RR^d$, such that the pairwise distances of estimates are consistent with a monotonic transform of the given dissimilarities. Key to the method is that it only takes the rank order of the dissimilarities into consideration. The reason is that in many psychophysics experiments, the magnitude of dissimilarities cannot be quantitatively measured, whereas the rank order of distances is considered to be more reliable---an argument we have made in favour of ordinal embedding, too (see above).

If $d_{ij} = \Vert y_i - y_j \Vert$ is the Euclidean distance of embedded items $y_i$ and $y_j$ in $\RR^d$, then the quality of a Euclidean representation is measured by a quantity called \textbf{stress}~\citep{kruskal1964nonmetric}:

\begin{equation}\label{eq:stressKruskal}
\text{stress} = \frac{\sum_{ij}{(d_{ij} - g(\delta_{ij})})^2}{\sum_{ij}{{d_{ij}}^2}}, 
\end{equation}

where $g$ is a monotonic function to be determined. 
The smaller the stress, the better the Euclidean representation. The numerator measures the squared loss between the transformed input dissimilarities  $g(\delta_{ij})$ and the Euclidean distances $d_{ij}$. By minimizing the stress we try to achieve that the distances $d_{ij}$ are as close as possible to the monotonic transform of dissimilarities $g(\delta_{ij})$. The role of the denominator is to prevent the degenerate solution, where $d_{ij}$ and $g(\delta_{ij})$ become infinitesimal together.

The goal of NMDS is to find the Euclidean representation of items that minimizes the stress function, where $g$ can be chosen from the set of all monotonic transform functions. The approach by Kruskal~\citep{kruskal1964multidimensional} finds an estimation of the optimal solution through a two-step optimization procedure. In the first step, a configuration of embedding points $y_1,y_2,\ldots y_n$ is fixed; this means that the distance values $d_{ij}$ are also fixed. Then a greedy algorithm is suggested (later called isotonic regression) to find the monotonic function $g$ that minimizes the stress function. In the second step of optimization the values of $g(\delta_{ij})$ are fixed and the embedding points $y_1,y_2,\ldots y_n$ are adjusted by a gradient descent algorithm to minimize the stress. The two steps are repeated consecutively until the stress value shows no further improvement or it becomes smaller than a certain threshold.

The NMDS algorithm has been used extensively in psychology~\cite{Reed1972,Smith1985,barsalou2014cognitive}, neuroscience~\citep{OpDeBeeck2001,Kayaert2005,Kaneshiro2015} and broader fields~\citep{Lester1987,Machado2015}. The non-parametric flavor of the method makes it a general purpose algorithm that is easy to apply. In addition, it can find representations in multi-dimensional spaces. However, there are two major drawbacks: First, the proposed optimization algorithm tries to solve a highly non-convex optimization problem, and typically gets stuck in a local, but no the global minimum of the stress function. This local optimum can be arbitrarily far off from the global optimum. 
The second issue is the requirement on the input data: as described above, the algorithm needs the full dissimilarity matrix as input. Alternatively, in a setting of triplet comparisons one can also implement the algorithm with just the knowledge on the ranking (ordering) of all the distance values $\delta_{ij}$. This ordering can be computed from triplet questions, but it requires in order of $n^2 \log n$ triplet questions to sort all pairwise distances. This property makes NMDS infeasible for many applications in psychophysics, as the number of required triplet comparisons get too large.

To get a feeling, consider the cases of  $n=15$ and $n=50$ stimuli. Suppose that we have $n=15$ stimuli. There exist ${15 \choose 2}=105$ dissimilarity values. The NMDS method requires full order of distances. This means on average it needs to ask $105 \cdot \log_2{105} \approx 700$ triplet questions. The amount of triplet comparisons for embedding methods depends on the embedding dimension. Let assume the embedding is in 2-dimensional space. Then, the embedding methods require in order of $nd\log_2{n}$ triplet comparisons. In this example, this amount would be $15\cdot2\cdot\log_2{15} \approx 120$. Figure~\ref{fig:SimsColor} shows the result of a simulation in a similar setting, where $n=14$. Even though the embedding method (Figure~\ref{fig:SimsColor}~(c)) uses less information, it produces an embedding with higher quality. The number of stimuli is $n=14$ and embedding dimension is 2.The difference is even more drastic with larger $n$. If we assume $n=50$, with the same calculations, NMDS requires about 12570 triplet comparisons, whereas the ordinal embedding methods require about 570 triplet comparisons.

\subsection{Maximum likelihood difference scaling (MLDS)} 

Decades after the introduction of NMDS, maximum likelihood difference scaling (MLDS) was proposed to solve a specific instance of the difference scaling problem~\citep{knoblauch1998difference,Maloney_Yang03,krantz2007foundations}. Originally MLDS asked quadruplet questions which involve four stimulus levels. If we denote the perceptual scale of four stimuli $S_i,S_j,S_k,S_l$ by $\psi_i,\psi_j,\psi_k,\psi_l$, then a quadruplet question asks whether the difference in perception $\vert \psi_i -\psi_j\vert$ is larger or smaller than the difference of perception $\vert \psi_k -\psi_l\vert$. Note, however, that triplet questions are indeed a subset of quadruplet questions, implying that the MLDS method is also applicable to triplet questions.

There are two main assumptions in the MLDS model. First, it assumes that the perceptual scale is a scalar (one-dimensional) denoted by $\psi$. Second, the MLDS method assumes that the perceptual scale grows monotonically with respect to the stimulus. In particular, it assumes that the order of two stimuli in the physical space implies the same order in the perceptual scale: $S_i<S_j \Rightarrow \psi_i < \psi_j$.

In contrast to NMDS, the MLDS method uses a parametric model; for a quadruplet of stimulus levels $S_i,S_j,S_k,S_l$, for simplicity denoted by $(i,j;k,l)$, a decision random variable is defined as 

\[
Dec(i,j;k,l) = \vert \psi_i - \psi_j \vert - \vert \psi_k - \psi_l \vert + \epsilon,
\]

where $\epsilon \sim \mathcal{N}(0,\sigma^{2})$ is a zero-mean Gaussian noise with standard deviation $\sigma>0$. If $Dec(i,j;k,l) > 0$, then the observer would respond that the pair $(i,j)$ has a larger difference than the pair $(k,l)$. In this case the response to the quadruplet $q=(i,j;k,l)$ is set to $R_q=1$, otherwise the response is $R_q=0$. The goal of the MLDS is now to estimate the perception scale $\psi$ that  maximizes the likelihood of the observed quadruplet answers. We first set $\psi_1 = 0, \psi_n = 1$ to remove degenerate solutions. Now, assuming that $R_1,R_2, \ldots, R_m \in \{0,1\}$ denote the independent responses to $m$ quadruplet questions, the likelihood of the perceptual scales given the quadruplet answers is 

\[
\mathcal{L}\left( \psi_2,\ldots,\psi_{n-1},\sigma \vert R_1,\ldots,R_m\right) = \prod_{q=1}^m{\Phi(\Delta_q)^{R_q}[1-\Phi(\Delta_q)]^{1-R_q}},
\]

where $\Phi(.)$ denotes the cumulative distribution function of $\epsilon \sim \mathcal{N}(0,\sigma^{2})$, and $\Delta_q = \vert \psi_i - \psi_j \vert - \vert \psi_k - \psi_l \vert$ for the quadruplet $q=(i,j;k,l)$. This likelihood is not convex with respect to the perceptual scale values $\psi_i$. Thus, the proposed numerical methods to maximize this likelihood might end get stuck in only a local maximum.

There are a number of advantages of the MLDS method: The maximum likelihood estimator is unbiased and has minimum variance among the unbiased estimators. As a practical advantage, it has been shown empirically that a small subset of quadruplets is enough for the convergence of the algorithm. Finally, it has been shown that the variance of the output behaves reasonable with respect to the input noise level~\citep{Maloney_Yang03}.

However, MLDS also has a some drawbacks. First, the algorithm only works for a one-dimensional perceptual spaces. In some cases (see the examples of color and pitch perception in Figure~\ref{fig:MultiDimExamples}) the scales need more than one dimension, however. Second, even in the one-dimensional case, the assumption on the monotonicity of the scaling function is restrictive and may not hold for all psychophysical settings. Finally, the nice theoretical properties (unbiasedness, minimum variance solution) only hold for the global optimum of the MLDS likelihood function, but not for the local optima that are realistically obtained by the optimization algorithm. 

\subsection{Ordinal embedding}

\subsubsection{General setup}

The comparison-based setting has recently become popular in machine learning literature~\citep{Schultz_Joachims2003,AgaEtAl07,VanDerMaaten2012,amid2015multiview,ukkonen2015crowdsourced,balcan2016Learning}. Instead of stimulus levels, in machine learning we deal with a set of abstract items, say $x_1,x_2,\ldots,x_n$ that come from some abstract space  $\mathcal{X}$. Furthermore, we assume that there exists 
a dissimilarity function $\delta: \mathcal{X} \times \mathcal{X} \to \mathbf{R}$ that describes the dissimilarity of the items. Often, in machine learning we assume that $\delta$ is symmetric, but not necessarily a metric. In our current setting in psychophysics, we now assume that the function  $\delta$ is not available, yet we have access to an oracle which responds to a triplet question $t = (i,j,k)$, based on the unknown dissimilarity. The triplet question will be ``Is item $x_i$ more similar to item $x_j$ or item $x_k$''? We denote this triplet question by $t = (i,j,k)$. The response to the triplet is denoted by $R_t$ and stored as the following:
\begin{equation}\label{eq:tripletAnswer}
R_t =
\begin{cases}
1 & \text{if the oracle responds } x_j \text{ is more similar to } x_i \\
-1 & \text{if the oracle responds } x_k \text{ is more similar to } x_i
\end{cases}
\end{equation}

Assume that the answers to a subset of triplet questions $T \subset \{(i,j,k)\vert x_i,x_j,x_k \in \mathcal{X}\}$ are collected from the oracle. Given an embedding dimension $d$ and the answers to the triplet questions $T$, the ordinal embedding aims to find points $y_1,y_2,\ldots y_n \in \RR^d$ in a d-dimensional Euclidean space whose Euclidean distances are consistent with the answers of the queried triplet questions. The consistency of an embedding with respect to triplet $t= (i,j,k)$ can be judged as following: 

\[ 
R_t \cdot \sgn(\Vert y_i - y_j \Vert^2 - \Vert y_i - y_k \Vert^2) =
\begin{cases}
1, \hspace{1cm}\text{if the embedding is consistent with } R_t \\
-1, \hspace{.75cm}\text{if the embedding is not consistent with } R_t
\end{cases}
\]

where function $\sgn$ returns the sign of a real value. 
The goal of \textbf{ordinal embedding} is to find an embedding $y_1, ..., y_n$ that maximizes the number of consistent triplets. Intuitively, we would like to solve the following optimization problem:
\begin{align}\label{eq:ordinalEmbedding}
\max_{y_1,\ldots, y_n \in \RR^d}{\sum_{t = (i,j,k)\in T}{R_t \cdot \sgn(\Vert y_i - y_j \Vert^2 - \Vert y_i - y_k \Vert^2)}}. 
\end{align}
However, there are major algorithmic obstacles. It is not always possible to find a perfect $d$-dimensional embedding for an arbitrary dissimilarity function $\delta$. Moreover, in a practical setting the answers to the triplets might be noisy. Therefore, the optimal solution is not necessarily consistent with the full set of triplets $T$. And finally, as written above the objective function is discrete-valued, which makes it even harder to optimize. Hence, various adaptations of the stress function and optimization heuristics are used to address these problems. For the purpose of this exposition, we want to keep it at this intuitive level, below we describe one particular algorithm in more detail.

\subsubsection{Connection to the scaling problem} 

One can see that ordinal embedding solves the scaling problem of psychophysics in the following way: the different stimuli $S_i$ play the same role as the abstract items $x_i$ in the ordinal embedding problem, and the perception values $\psi_i$ correspond to the embeddings $y_i$. Concretely, given a standard scaling function as in Figure~\ref{fig:slantExample} (bottom right), the ordinal embedding output corresponds to the positions of the perception values on the y-axis (yellow points in Figure 2, bottom right). Thus, given the ordinal embedding output (y-values) and the values of the physical stimuli, we can reconstruct the scaling function.%

To make it concrete once more in the specific example of the slant-from-texture problem in  Figure~\ref{fig:slantExample}): Given the slant stimuli $S_1, ..., S_n$, participants were asked a number of triplet questions involving the stimuli $S_1, ..., S_n$. Then we fed the answers of these triplet questions to an ordinal embedding algorithm and asked the algorithm to construct a 1-dimensional embedding. This resulted in the yellow points $y_1, ... y_n$ on the y-axis. We only depicted three stimuli out of eight with yellow points, in order to keep the plot neat. These points can now be identified as the perception values $\psi_1, ..., \psi_n$, so we can finally draw the scaling function by connecting the points $(S_i, \psi_i)$. More details on this experiment are provided in the Experiments section.

\begin{figure}
\centering
\begin{subfigure}
    \centering
    \includegraphics[width = .3\linewidth]{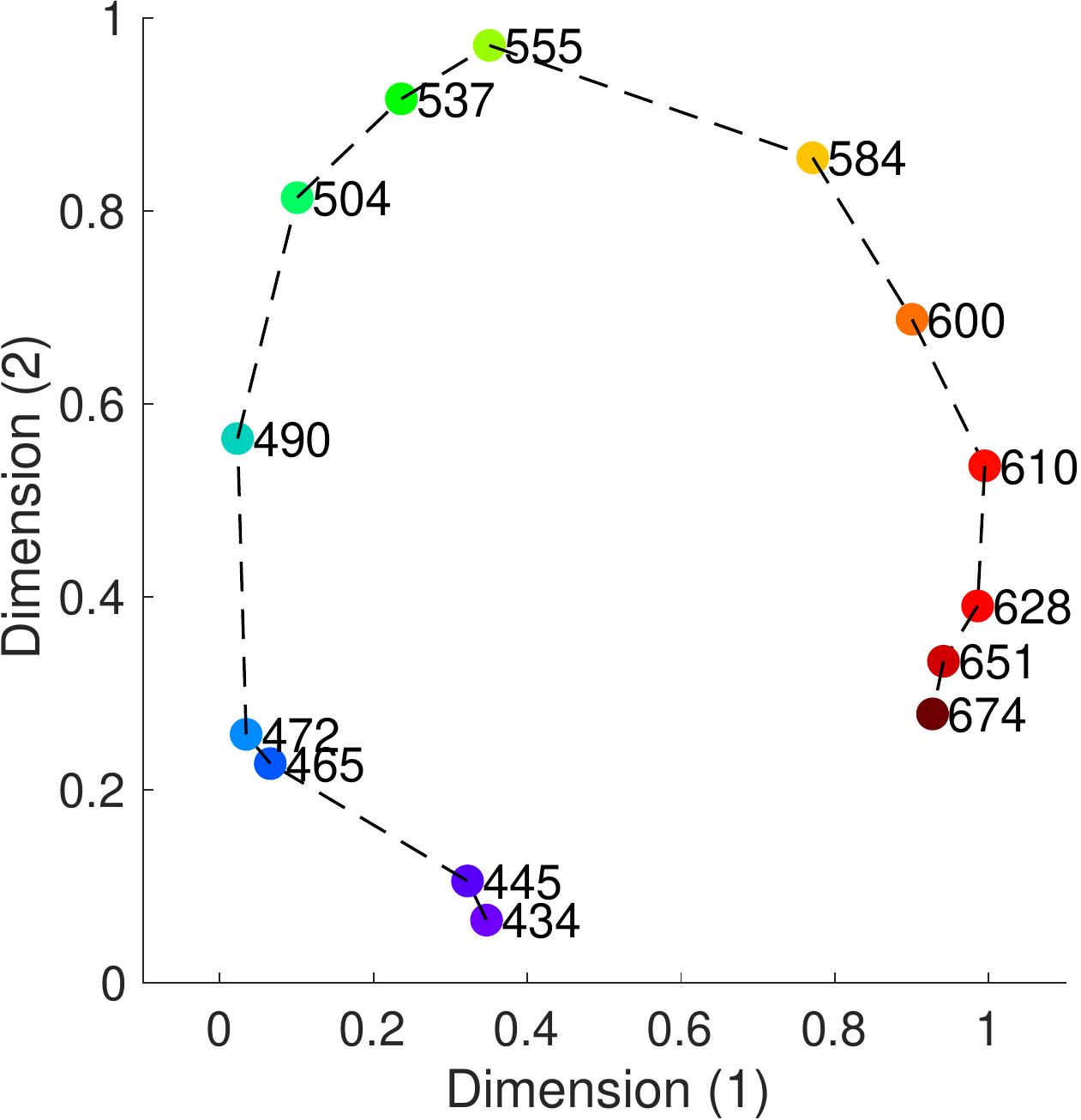}
\end{subfigure}
\begin{subfigure}
    \centering\hspace{10mm}
    \includegraphics[width = .25\linewidth]{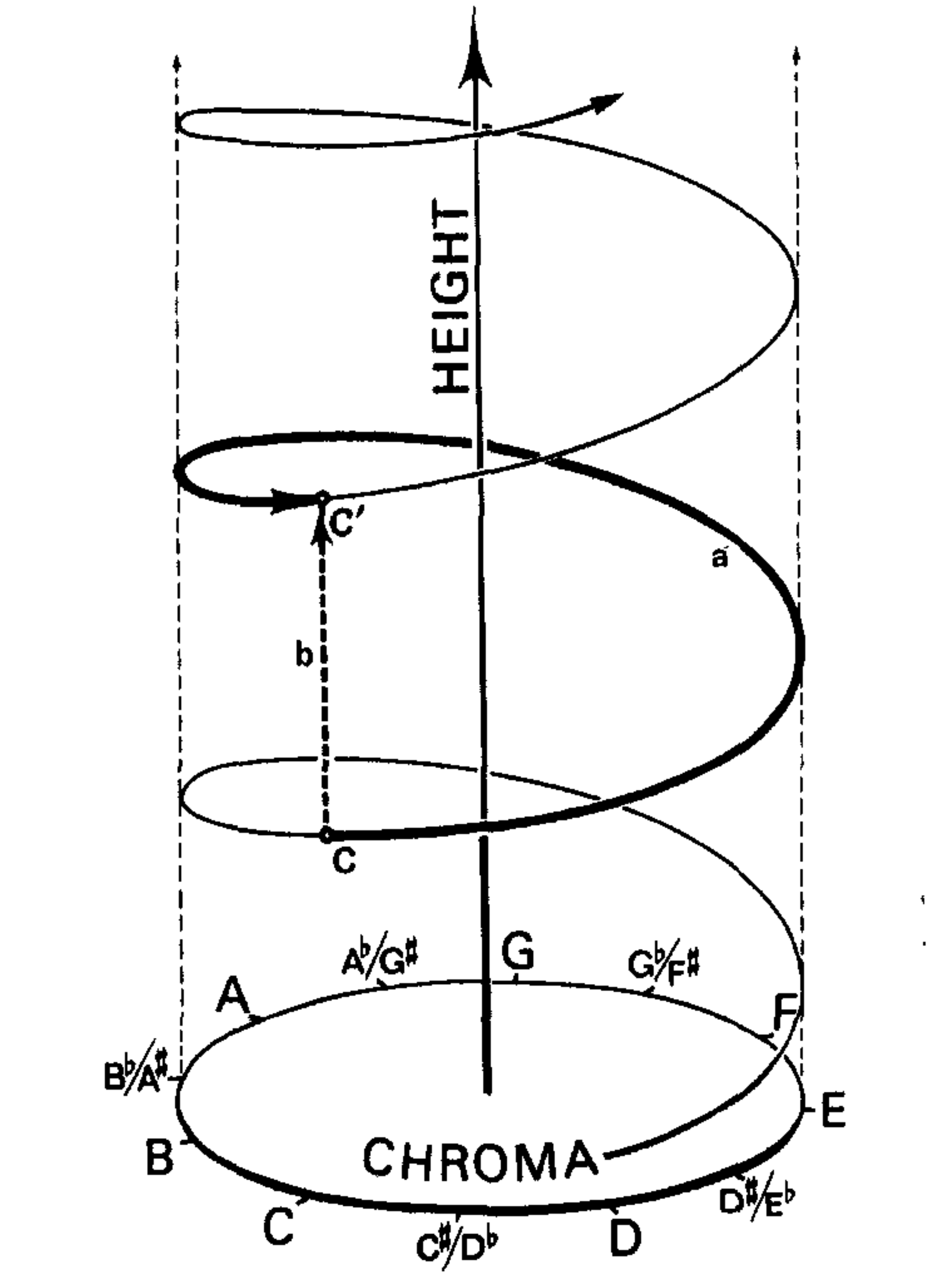}
\end{subfigure}
\caption{\label{fig:MultiDimExamples}Left : The two-dimensional circle of color perception gathered by similarity measurements between 14 colors~\protect\citep{shepard1962analysis}. The \textit{wavelength} of each color is written on the right side of the colored dot. Note that we have reconstructed the color circle with the non-metric MDS method based on the original dissimilarity data. Right: The helix proposed by Shepard for the pitch perception. The physical stimulus, i.e. pitch, varies along the spiral path of the curve and the three-dimensional space describes the perception~\protect\citep{Shepard1982pitch}. Note that the visualization is done in three dimensions, however, two parameters, called \textit{chroma} and \textit{height} by Shapard, are enough to describe the perception. %
}
\end{figure}

While in the example of the slant experiment we used a one-dimensional embedding, ordinal embedding methods can also construct a \textit{multi-dimensional} embedding that describes the perceptual space of humans. Let us discuss two examples that demonstrate why this might be important. 
piOne almost ``famous'' example is {\textit color perception.}  Figure~\ref{fig:MultiDimExamples}~(Left) shows the two-dimensional color circle proposed by Shepard and Ekman~\citep{shepard1962analysis,ekman1954dimensions}. The figure has been constructed with the NMDS algorithm based on a 14$\times$14 similarity judgment matrix. The wavelength of each color is written at the right side of each colored dot. In our context the important observation is that human observers perceive the the violet colors with low wavelengths as similar to the red colors with high wavelengths. This suggests a circular perceptual internal space, which can only be realized in at least two dimensions. A second example is \emph{pitch perception} of sounds. Even though auditory frequency is again one-dimensional, the pitch is perceived along a three-dimensional helix~\citep{Shepard1982pitch,houtsma1995pitch}. Figure~\ref{fig:MultiDimExamples}~(Right) shows the proposed perception space by Shepard. In both cases, pitch and color, the multi-dimensional ordinal embedding can enable the researcher to find perceived values in a higher-dimensional Euclidean spaces that properly capture the similarity-structure of perception.

\subsubsection{Stochastic triplet embedding}

In recent years, there has been a surge of methods to address the ordinal embedding problem in the machine learning community, for example generalized non-metric multidimensional scaling (GNMDS)~\citep{AgaEtAl07}, the crowd-median kernel~\citep{tamuz2011adaptively}, stochastic triplet embedding (STE)~\citep{VanDerMaaten2012} and local ordinal embedding(LOE)~\citep{TerLux14}. In general, the focus of the machine learning community is to build methods that require only a small number of triplets to embed a large number of items, make as few assumptions as possible, and to be robust towards noise in the data. 

In the following we focus on one particular class of methods, stochastic triplet embedding (STE) and its variant t-distributed stochastic triplet embedding (t-STE), because in our experience they work very well and are based on a simple model that is also plausible in a psychophysics setting. The STE method introduces the probabilistic model defined in Equation~\eqref{eq:tsteModel} to solve the ordinal embedding problem. Assume that $y_1, \ldots, y_n \in \RR^d$ were the correct representations of our objects. The model assumes that if a participant is being asked whether $y_i$ is closer to $y_j$ or to $y_k$, then he gives a positive answer with probability 

\begin{align}\label{eq:tsteModel}
p_{ijk} = \frac{\exp(-{\Vert y_i - y_j \Vert^2})}
{
\exp(-{\Vert y_i - y_j \Vert^2})+
\exp(-{\Vert y_i - y_k \Vert^2})
}.
\end{align}

Intuitively, ``easy'' triplet questions (where the distances 
$\Vert y_i - y_j \Vert$ and $\Vert y_i - y_k \Vert$ are very different) will be answered correctly in most of the cases, whereas difficult triplet questions (where $\Vert y_i - y_j \Vert$ is about as large as $\Vert y_i - y_k \Vert$) can easily be mixed up. 
Given the answers to a set of triplets, the STE algorithm attempts to maximize the likelihood of the embedding point configuration with respect to the answered triplets. If the answer to a triplet question $t=(i,j,k)$ is given according to Equation~\ref{eq:tripletAnswer}, and if we assume that triplet questions are answered independently, the likelihood of an embedding given the answers to a set of triplets $T$ is given as 

\[
\mathcal{L}\left( y_1, \ldots, y_n \vert R_1, \ldots, R_{\vert T \vert}\right) = \prod_{t=(i,j,k) \in T, R_t = 1}{p_{ijk}}\cdot \prod_{t=(i,j,k) \in T, R_t = -1}{(1-p_{ijk})}.
\]

The log-likelihood is maximized to find the solution of ordinal embedding. In the above formulation, the probability of satisfying a triplet goes rapidly to zero when the difficulty of a triplet question increases. As a result, a severe and a slight violation of a triplet are penalized almost the same. To make the statistic more robust, the authors propose to replace the Gaussian functions with Student-t functions with a heavier tail kernel~\citep{VanDerMaaten2012}. The modified method is called t-distributed STE (t-STE).

This algorithm can deal with a large number of items (stimulus levels) and reasonable number of triplets, and it is robust to noise, which is an important characteristic when dealing with psychophysics data. Unlike MLDS, one-dimensional functional relations (embeddings) are not restricted to monotonic functions, and the algorithm is capable of embedding in higher dimensional Euclidean spaces. However, as all other ordinal embedding methods, the proposed optimization problem is not convex, which makes it vulnerable to inappropriate local optima.

\subsection{Summary of embedding methods}

In Table~\ref{tab:EmbSummary}, we summarize the properties of the different embedding methods. The ordinal embedding methods can produce high quality results with a small set of triplet answers. This property makes them superior to traditional NMDS that requires the full order of distances. On the other hand, the embedding methods are not limited to the case of one-dimensional monotonic functions, as it is the case for MLDS.

As the number of items (and consequently the number of triplets) grows, the ordinal embedding algorithms become drastically slow. This is however, more of a concern for machine learning purposes which deal with thousands of items and hundreds of thousands of triplets. The algorithms (particularly STE and t-STE) have a quite acceptable running time for standard psychophysics experiments. Our experiments are performed on an iMac 18.3 (2017) with a 3.4 GHz i5 quad-core processor. On this machine, the (t)-STE algorithm, implemented in MATLAB, requires about 30 minutes to embed 100 items in two dimensions using 2000 triplet answers.

\begin{table}[]
    \centering
    \begin{tabular}{c|c|c|c|c}
         Method &  Data required & Statistical noise model & Multi-dimensional & Restrictions \\ \hline\hline
         NMDS & Complete order of distances & No & Yes &  --- \\
         MLDS & Partial set of quadruplets & Yes & No & Monotonic functions \\
         t-STE & Partial set of triplets & Yes & Yes & ---
    \end{tabular}
    \caption{The comparison of ordinal embedding methods. Each row corresponds to one method, while the properties are listed in the columns.}
    \label{tab:EmbSummary}
\end{table}

\section{Simulations}

In this section, we compare ordinal embedding with the traditional embedding approaches in psychophysics (NMDS and MLDS) with diverse simulations. We consider one-dimensional and two-dimensional perceptual spaces. %

\subsection{Simulation setup}

{\bf Stimulus and perceptual scale:} 
We assume that the stimulus and perception are measured on a scale from 0 to 1, and the true relation between the physical stimulus and the perception is encoded by a function $f: (0,1) \to (0,1)^{d}$, where the dimension $d$ of the perceptual space will be 1 or 2. We consider $n$ uniformly chosen steps for the stimulus levels, denoted by $S = \{S_1,S_2,\ldots,S_n\}$. In our simulations we assume that a true perceptual scale exists, for the stimulus $S_i$ is denoted by $y_i = f(S_i)$. We will choose different functions $f$ for our different simulations below.\\

{\bf Generating subsets of triplet questions:} Our goal will be to feed the same number of triplet questions to each of our algorithms. However, we need to be a bit careful regarding which comparisons are ``valid'' for each of the algorithms. 
The MLDS method assumes that the perceptual function is monotonic. If we consider three stimuli $S_i,S_j,S_k$ such that $i<j<k$, then MLDS always assumes that $\delta(i,k) > \delta(i,j)$ and $\delta(i,k) > \delta(j,k)$---from the MLDS point of view, it would be useless to ask a participant for the comparison of $\delta(i,k)$ to $\delta(i,j)$. The only useful triplet question for MLDS is whether $\delta(j,i)$ is smaller or larger than the distance $\delta(j,k)$. Thus, for any set of three stimuli, there is only one ``valid'' triplet question, leading to a total of ${n \choose 3}$ valid triplet questions for MLDS. All other algorithms under consideration do not make any monotonicity assumption. Here, for any set of three different stimuli we can ask three useful triplet questions, leading to a total of $3{n \choose 3}$ valid triplet questions. 

In all our simulations, we feed the same number of triplets to all embedding algorithms. A random subset of triplets are chosen without replacement from the set of all valid triplets for each algorithm, where this set of valid triplets is different for MLDS and the other algorithms, as described above. In our simulations, the size of the random subset of triplets will be chosen in the range $r\cdot{n \choose 3}$ with $r \in \{0.2,0.4,0.6,0.8,1\}$. The value $r=1$ is equivalent to choosing the whole set of valid triplets for the MLDS method and a third of the set of valid triplets for the other methods.

Note that MLDS and ordinal embedding methods get triplet answers as input. However, NMDS needs a dissimilarity matrix. Therefore, we design a fair procedure, explained later in this section, to construct the required dissimilarity values.

{\bf Underlying model to generate triplet answers:}
In order to generate answers to the triplet questions, we construct a model that resembles a typical observer of a psychophysical experiment. Given a fixed perceptual scale function $f$, we assume that the simulated observer answers the triplet questions based on a noisy version of this function, denoted by $\tilde y_i = f(S_i) + \epsilon$, where $\epsilon \sim \mathcal{N}(0,\sigma \cdot{I}_d )$ is a zero-mean Gaussian noise with unit covariance matrix and standard deviation $\sigma$ in d dimensions. In our simulations we use $\sigma$ in the range of $\{0.01,0.05,0.1,0.5\}$. The simulated observer produces the answer to the queried triplet question $t = (i,j,k)$ by 

\[
R_t=
\begin{cases}
1 & \text{if } \Vert \tilde y_i - \tilde y_j \Vert < \Vert \tilde y_i - \tilde y_k \Vert, \\
-1 & \text{otherwise}.
\end{cases}
\]

Note again that the embedding values $y$ play the same role as the perceptual scale values $\psi$ in the psychophysics notation. We sometimes use a different notation to emphasize that the embedding values $y$ can be multi-dimensional, and to make a clear distinction to scalar values of $\psi$. The scalar $\psi$ is depicted in Figure~\ref{fig:scalingExample} and also in the MLDS method.

{\bf Feeding triplet answers to the algorithms.}
The above-mentioned model produces answers to the triplet questions. In case of (t)-STE, these triplet answers directly serve as the input to the algorithm. The same is true for MLDS, because we took care to only sample valid triplets. 
For NMDS, however, we need to proceed differently. NMDS requires dissimilarities between pairs of items (but in the end only makes use of the order between these values due to the monotonic transformation function $g$ used in the definition of stress in Equation~\eqref{eq:stressKruskal}). %
For our simulations, we generate a set of noisy perceptual values for $n$ stimuli levels as before, $\tilde y_i = f(S_i) + \epsilon$, and then explicitly compute all dissimilarity values $\delta_{i,j} = \Vert \tilde y_i - \tilde y_j \Vert$. These are then the values that we give to the NMDS algorithm. Strictly speaking, the NMDS algorithm thus gets to see more information about the data, but it does not access it because in the end it only considers the ordering between the dissimilarities. Also note that because NMDS requires the full matrix of dissimilarities, we only apply and compare the NMDS algorithm with MLDS and the embedding methods when $r=1$, that is all triplet questions are being asked. This procedure makes sure that all three algorithms get the same amount of information.

{\bf Embedding methods:} We now apply various algorithms to generate embeddings or perceptual scales. For STE and t-STE we use the MATLAB implementation by \citet{VanDerMaaten2012}~\footnote{\url{https://lvdmaaten.github.io/ste/Stochastic_Triplet_Embedding.html}}. We use the default optimization parameters for both methods. The degree of freedom for the t-Student kernel is set to $\alpha=1$ for the t-STE method. We also use the R-implementation of a second algorithm from the machine learning community, local ordinal embedding (LOE)\footnote{\url{https://cran.r-project.org/web/packages/loe}}, with the default parameter settings. For MLDS, we use the R-package available on CRAN repository\footnote{ \url{https://cran.r-project.org/package=MLDS}}, again with the default optimization parameter settings. For the NMDS algorithm, we use MATLAB implementation, which is available by calling the function ``mdscale''. The implementation optimizes the stress function defined defined by~\citet{kruskal1964multidimensional}; see Equation~\eqref{eq:stressKruskal}.

In all cases, we set the embedding dimension to the dimension of true perceptual function. In the section on real experiments we also consider cases where the embedding dimension is not known. 

All embedding methods solve a non-convex optimization problem and thus are prone to find inaccurate local optima. To reduce this effect, we run all the algorithms 10 times with random initializations. Among the 10 embedding outputs we choose the one which has the least triplet error (see next subsection for a definition). 

Independent of the above repetition, which is done to remove the effect of local minima, each embedding method is executed 10 times, on 10 independent draws of the random input data.  This repetition is meant to analyze the average behaviour and the variances of the algorithms. We finally report the average values of the evaluations over the 10 repetitions. The standard deviations are reported in the supplementary material.

\begin{figure}[t]
\centering
\subfigure[\hspace{-6mm}]{
    \centering
    \includegraphics[width = .175\linewidth]{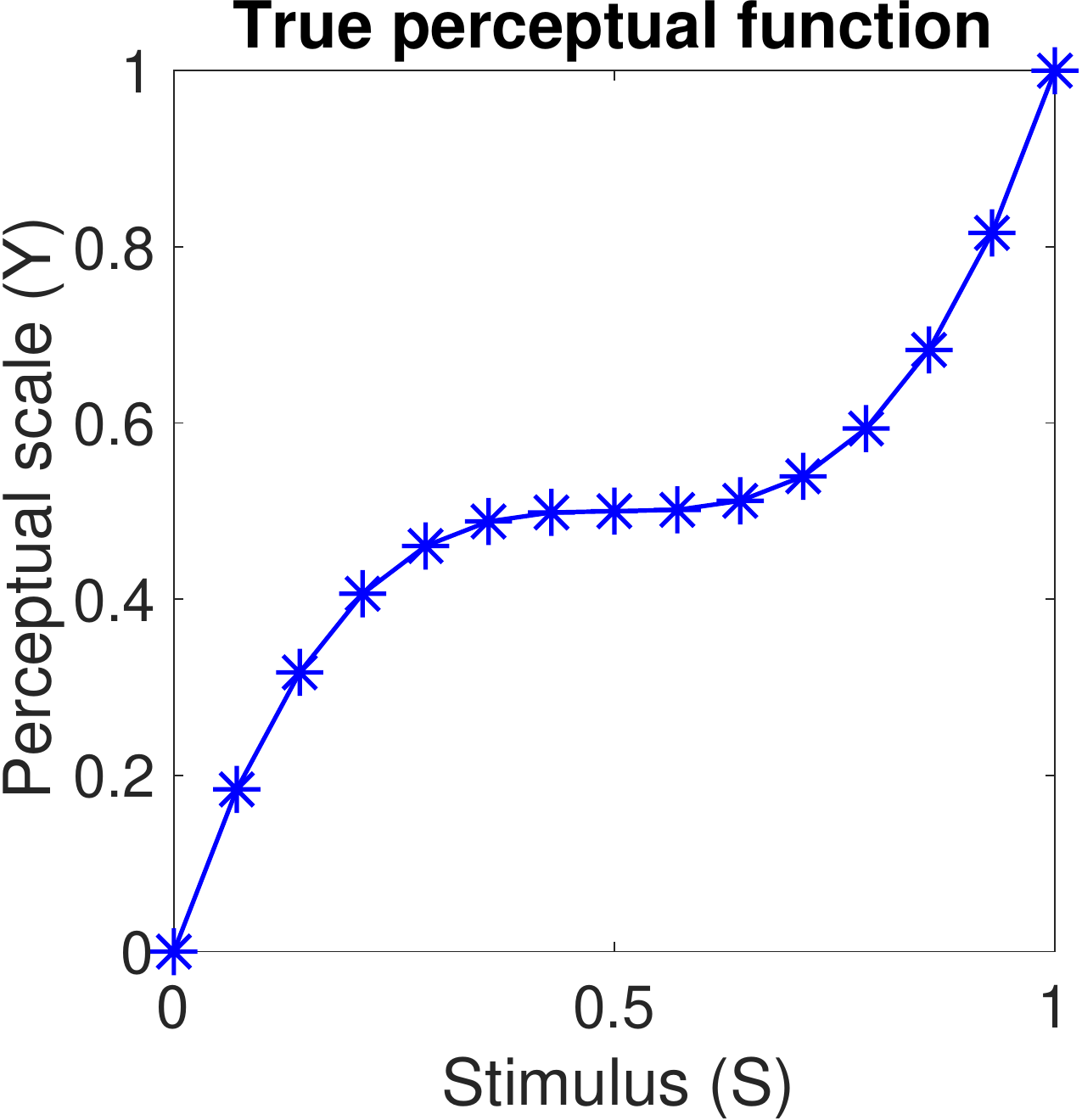}
}
\subfigure[\hspace{-5mm}]{
    \centering
    \includegraphics[width = .175\linewidth]{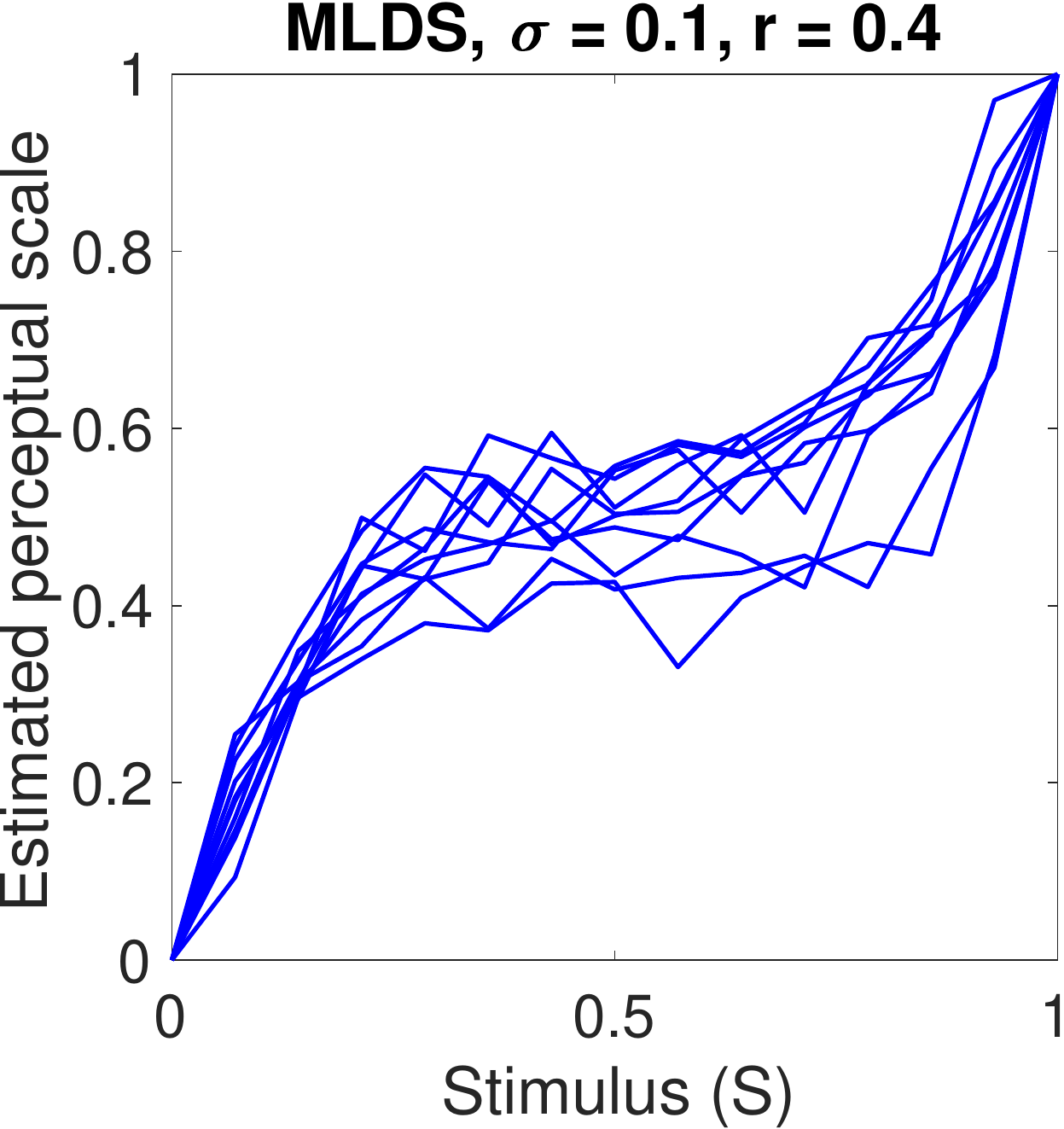}
}
\subfigure[\hspace{-5mm}]{
    \centering%
    \includegraphics[width = .175\linewidth]{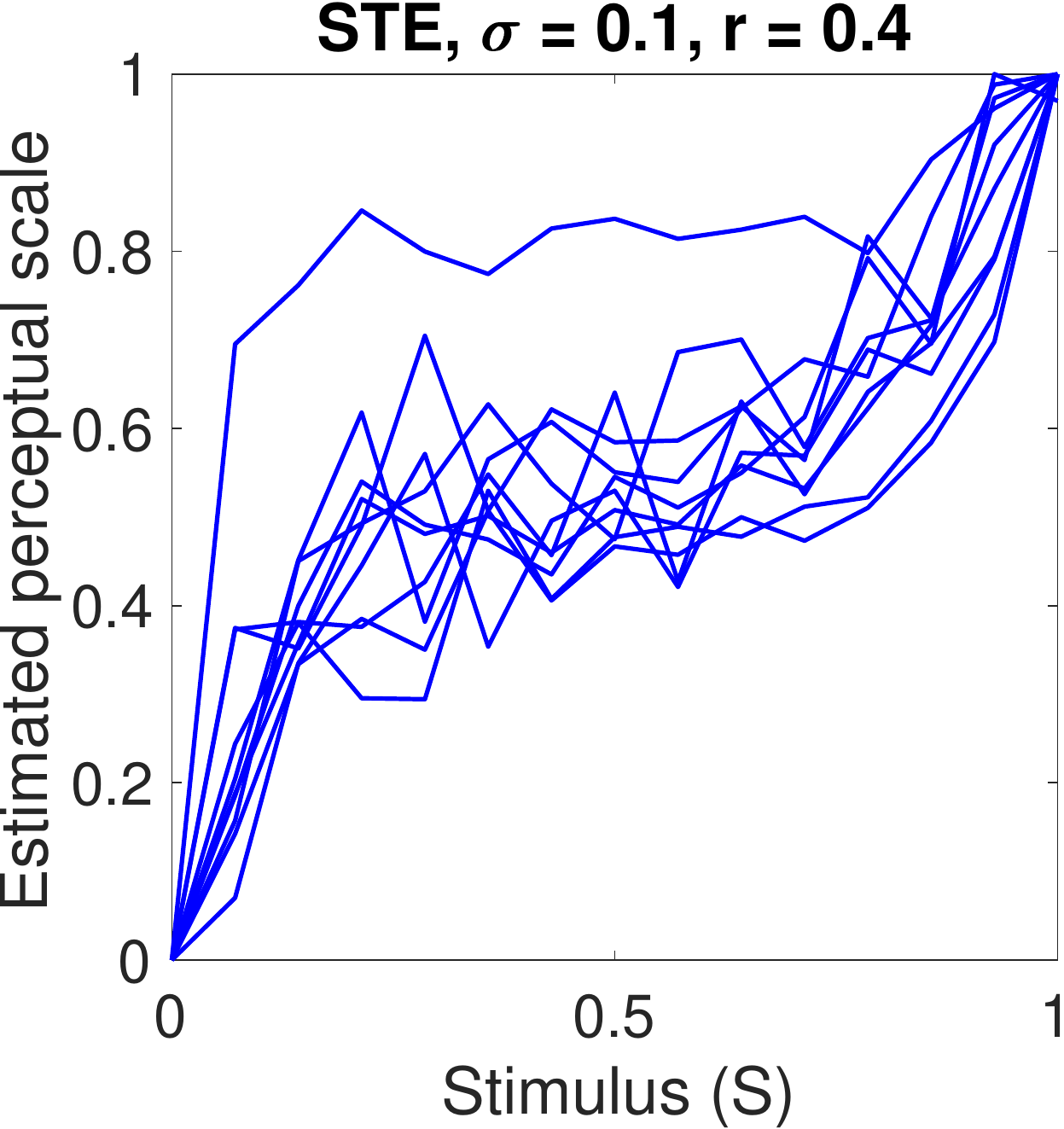}
}
\subfigure[\hspace{-5mm}]{
    \centering%
    \includegraphics[width = .185\linewidth]{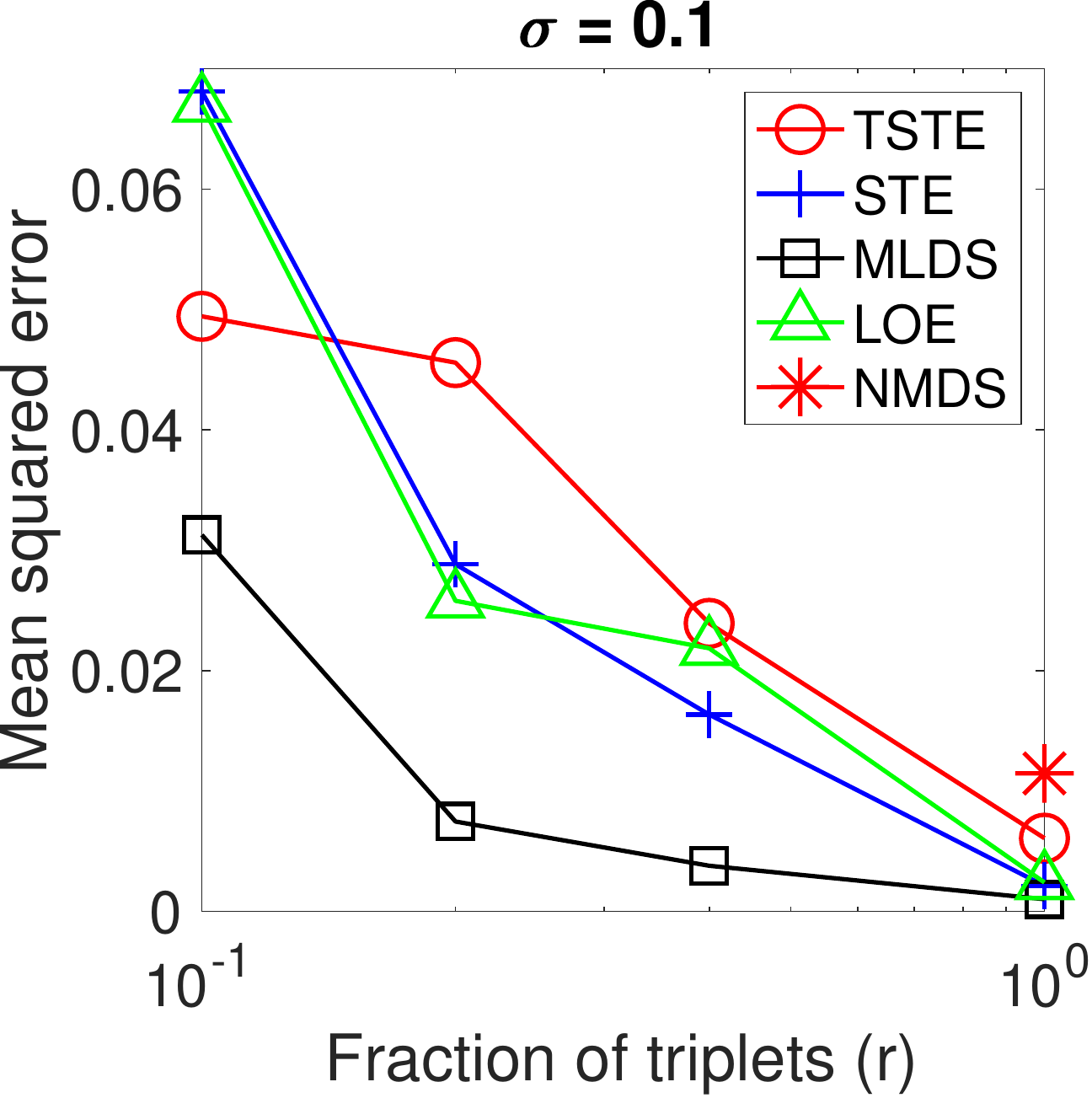}
}
\subfigure[\hspace{-5mm}]{
    \centering%
    \includegraphics[width = .185\linewidth]{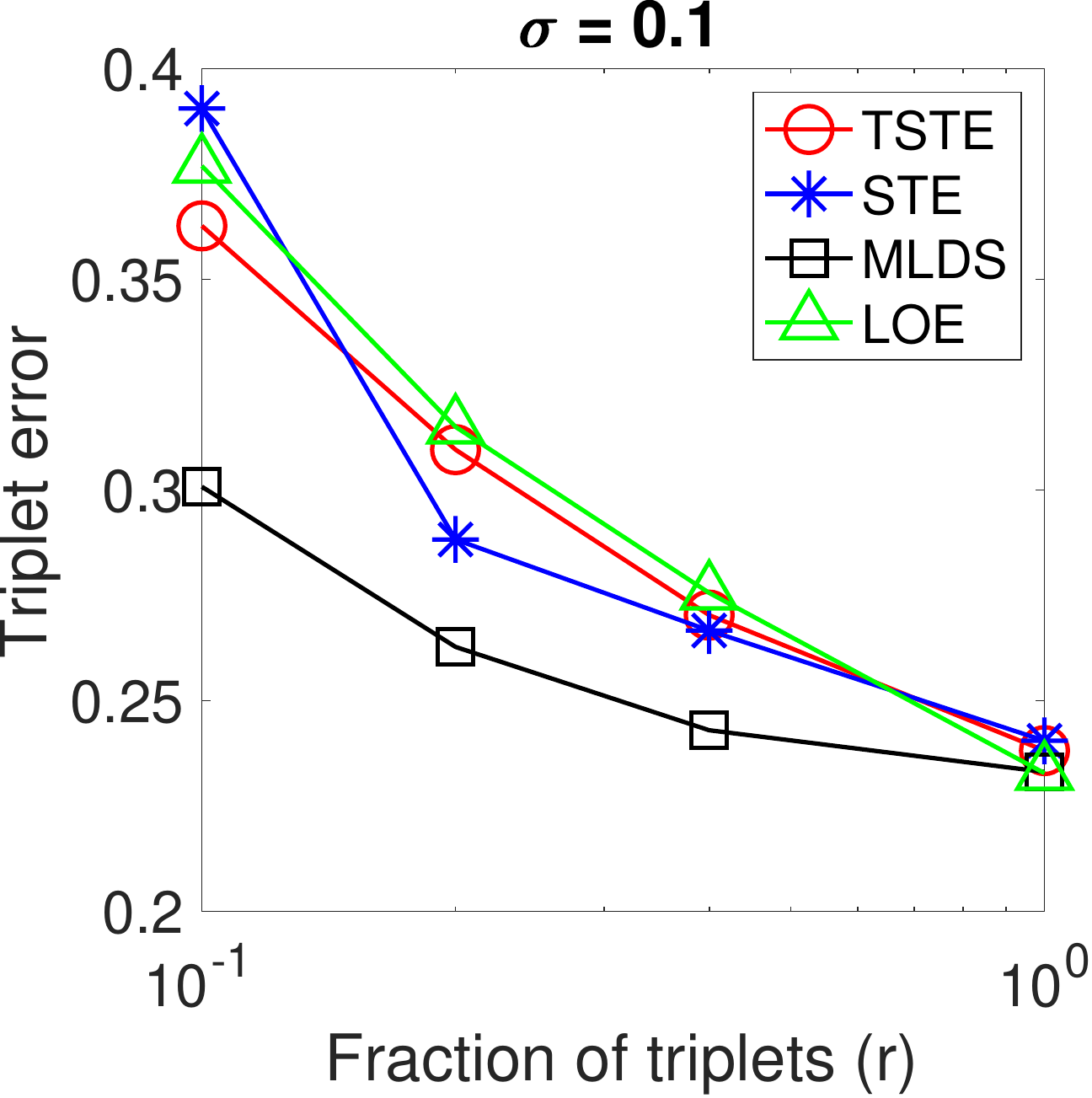}
}
\caption{\label{fig:Simsf1} Comparison of various ordinal embedding methods (LOE, STE, t-STE) against the traditional embedding methods in psychophysics (MLDS and NMDS) for a monotonic one-dimensional perceptual function (Sigmoid). (a) The true perceptual function ($y$). (b) Ten embedding results ($\hat y$) of the MLDS method for a fixed value of standard deviation $\sigma$ and triplet fraction $r$. (c) Ten embedding results ($\hat y$) of the STE method for a fixed value of standard deviation $\sigma$ and triplet fraction $r$. (d) The average MSE of embedding methods. (e) The average triplet error of embedding methods.}\end{figure}

{\bf Evaluating the results:} We consider two approaches to evaluate the performance of the various methods: 
\begin{enumerate}
    \item Mean-squared-error (MSE): For one-dimensional perceptual spaces, we can compute the mean-squared-error (MSE) between the estimated scales $\hat y$ and the true perceptual function values $y$. Since the embedding result is unique only up to similarity transformations (scaling, rotation and translation), we need two steps of \textbf{normalization} before computing the MSE. First, we transform the output of embedding to be in the range of $(0,1)$ as our scaling functions are defined in this range. More precisely, we shift (translate) the minimum value to zero and divide all the values by the maximum. Secondly, If we get the output $\hat{y}$ as a result of embedding, this answer is not unique. More precisely, $-\hat{y}$ can also be considered as an answer without violating any triplet of the input set. Therefore, we choose between $\hat{y}$ and $-\hat{y}$, the output which shows a better performance with respect to the MSE. In this way we choose the best rotation of the output.

    \item Triplet error: The MSE criterion is cumbersome to compute in multivariate scenarios, because we have to take into account all possible rotations of the embeddings. Moreover, in real-world scenarios, MSE cannot be computed anyways because the underlying ground truth is unknown. 
    As an alternative, we propose to evaluate the quality of an embedding by its ability to predict the answers to (potentially new) triplet questions.  To this end, we compute a quantity called the \textbf{triplet error}. Given an embedding $\hat y_1, ..., \hat y_n$ and a validation set $T^\prime$ of triplets, the triplet error of the embedding with respect to $T^\prime$ is defined as 

\begin{align} \label{eq:TripletError}
\text{triplet error} = \frac{1}{\vert T^\prime \vert}\sum_{t=(i,j,k)\in T^\prime}{\mathbbm{1}\left\{R_t \cdot \sgn(\Vert \hat{y}_i - \hat{y}_j \Vert^2 - \Vert \hat{y}_i - \hat{y}_k \Vert^2) = 1\right\}},
\end{align}

where the characteristic function $\mathbbm{1}$ takes the value 1 if the  expression in the curly parenthesis is true (that is, if the estimated embedding is not consistent with the new triplet $t$),  and it takes the value 0 otherwise. 

In words, the triplet error counts how many of the triplets in $T^\prime$ are not consistently represented by the given embedding.  In practice, we are typically provided with only \textit{one} set of answered triplets; this set then has to be used both for constructing the embedding and for evaluating its quality. 

The first way is to set $T^\prime=T$, meaning that we use the same set of triplets to construct the embedding and to measure its quality. In a second way, we perform $k$-fold cross-validation to avoid overfitting: We partition the set of input triplets $T$ into $k$ non-intersecting subsets (``folds''). We perform the embedding and the evaluation $k$ times. In each iteration we pick one of the folds as the validation set ($T^\prime$) and the rest of the folds as the training set (the input to the embedding algorithm). The final triplet error is the average over the triplet errors of the $k$ validation sets. Throughout the rest of the paper, we refer to the latter approach as \textbf{cross-validated triplet error}, while the first approach is simply called the \textbf{triplet error}.
\end{enumerate}

\subsection{One-dimensional perceptual space}
\subsubsection{Simulations with monotonic scales}

\begin{figure}
\centering
\subfigure[\hspace{-5mm}]{
    \centering
    \includegraphics[width = .175\linewidth]{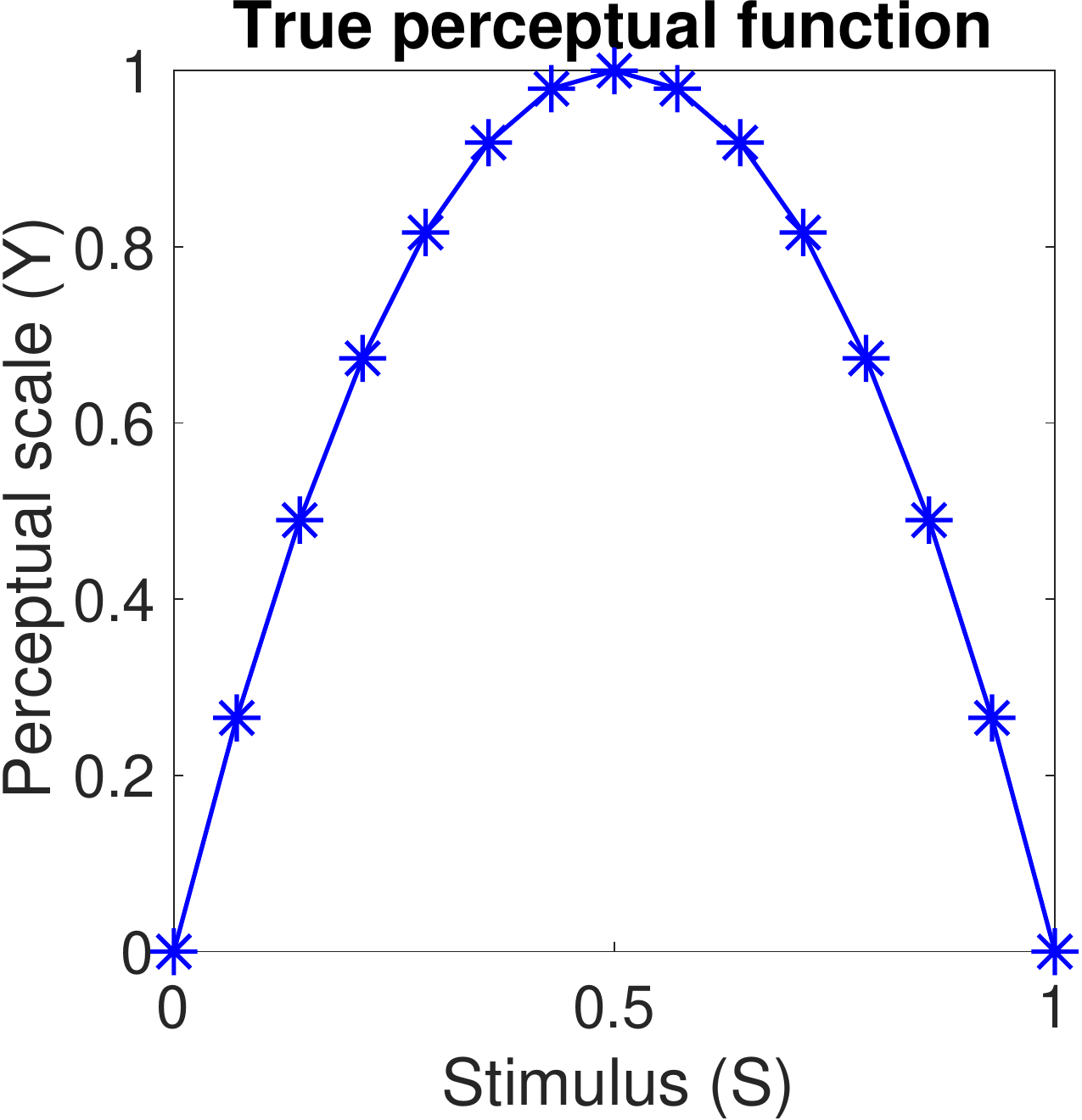}
}
\subfigure[\hspace{-5mm}]{
    \centering
    \includegraphics[width = .175\linewidth]{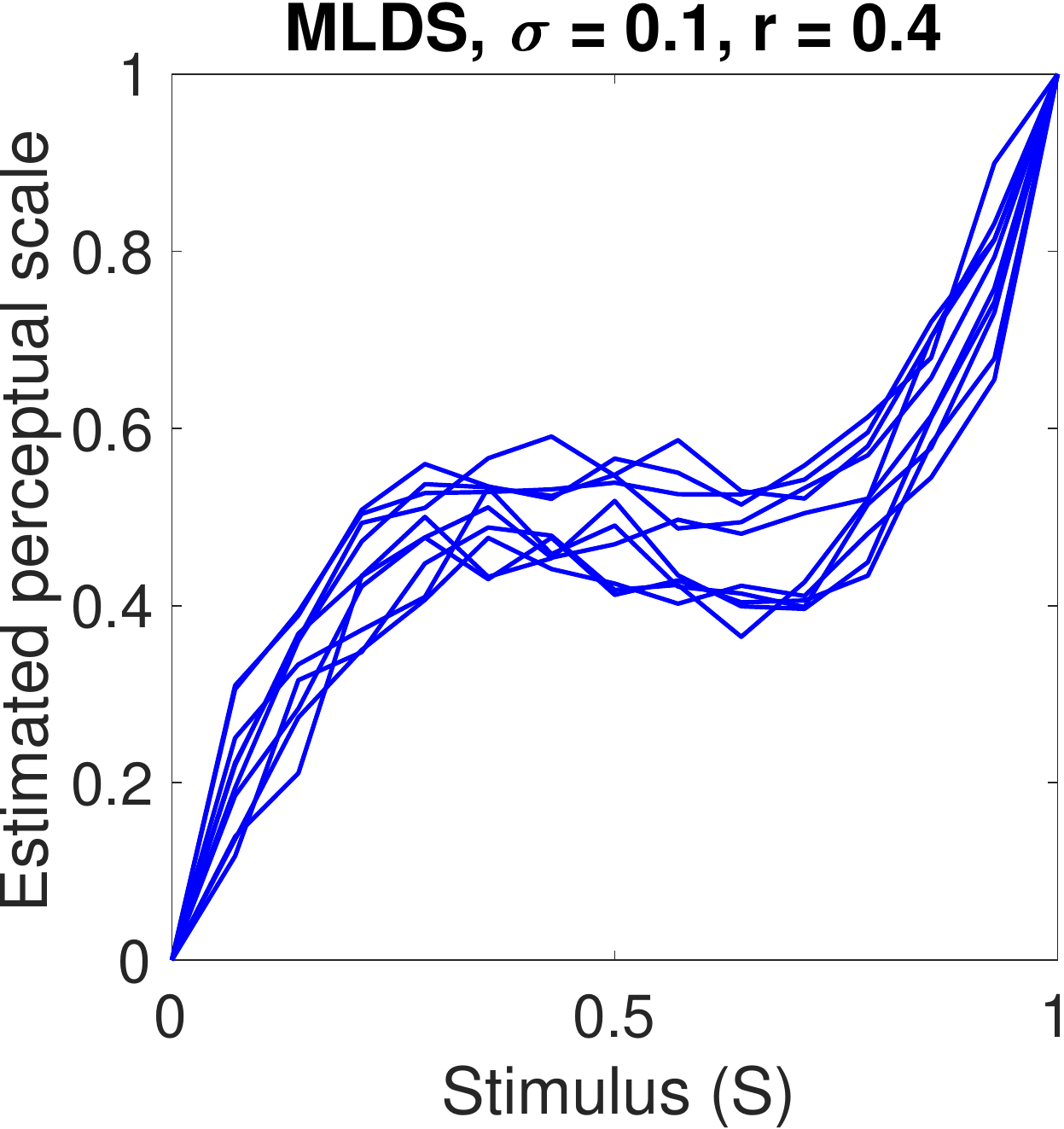}
}
\subfigure[\hspace{-5mm}]{
    \centering%
    \includegraphics[width = .175\linewidth]{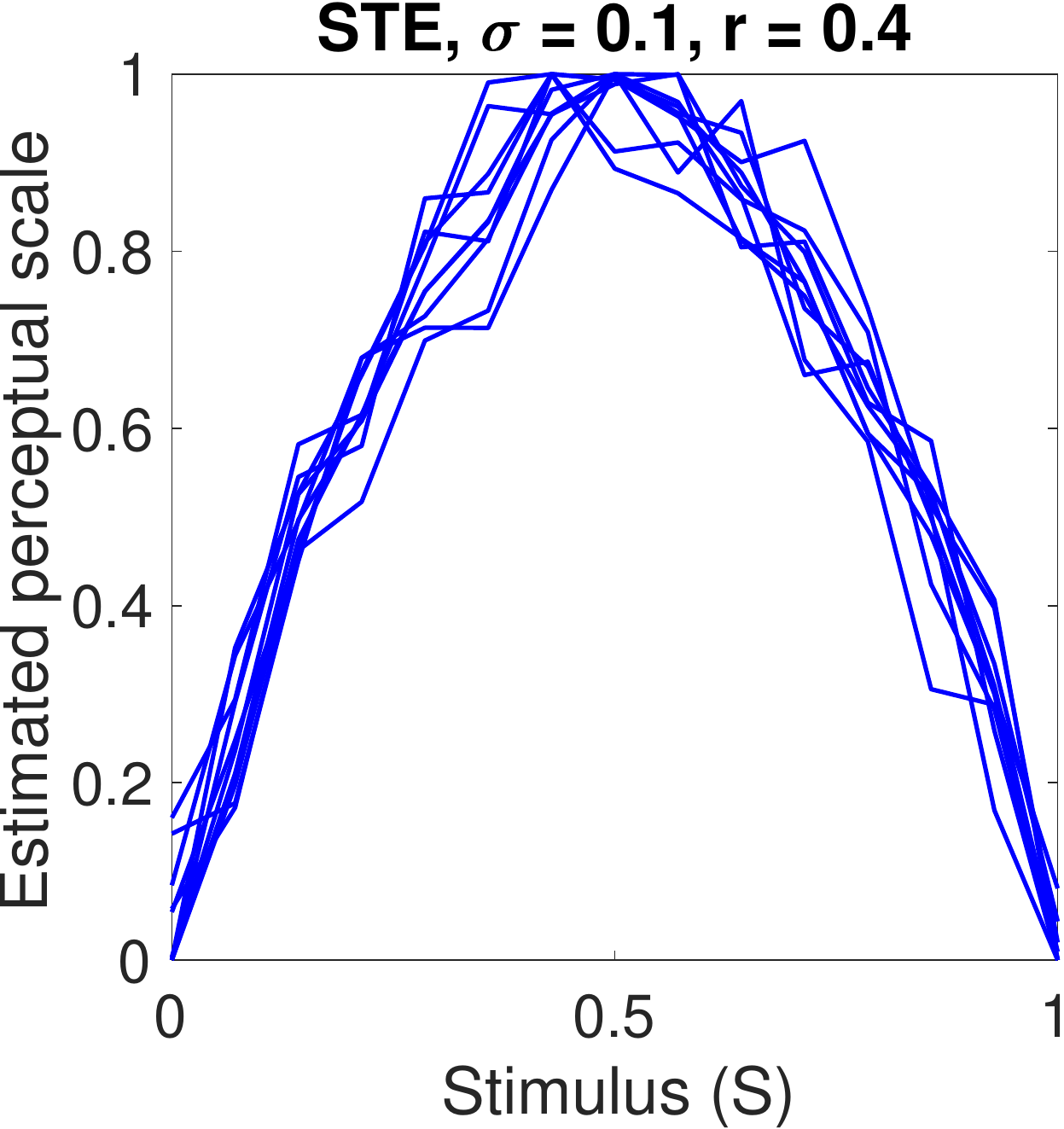}
}
\subfigure[\hspace{-5mm}]{
    \centering%
    \includegraphics[width = .185\linewidth]{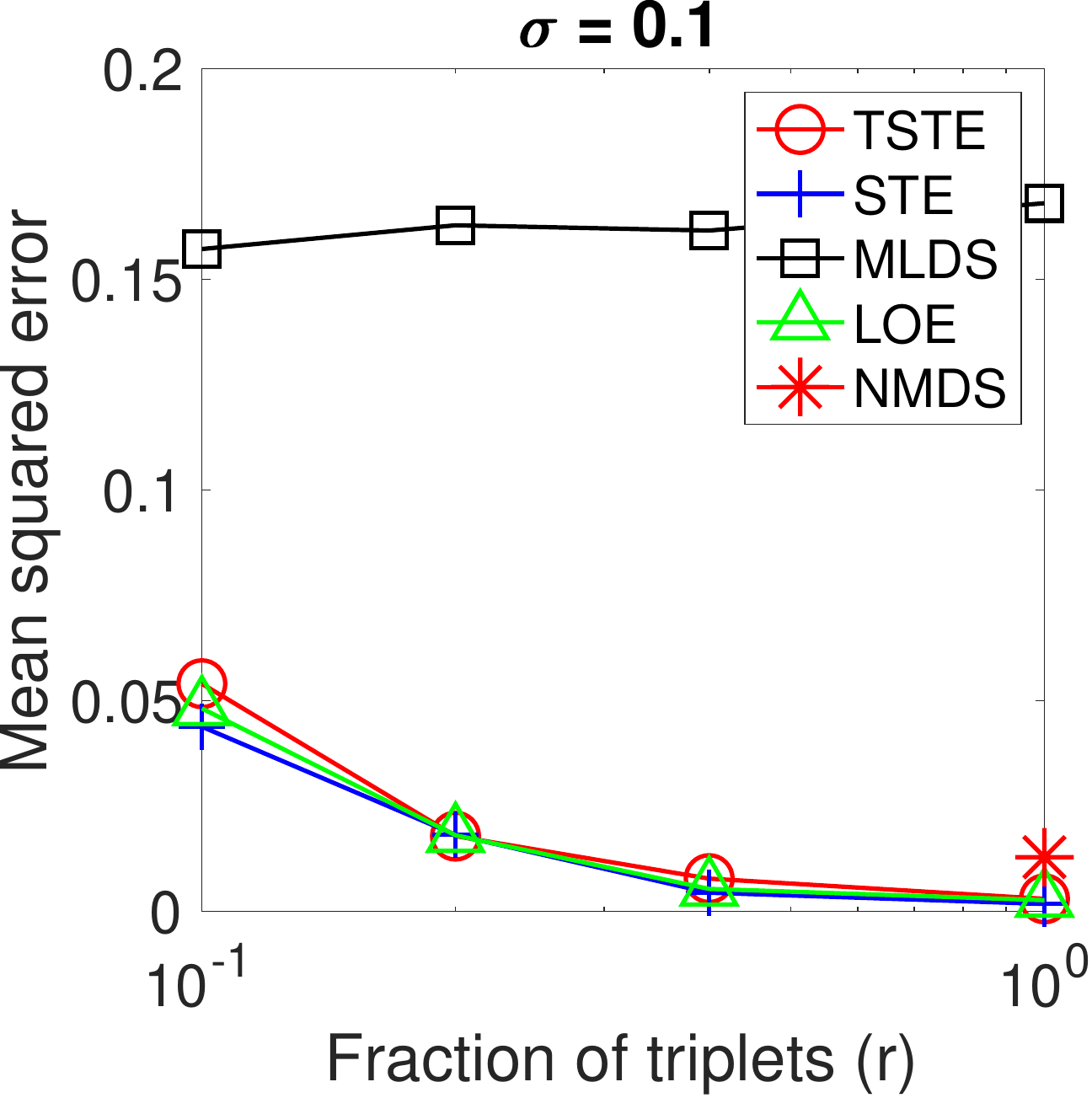}
}
\subfigure[\hspace{-3mm}]{
    \centering%
    \includegraphics[width = .185\linewidth]{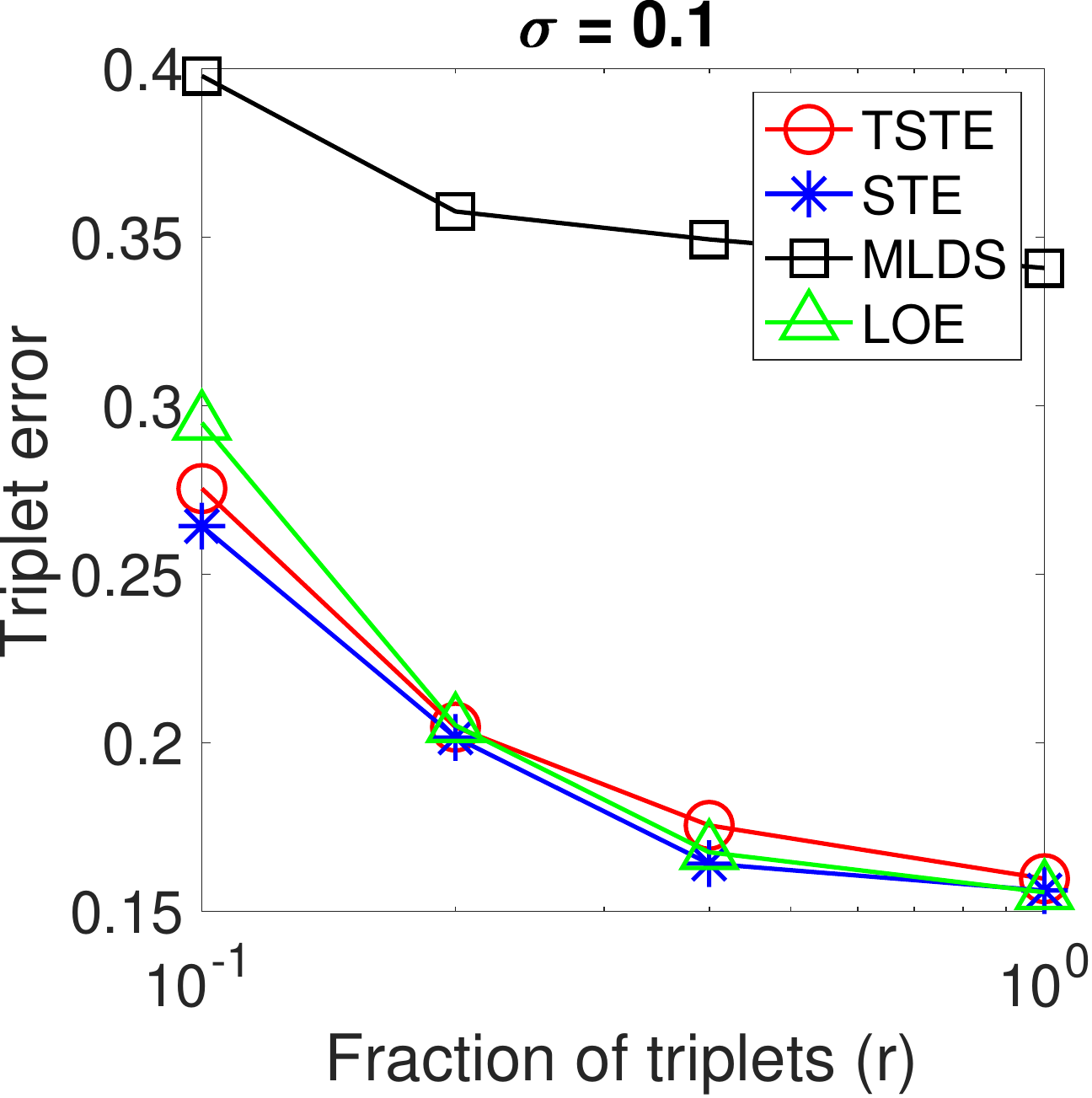}
}
\caption{\label{fig:Simsf4} Comparison of various ordinal embedding methods (LOE, STE, t-STE) against the traditional embedding methods in psychophysics (MLDS and NMDS), for a non-monotonic one-dimensional perceptual function (second degree polynomial). (a) The true perceptual function ($y$). (b) Ten embedding results ($\hat y$) of the MLDS method for a fixed value of the standard deviation $\sigma$ and triplet fraction $r$. (c) Ten embedding results ($\hat y$) of the STE method for a fixed value of standard deviation $\sigma$ and triplet fraction $r$. (d) The average MSE of embedding methods. (e) The average triplet error of embedding methods.}\end{figure}

Our first simulation involves a typical monotonic function as it occurs in many psychophysics experiments. The true perceptual function $f$ (a Sigmoid function) is shown in Figure~\ref{fig:Simsf1} (a). Figure~\ref{fig:Simsf1} (b) and (c) show the output embedding of the MLDS and STE algorithms for 10 iterations, respectively. The other ordinal embedding methods have a similar performance and the output embeddings are reported in the supplementary material. The average (over 10 runs) MSE and triplet errors of various embedding algorithm are depicted in Figure~\ref{fig:Simsf4} (d) and (e),  respectively. 

In both error measures the MLDS method performs better than the ordinal embedding algorithms. The obvious reason is that MLDS makes a strong model assumption, namely the monotonicity of the scaling function, and this assumption is satisfied in this example. Hence it has an inductive advantage that pays off.  The ordinal embedding algorithms also show an acceptable performance, however. In particular, when we provide more triplet answers ($r=1$) the average errors of MLDS and the ordinal embedding algorithm tends to be the same.
More detailed results regarding this simulation, including the four ordinal embedding outputs and the performance of algorithm with other values of $\sigma$, can be found in the supplementary material, see Figure~\ref{fig:simsf1Extra}. We also examine another monotonic function in  Figure~\ref{fig:simsf2Extra} of the supplementary material, and the results are consistent with the ones presented here. 

\subsubsection{Simulations with non-monotonic scales}

We now perform the same experiment with a non-monotonic function: a second-degree polynomial function is chosen as the true perceptual function $f$; see Figure~\ref{fig:Simsf4} (a). Figure~\ref{fig:Simsf4} (b) and (c) show the output embedding of the MLDS and STE algorithms for 10 iterations respectively (the embeddings produced by LOE and t-STE are quite similar to the STE, see supplementary material). The average (over 10 runs) MSE and triplet error of various embedding algorithm are depicted in Figure~\ref{fig:Simsf4} (d) and (e) respectively. 

The function shapes depicted in Figure~\ref{fig:Simsf4} (b) show the poor performance of the MLDS method for the non-monotonic function: MLDS tries to fit the most consistent monotonic function to the input triplets; but as the ground truth is far from being monotonic, the MLDS result is far off. The average MSE and triplet errors are also significantly larger for the MLDS. In contrast, the ordinal embedding algorithms can correctly estimate the true function shapes. Note that the ordinal embedding methods are capable of discovering a much broader range of scaling functions (non-monotonic scales) with the same number of triplets as we used for the monotonic scales.

Similar to the monotonic functions we report the full details of the simulation in supplementary material; see Figure~\ref{fig:simsf4Extra}. We also perform the simulation on a Sinosoid function. The results are quite similar to the second-degree polynomial function and are demonstrated in the Figure~\ref{fig:simsf3Extra} of the supplementary material. 

\begin{figure}
\centering
\subfigure[\hspace{-5mm}]{
    \centering
    \includegraphics[width = .215\linewidth]{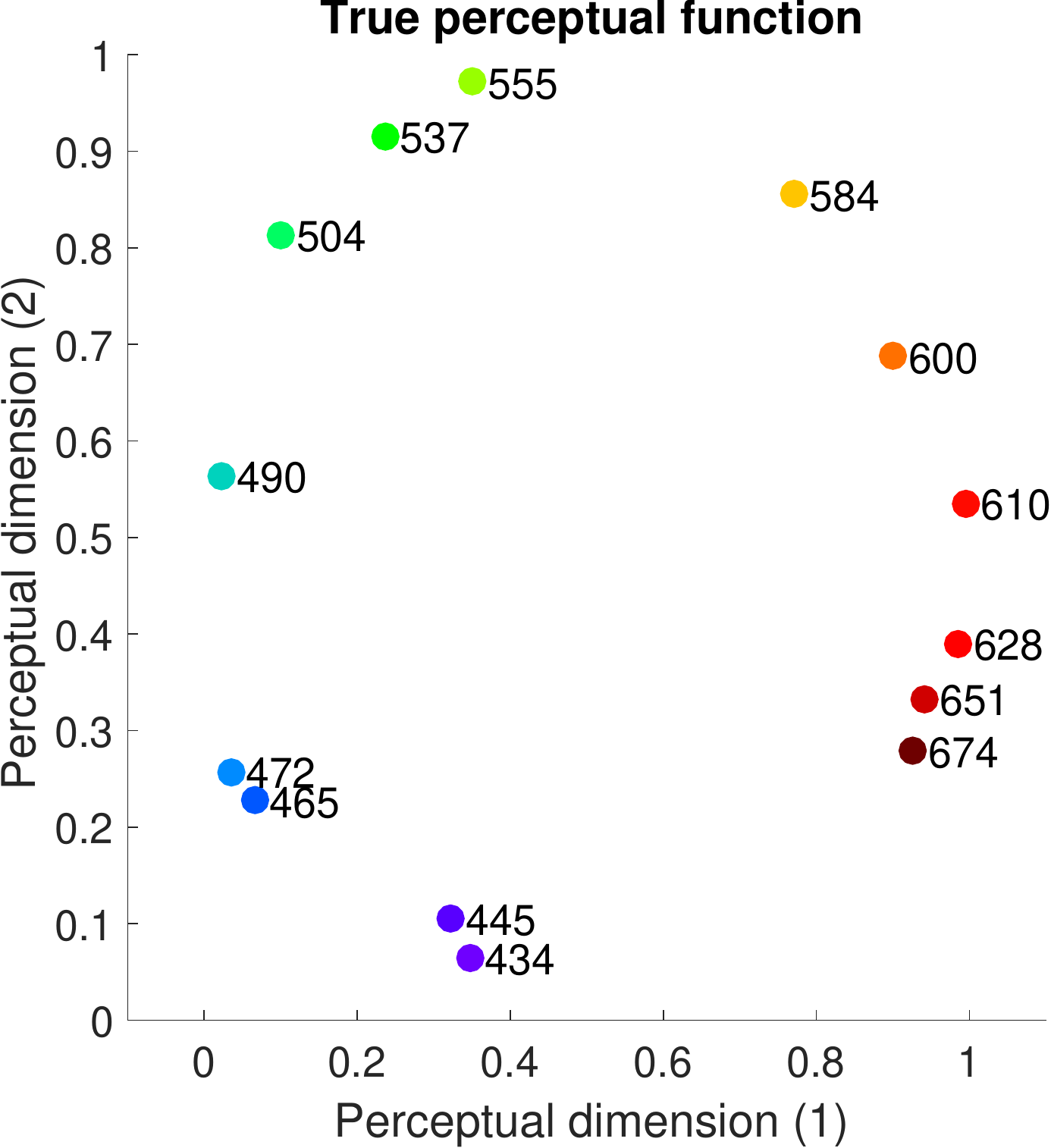}
}
\subfigure[\hspace{-5mm}]{
    \centering
    \includegraphics[width = .22\linewidth]{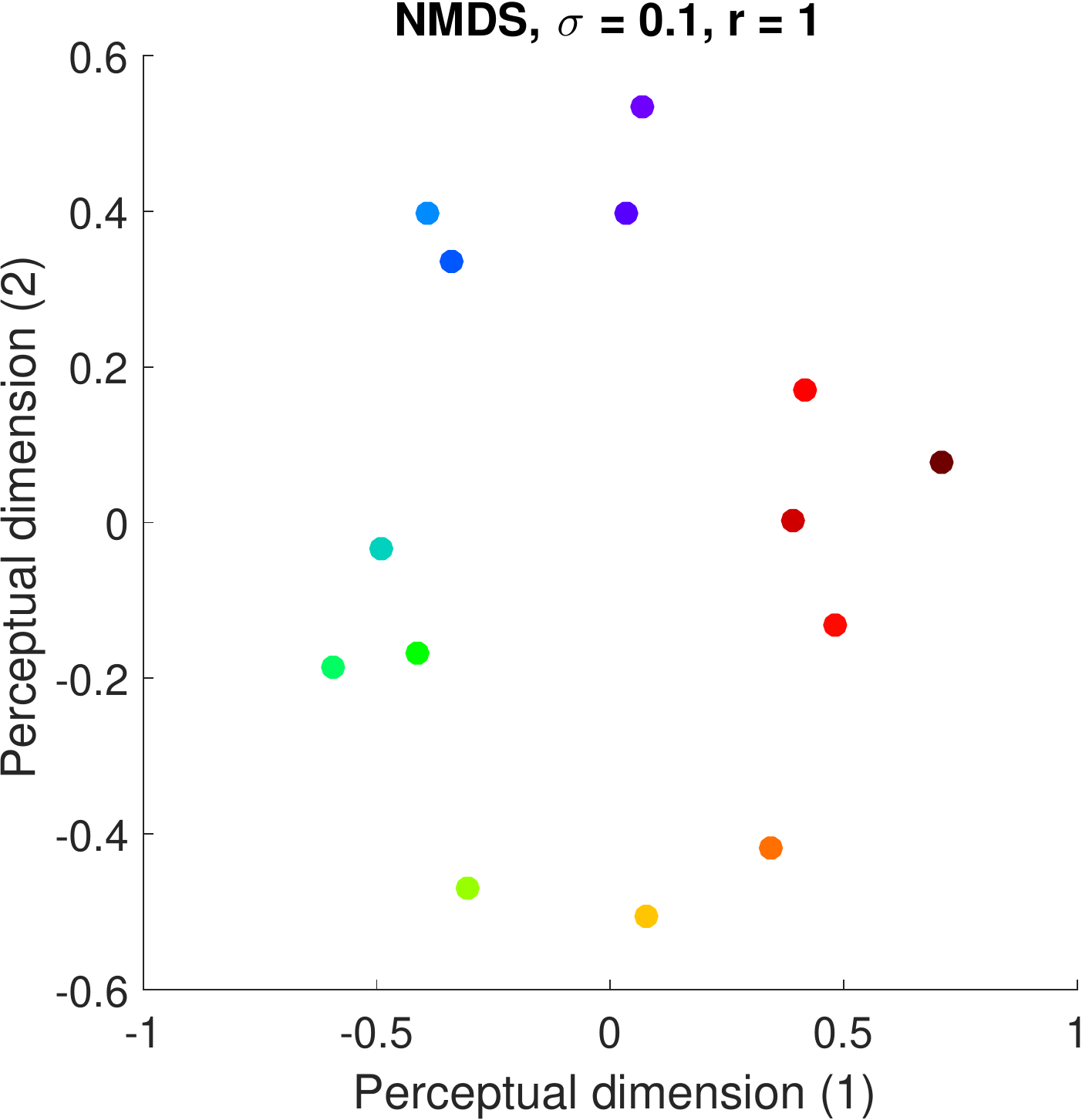}
}
\subfigure[\hspace{-5mm}]{
    \centering%
    \includegraphics[width = .22\linewidth]{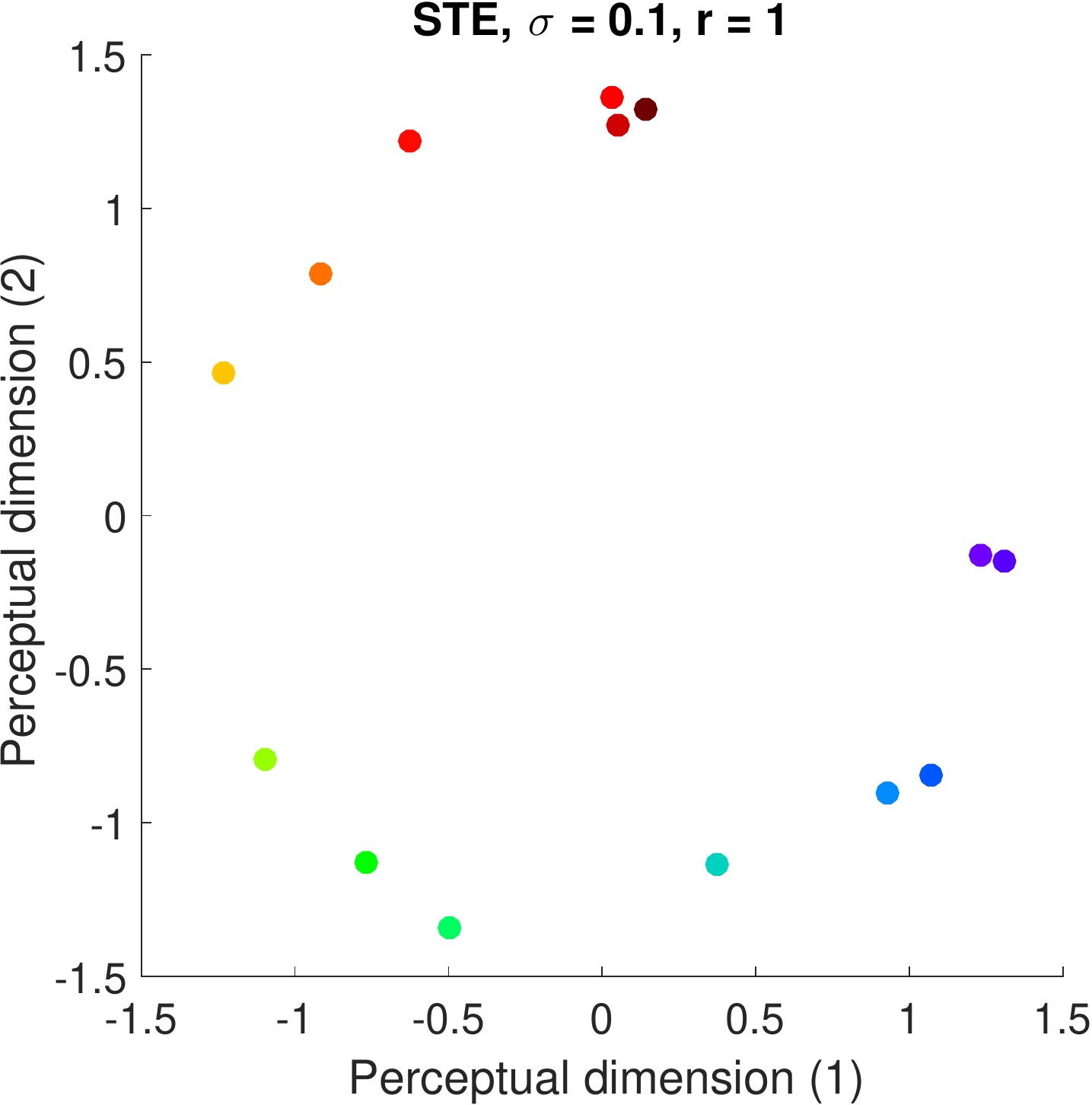}
}
\subfigure[\hspace{-5mm}]{
    \centering%
    \includegraphics[width = .23\linewidth]{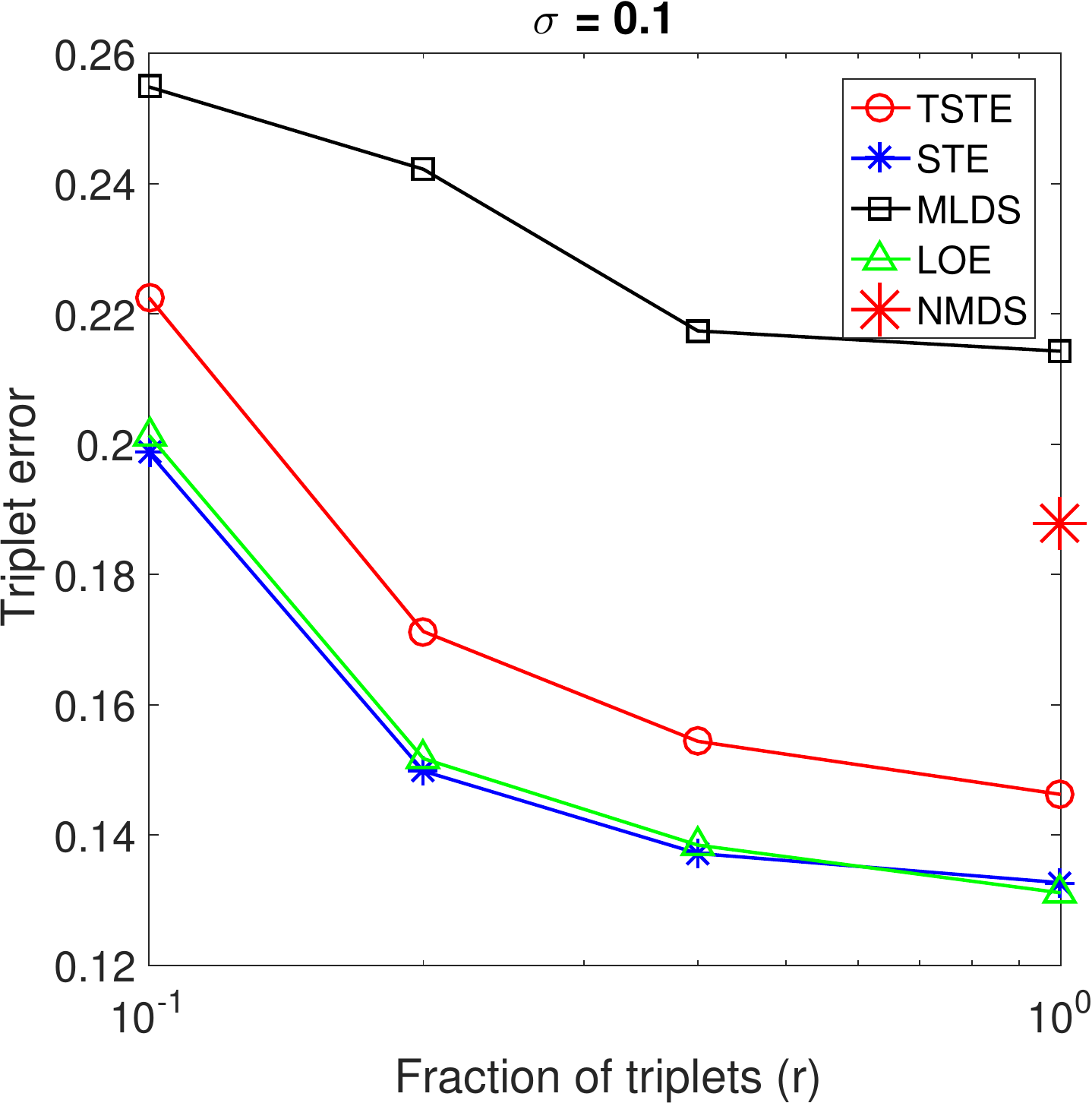}
}
\caption{\label{fig:SimsColor} Comparison of the ordinal embedding methods (MLDS, STE and TSTE) against the traditional NMDS method of psychophysics for the two-dimensional color perception function. (a) The true perceptual function in two dimensions. The stimulus value, color wavelength, is written beside each color. The two-dimensional vector space represents the perceptual space. (b) The embedding result of the NMDS method depicted in two dimensions for a fixed value of standard deviation $\sigma$ and triplet fraction $r$. (c) The embedding result of the STE method depicted in two dimensions for a fixed value of standard deviation $\sigma$ and triplet fraction $r$. (d) The average triplet error of various ordinal embedding methods in comparison with the NMDS method.}\end{figure}
\subsection{Multi-dimensional perceptual space}

So far, we considered simulations in which the perception could be represented in a one-dimensional Euclidean space. However, in some cases such as the examples of color and pitch perception in Figure~\ref{fig:MultiDimExamples}, more than one dimension is required to represent the perception. Here, we perform a simulation with a function mapping from one-dimensional stimulus space into a two-dimensional perceptual space. 

In order to construct a realistic psychometric function $f$, we use the color similarity data\footnote{\url{https://faculty.sites.uci.edu/mdlee/similarity-data/}} presented in~\citet{ekman1954dimensions}. We first construct a two-dimensional  embedding using NMDS; see Figure~\ref{fig:SimsColor}~(a). In the following, this embedding will be considered our ground truth, which will then be used to generate further data (let us stress: we do not argue that this embedding is ``correct'' in any way; we just use it as a ground truth to generate further simulations). Figure~\ref{fig:SimsColor} (a) shows this embedding.  The wavelength of each color is also denoted beside the color. The X-Y axes of the plot correspond to the two perceptual dimensions of the color. 

To generate noisy triplets from our ground truth, we essentially proceed as before: we rescale the stimulus sizes (wavelengths) to the range of $S \in (0,1)$ to be consistent with the underlying model we defined earlier. We define the true ground truth function $f$ by our ground truth embedding, which is the true two-dimensional representation $y_i = f(S_i)$ of a stimulus $S_i$ is given by the values in Figure~\ref{fig:SimsColor}. Now we generate noisy versions $\tilde y_i$ of the perceptual scale functions, random subsets of triplets and noisy answers to triplet questions as described in the beginning of this section, and use the various algorithms to compute an estimated embedding $\hat y_i$ of all the stimuli. 
We fix the embedding dimension to $d=2$ for the following embedding methods: NMDS, LOE, STE and t-STE. However, MLDS is only capable of embedding in one dimension. Thus, we perform MLDS with $d=1$. Figure~\ref{fig:SimsColor} (b) and (c) show the two-dimensional embedding output of the NMDS and STE algorithms respectively. The embedding are shown for the parameter values $\sigma=0.1$ and $r=1$. The average triplet error of various embedding methods is shown in Figure~\ref{fig:SimsColor} (d) for the parameter value $\sigma=0.1$.

The comparison of Figure~\ref{fig:SimsColor}~(b) and (c) reveals the different  performances of NMDS and ordinal embedding methods in the presence of noise. NMDS is known to be quite vulnerable to noise, and this can be seen from the figures. While STE produces a circle of colors fairly similar to the true perceptual function, the colors are somewhat mixed up in the NMDS embedding. The triplet error also shows that ordinal embedding algorithms outperform the NMDS method significantly---even if we have only half or less of the triplets available! Finally, and as expected by design, MLDS cannot produce an embedding in two dimensions. When evaluated on its one-dimensional embedding, it unsurprisingly produces triplet errors that is much larger than the one of the two-dimensional embedding methods. More details regarding this experiment can be found in the supplementary material; see Figure~\ref{fig:simsf5Extra}.

\begin{figure}
\centering
\begin{subfigure}
    \centering
    \subfiguretopcaptrue
    \includegraphics[width = .55\linewidth]{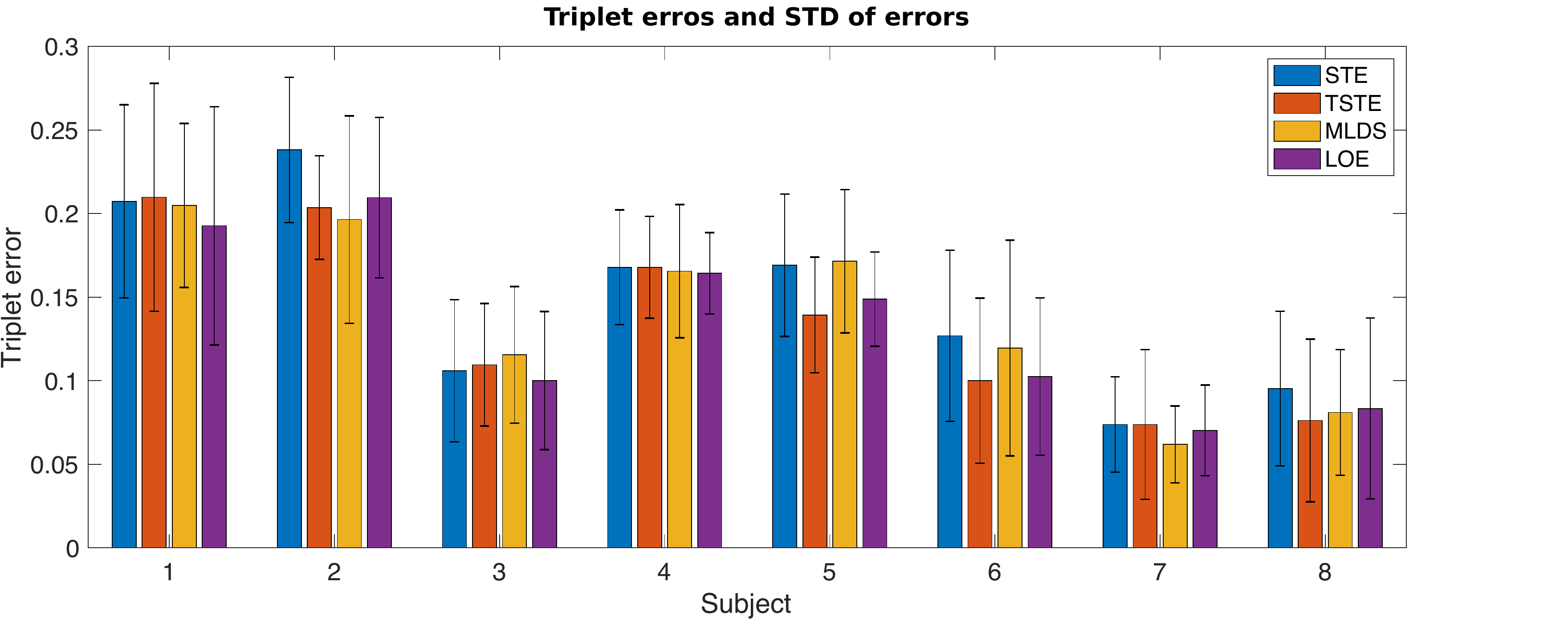}
\end{subfigure}\\
\begin{subfigure}
    \centering
    \includegraphics[width = .85\linewidth]{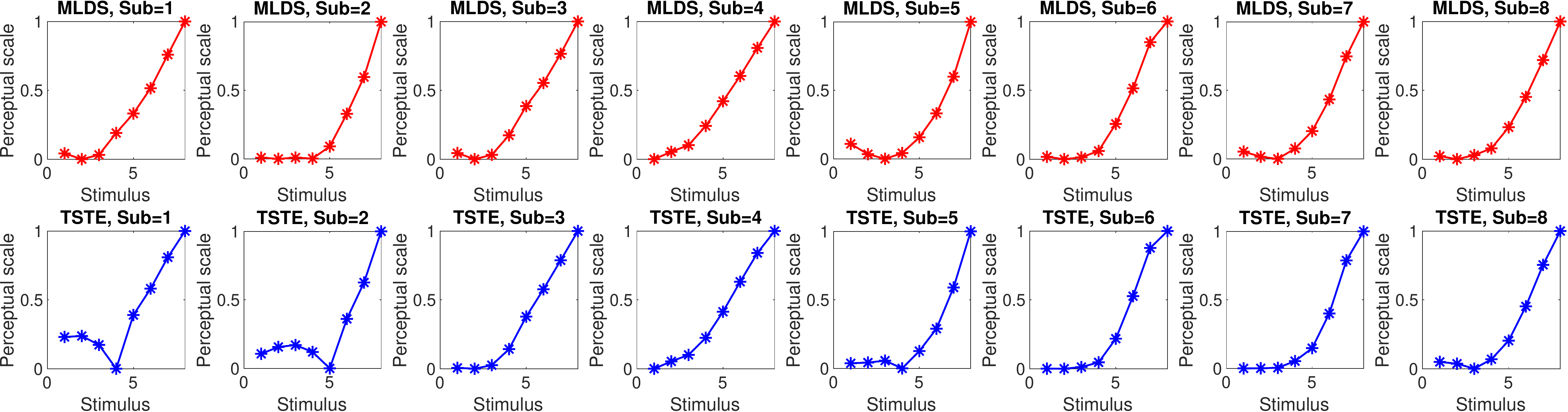}
    \vspace{5mm}
\end{subfigure}
\caption{\label{fig:SlantRes} (Top) Average and standard deviation of cross-validated triplet error for 8 subjects of the slant-from-texture experiment. Each group of bar shows the error for one subject, as each bar in the group corresponds to one of the embedding methods shown with different colors. (Bottom) The embedding outputs for 8 subjects with two embedding methods: MLDS and t-STE. The MLDS method is depicted at the top row while the t-STE is shown at the bottom.}
\end{figure}

\section{Experiments}

In this section, we apply the comparison-based approach and ordinal embedding methods to two real experiments in visual perception: the slant-from-texture experiment that we have already mentioned above, and a more complex ``Eidolon'' experiment. 

\subsubsection{Slant-from-texture experiment}
\label{sec:slantExperiment}

This experiment intends to find the functional relation between the perceived angle of the slant with a dotted surface and the actual physical degree of slant. The dataset has originally been generated in~\citep{Aguilar2017} (see the paper for more information on the experimental settings). Figure~\ref{fig:slantExample}~(top) shows the eight stimuli used in this experiment. The degree of slant is varied from 0 to 70 degrees in steps of 10 degrees, making 8 stimulus levels. Then participants had to answer triplet comparisons. As the experiment has initially been performed with the assumption of a monotonic relation of slant degree and the perception, for each combination of three stimuli $S_i<S_j<S_k$ (three degrees of tilting) only one triplet question has been asked:  whether $\delta(S_j,S_i) < \delta(S_j,S_k)$. With 8 levels of the stimulus, this results in  ${8 \choose 3}=56$ possible triplet questions. Eight subjects participated in the study. Each subject has answered all 56 triplet questions many times, in order to reduce the effect of noisy responses. Subjects $\{1,6,8\}$ have answered 420 triplet question in total, while the other subjects answered 840.

Since the ground truth embedding is unknown, we can only rely on the triplet error for evaluation of the embeddings. To avoid overfitting we use 10-fold cross-validation to compute the \textit{cross-validated triplet error} (see the definition in the simulation setup). Figure~\ref{fig:SlantRes} (top) shows the average and standard deviation of the cross-validated triplet error for 8 subjects and four embedding methods, including: MLDS, STE, t-STE and LOE. All ordinal embedding algorithm have very similar performance to MLDS---thus in this real-world example the advantage of MLDS over the ordinal embedding algorithms see in Figure~\ref{fig:Simsf1} appears to have vanished.

In addition to the triplet error, we also show the embedding outputs of MLDS and t-STE for 8 subjects in Figure~\ref{fig:SlantRes}~(bottom). Note that these plots are generated with the full set of triplets, not only the training folds that are used to evaluate the triplet error. The resulting functions are similar, both across the two methods and across the participants. However, it is curious to see  that the t-STE embeddings produced for observers 1 and 2 are not monotonous. This is an effect that could not happen for MLDS, because MLDS always outputs monotonous functions. On the other hand, the triplet errors in both cases are comparable. It would now be interesting to further investigate the perceptions of observers 1 and 2 in more details; however this would require more lab experiments involving the two observers, which is beyond the scope of this paper. At this point we can only stress that ordinal embedding methods at least have the potential to discover interesting, non-standard patterns in perceptual data that might be overlooked by MLDS.

\subsection{The eidolon experiment}
\begin{figure}[t]
\newcommand{\myimagewidth}{2.5cm}
    \centering
    \begin{minipage}{0.3\textwidth}
    \centering
    \includegraphics[width=\myimagewidth]{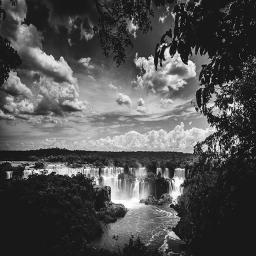}
    \end{minipage}
    \begin{minipage}{0.44\textwidth}
    \centering
    \includegraphics[width=\myimagewidth]{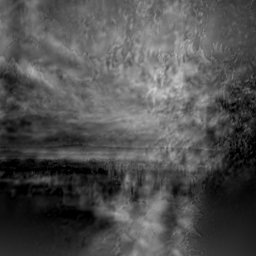} \\
    \includegraphics[width=\myimagewidth]{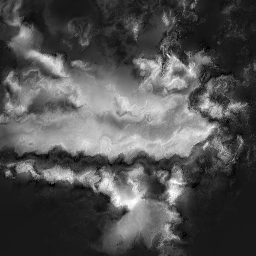}
    \includegraphics[width=\myimagewidth]{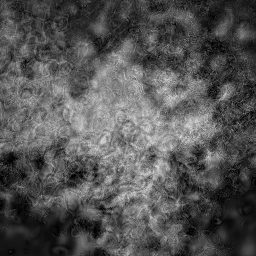}
    \end{minipage}
    \caption{\label{sec:eidolonExperiment}Left: The original image from our Eidolon experiment. Right: An example triplet question --- \textit{``Which of the bottom two images is more similar to the top image?''}}
    \label{fig:compare_methods_synthetic_eidolon_example}
\end{figure}

Our final setup concerns the comparison of images. To generate images we use the Eidolon Factory by \citet{KoeEtAl17}---more specifically, its \mbox{partially\_coherent\_disarray()} function. In this toolbox, a given basis image can be  distorted systematically using three different parameters called \reach, \grain{} and \coherence. An Eidolon of a basis image corresponds to a parametrically altered version of this image. Reach controls the strength of a distortion (the higher the value, the stronger the amplification), grain modifies how fine-grainedness of the distortion (low values correspond to `highly fine-grained'), whereas a parameter value close to 1.0 for coherence indicates that ``local image structure [is retained] even when the global image structure is destroyed'' \citep{KoeEtAl17}. From a perceptual point of view we want to know which and to what degree the image modifications influence the percept. Starting with a black and white image of a natural landscape as basis image (see \fig{fig:compare_methods_synthetic_eidolon_example}, left), we generate 100 altered images, using reach and grain in $\{5, 12, 26, 61, 128\}$ and coherence in $\{0.0, 0.33, 0.67, 1.0\}$. All possible combinations of these parameter values result in $5 \cdot 5 \cdot 4 = 100$ different images.

\paragraph{Lab experiment setup:} In our lab, we asked three participants aged 19 to 25 to answer triplet questions, see \fig{fig:compare_methods_synthetic_eidolon_example} (right) for an example question. For this purpose, participants use a standard computer mouse to click on the one of the two bottom images that they deemed more similar to the top image. Stimuli were presented on a $1920 \times 1200$ pixels ($484 \times 302$ mm) VIEWPixx LCD monitor (VPixx Technologies, Saint-Bruno, Canada) at a refresh rate of 120~Hz in an otherwise dark chamber. Viewing distance was 100 cm, corresponding to $3.66 \times 3.66$ degrees of visual angle for a single $256 \times 256$ pixels image. The surround of the screen was set to a grey value of $0.32$ in the $[0, 1]$ range, the mean value of all experimental images. The experiment was programmed in MATLAB (Release 2016a, The MathWorks, Inc., Natick, Massachusetts, United States) using the Psychophysics Toolbox extensions version 3.0.12 \citep{Brainard97, KleEtAl07} along with the iShow library of the Wichmann-lab (\url{http://dx.doi.org/10.5281/zenodo.34217}).

Answers had to be given within $4.5$ seconds after a triplet presentation onset, otherwise the triplet was registered as unanswered and the experiment proceeded to the next triplet (this occurred in only $0.013$\% of all cases and can thus be safely ignored). The experiment was self-paced:  once a participant had answered a question, the next one appeared directly after a short fixation time of $0.3$ seconds during which only a white $20 \times 20$ pixels fixation rectangle at the center of the screen was shown. Before the experiment started all test subjects were given instructions by a lab assistant and performed 100 practice trials to gain familiarity with the task. The set of practice triplets is disjoint from the set of experimental triplets. Participants were free to take a break every 200 triplet questions. They gave their written consent prior to the experiment and were either compensated \euro 10 per hour for their time or gained course credit towards their degree. All test subjects were students and reported normal or corrected-to-normal vision. The experiments were carried out in accordance with the guidelines of the Deutsche Gesellschaft für Psychologie (DGPs). Informed consent was obtained for experimentation by all participants.

\paragraph{Experiment design:} The first step is to design a plausible subset of triplet question. There exist 100 stimuli, giving rise to around $10^6$ possible triplet questions! In contrast to the previous slant-from-texture experiment it is absolutely impossible to ask the whole set of triplet questions---NMDS would thus not be possible. Here the machine learning theory literature comes to aid: it has been proven theoretically that if the embedding dimension is $d$, then of the order of $dn \log n$ triplet questions are sufficient to reconstruct the Euclidean representation of $n$ items up to small error~\citep{JaiJamNow16}. Even though this is just an asymptotic statement and constants are completely ignored, it gives a guideline. 
If we assume that the perceptual embedding dimension is not more than $d \approx 3$ (because three parameters are involved to modify the images), then $dn\log n = 100\cdot3\cdot\log_2(100)\approx 2000$ triplet questions. In order to have sufficient triplets, and considering the training and test split, we hence decided to ask 6000 triplets from each participant of the experiment. The triplets are chosen uniformly at random from the set of possible triplets. The triplets have been asked in three sessions each consisting of 2000 triplets each.
\begin{figure}[t]
\centering
\includegraphics[width = .95\linewidth]{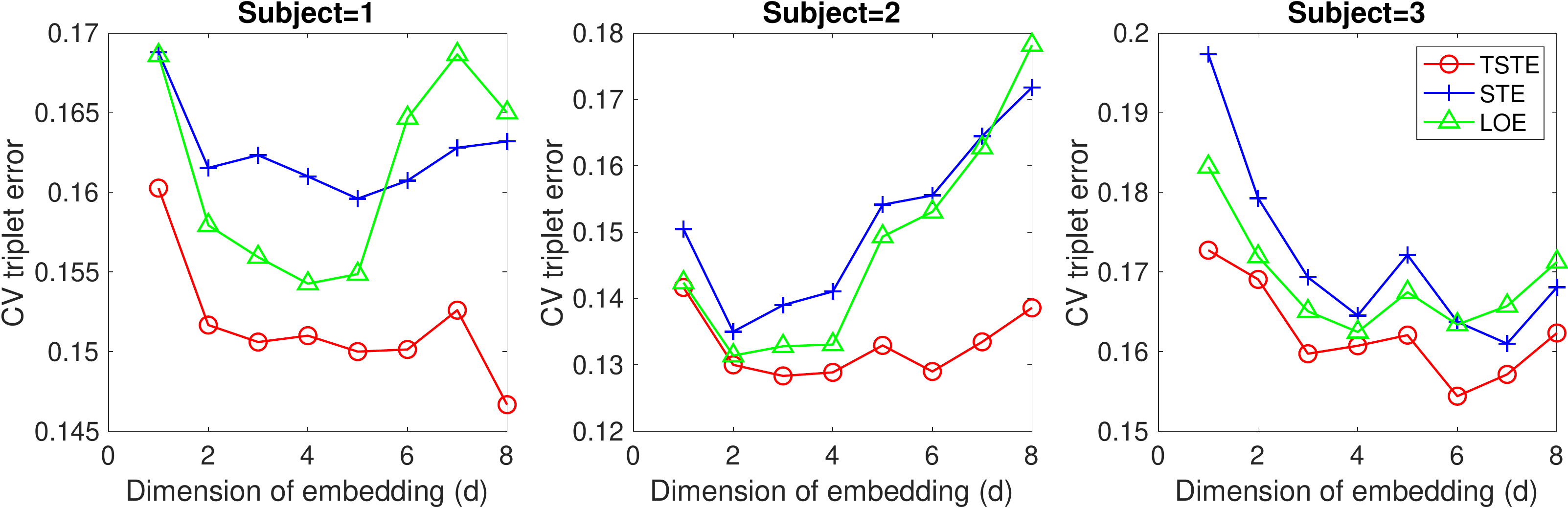}
\caption{\label{fig:EidolonRandRes} Cross-validated triplet error of three embedding methods for three subjects of the Eidolon experiment. Each plot corresponds to one subject and each curve denoted the cross-validated triplet error of one method. The x-axis is the dimension of embedding.}
\end{figure}

Based on the triplet answers, we now run the ordinal embedding algorithms (STE, t-STE, LOE). As the best embedding dimension is not known, we test dimensions in the range $d \in \{1,2,\ldots,8\}$. MLDS method is again performed only in one dimension, as it is not applicable in multi-dimensional cases. We perform 10-fold cross-validation, and the cross-validated triplet error (see Equation~\ref{eq:TripletError}) is reported as the evaluation criterion. 

Figure~\ref{fig:EidolonRandRes} shows the cross-validated triplet error for three subjects with various dimensions and three embedding methods. Each plot corresponds to one subject, while each curve shows the error corresponding to an embedding method. We observe that t-STE consistently outperforms the other methods. The cross-validated triplet error for  MLDS is larger than {0.25} for all three subjects. Thus, MLDS is not comparable to the performance of the best embedding methods, and omitted from the plots---it would be off the scale in each of the panels. For all three subjects, increasing the embedding dimension from one to two definitely improves the embedding error---hence we obviously need more than one dimension to describe the perceptual space. Adding further dimensions in most cases does not really help. It looks as if further investigations, and in particular more participants and a joint analysis over all participants would be necessary to come to a conclusion here if one wanted to know how the parameters of the Eidolon Factory are connected to perception.

The above results show that best embedding method (t-STE) leads to a cross-validated triplet error around $0.15$. How should we know if the error is acceptable, or whether it might be possible to reach a much lower error, for example by collecting more triplets? To answer this question, we would need to know what the error baseline of human participants is. In particular, there might be a proportion of ambiguous triplets for which no obviously ``correct'' answer exists. For example, if we knew that 80\% of the triplet questions had an easy, obviously correct answer, and 20\% of the questions were so hard that the answer was almost random, then the best error rate we could hope for would be around 10\% (on 80\% of the triplets we do not make any error, on 20\% of the triplets we guess randomly, getting about 10\% right and 10\% wrong). 
In case of the Eidolon experiment we do not have any external knowledge about the difficulty of triplets. Thus, we conducted a side experiment. We chose a set of 2000 random triplets and asked these questions three times to each subject (triplets have been shuffled such that subjects did not realize that they are answering the same triplets repeatedly). We now estimate the ``difficulty'' of a triplet by how consistent the repeated answers were: if a subject answers the same triplet question with different answers, we consider it as ``hard'', otherwise as ``easy''. We performed this experiment on three subjects. They show the following percentage of hard triplets: $9.2\%$, $9.8\%$ and $11\%$. Having these answers, we would expect at least $0.10$ triplet error. The cross-validated triplet errors reported in our plots above are pretty close to this value, suggesting that our ordinal embeddings are close to what is achievable.

\section{How to apply ordinal embedding methods in psychophysics}

In this section we would like to present some rules of thumb to make ordinal embedding methods more applicable for a researcher who is unfamiliar with the methods.

\subsection{How many triplets?}

For a set of $n$ stimuli, there exist $3 {n \choose 3}$ many triplet questions --- already for moderate $n$ this is by far too many to ask to a participant of an experiment. However, the good news is that for the ordinal embedding methods a small subset of triplets already contains enough information to accurately reconstruct the true embedding. It has been proven that if the required embedding dimension is $d$, then of the order $\mathcal{O}(dn\log(n))$ triplets are sufficient to reconstruct the true embedding of $n$ items (stimulus levels) up to a small error~\citep{JaiJamNow16}. According to this result, we suggest to start with a subset of size $dn\log(n)$ or $2dn\log(n)$ triplets and perform the ordinal embedding. If the time budget allows, one can still increase the number of triplets and see whether the error improves significantly, but $d n \log n$ should be good baseline. 

\subsection{How to make the subset of triplets?}

Consider a set of $n$ stimuli. At the fist step, one needs to consider the whole set of possible triplets. As we mentioned earlier in simulations, every combination of three items from the stimuli set gives rise to \textbf{three} questions. Therefore, the complete set of possible triplet questions contains $3{n \choose 3}$ triplets. The set of all possible triplet might be very large indeed, thus a small subset of triplets needs to be sub-sampled. A natural question is: Which of the triplet questions among the whole set of possible questions should be chosen? Over the course of many years we have tried many subsampling strategies in our group (Luxburg-lab): based on landmarks, based on active learning, based on estimated confidence values, based on the difficulty of triplet questions, etc. However, in all our experiments the simple strategy of selecting triplets uniformly at random from the set of all possible triplets outperformed all other strategies in terms of triplet error. Hence our general rule of thumb is to apply the straightforward random sub-sampling method.

\subsection{How to evaluate the quality of embedding?}

We reported the MSE in our simulations; however, the true perceptual scale is not available in a real experiment. The general approach that we suggest for the evaluation of ordinal embedding is through the \textbf{cross-validated triplet error} (see Equation~\ref{eq:TripletError})---indeed, we suggest that this may be a good idea for MLDS and NMDS, too. The chosen subset of triplets needs to be partitioned into training and validation sets. The embedding method finds a Euclidean embedding for the perceptual scales, given the training set of triplet as input. We then calculate the cross-validated triplet error on the validation set. This procedure is preferable to the triplet error that is evaluated on the very same set that is used to construct the embedding; the latter can be highly biased and typically underestimates the true triplet error (overfitting). 

\subsection{How to choose the embedding dimension?}

We suggest to run the embedding algorithms in various dimensions, say from 1 to 10, and to finally choose the smallest dimension that shows an acceptable cross-validated triplet error. The formal problem is that increasing the dimension can always produce less triplet error---in the extreme case, it is always possible to embed $n$ items in a space of $d=n$ dimensions without any error. In some cases, it might also be possible to estimate the dimension of the data based on particular distance comparisons \citep{KleLux15}.  

\subsection{Which algorithm, which implementation?} 

Considering the results of the various algorithms on the many tasks, and our experience in running ordinal embedding algorithms for many years, we consider t-STE as our method of choice. The original implementation of the authors is available at~\url{https://lvdmaaten.github.io/ste/Stochastic_Triplet_Embedding.html} implemented in MATLAB. We will also provide a general toolbox containing t-STE and MLDS in R upon the acceptance of the paper.

\section{Discussion}

In this paper, we introduced the ordinal embedding methods as a powerful approach to analyze the triplet comparisons gathered from the method of triads. As opposed to common belief, the ordinal embedding methods require a surprisingly small ($d n \log n$) subset of triplet comparisons to achieve acceptable results. This property makes them preferable to  traditional NMDS, which needs the rank order of all $n^2$ pairwise distances. On the other hand, ordinal embedding methods are capable of embedding in multi-dimensional Euclidean spaces without restrictions on the scaling function. In these situations, they have an advantage over  MLDS, whereas even in one-dimensional, monotonic scaling scenarios they are not much worse than MLDS. Hence ordinal embedding methods such as $t$-STE are promising candidates for ``default'' psychophysical scaling algorithm.

\subsection{Open issues}
As almost always there are a few open issues regarding the use of ordinal embedding methods that we think need to be mentioned and/or addressed in the future.

\textbf{Confidence intervals:} There have been considerable efforts to propose algorithms for the ordinal embedding problem. However, there exist no particular study which provides confidence intervals for the estimated embeddings. Although this issue is not taken very seriously in machine learning, for psychophysics this is an issue of high importance. Some first steps in this direction have been take in~\citet{lohaus19}, but there is definitely much room for improvement. 

\textbf{Interpreting the embedding:} A challenging yet important step is to interpret the embedding results. To make the point clear, consider the Eidolon experiment discussed in the previous section. After gathering a two-dimensional perceptual space and a mapping of stimuli in this space, there are a couple of natural questions arising. 
What does each perceptual dimension mean? How are the perceptual dimensions related to the parameters of the stimulus (in this case reach, coherence and grain)? These are essential questions which can lead to better understanding of human perception. 

\textbf{Conjoint measurement:} In addition to the general scaling problem, we believe that ordinal embedding is a promising candidate to tackle conjoint measurement problems. In a conjoint measurement experiment the sensory stimulus consists of more than one modality. Again we could ask participants to compare triplets of items, and subsequently  apply the ordinal embedding. The approach of using triplet comparisons and ordinal embedding would make much less restrictions than many of the approaches in conjoint measurement, which often rely on independence or additivity assumptions on the modalities.

\section{Acknowledgements}

The authors would like to thank Robert Geirhos and Patricia Rubisch for the programming and running of the Eidolon experiment, Uli Wannek for the help with experimental setup, Guillermo Aguilar for providing the slant-from-texture dataset and fruitful discussions, and Silke Gramer for administrative support. This work has been supported by the German Research Foundation DFG via SFB 936/ Z3, the Institutional Strategy of the University of Tübingen (Deutsche Forschungsgemeinschaft, DFG, ZUK 63), and the  DFG Cluster of Excellence “Machine Learning – New Perspectives for Science”, EXC 2064/1, project number 390727645. We also acknowledge support of the International Max Planck Research School for Intelligent Systems (IMPRS-IS).


\newpage
\section{Supplementary Material}

\subsection{Extended simulation results}
\begin{figure}
\centering
\begin{subfigure}
    \centering
    \includegraphics[width = .32\linewidth]{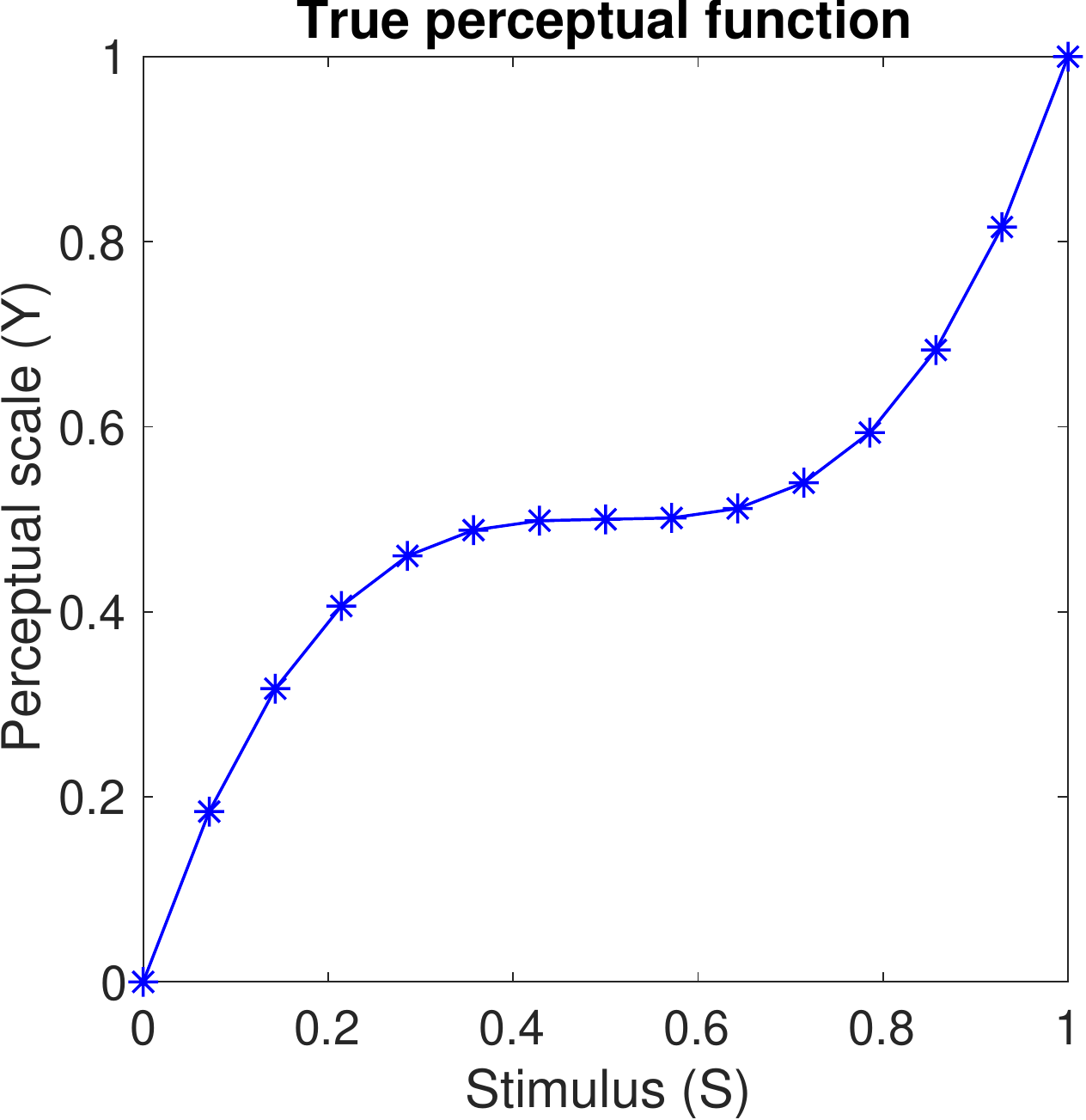}
\end{subfigure}\hspace{5mm}
\begin{subfigure}
    \centering
    \includegraphics[width = .30\linewidth]{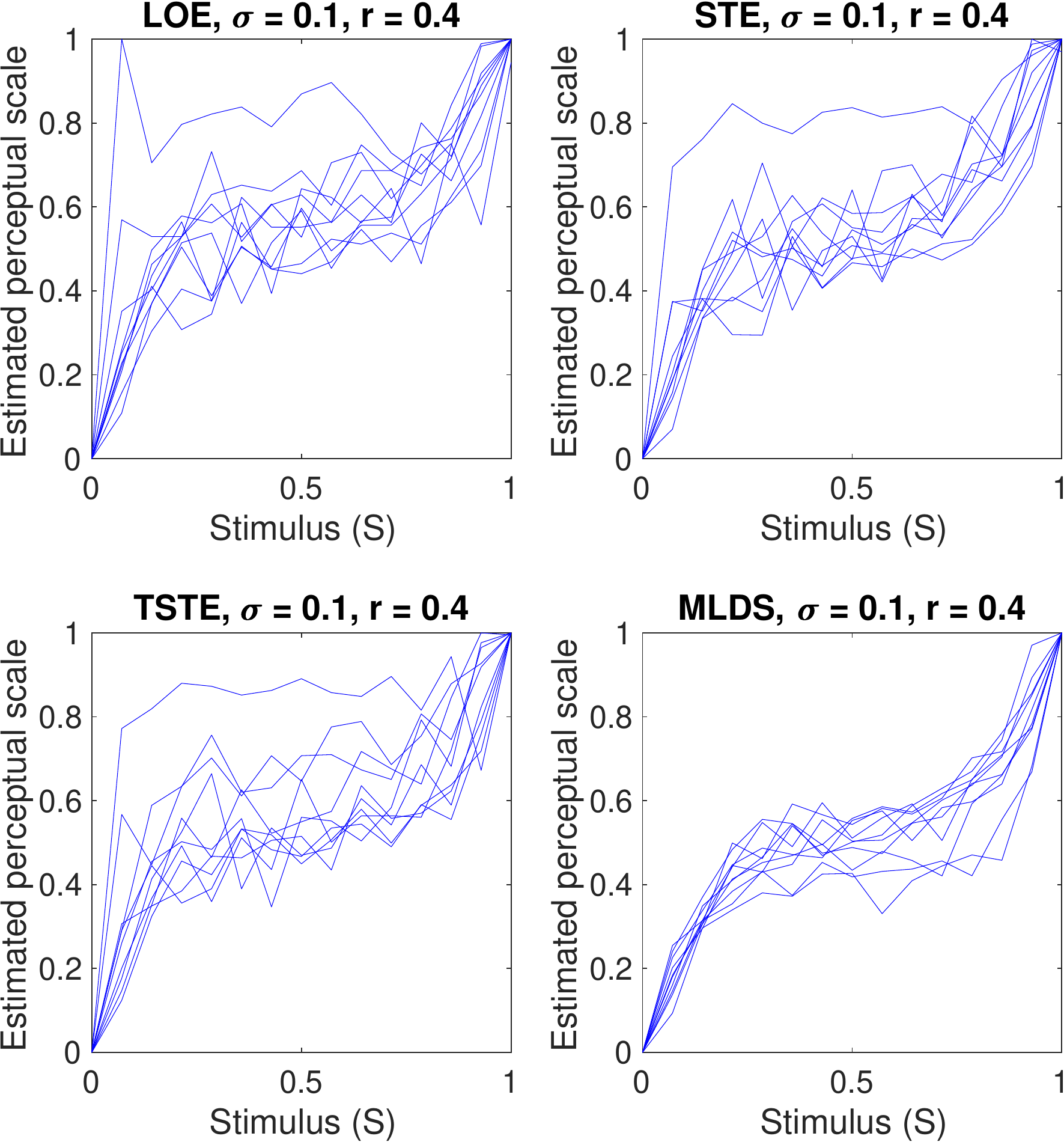}
\end{subfigure}
\begin{subfigure}
    \centering%
    \includegraphics[width = .80\linewidth]{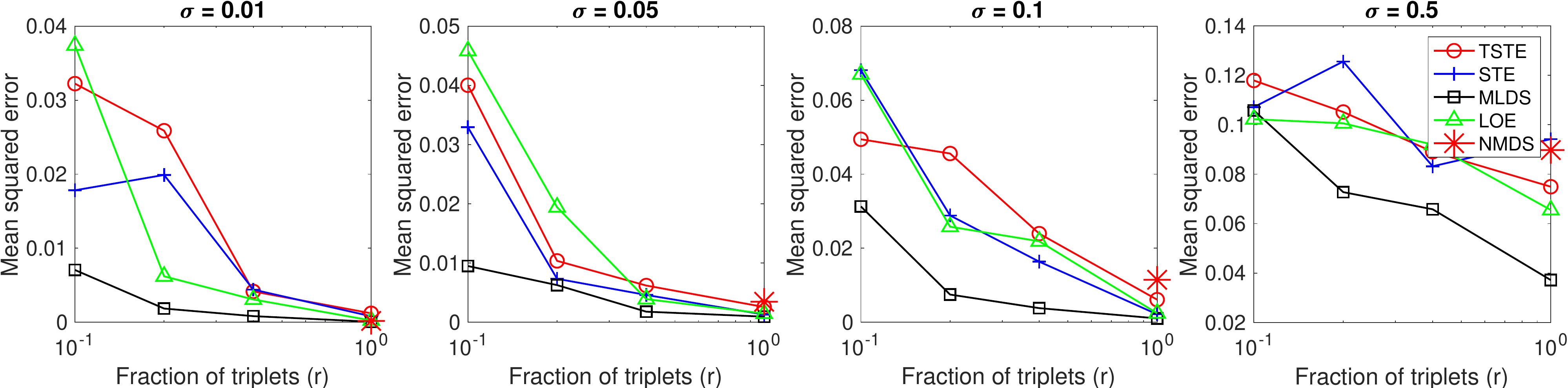}
\end{subfigure}
\begin{subfigure}
    \centering%
    \includegraphics[width = .80\linewidth]{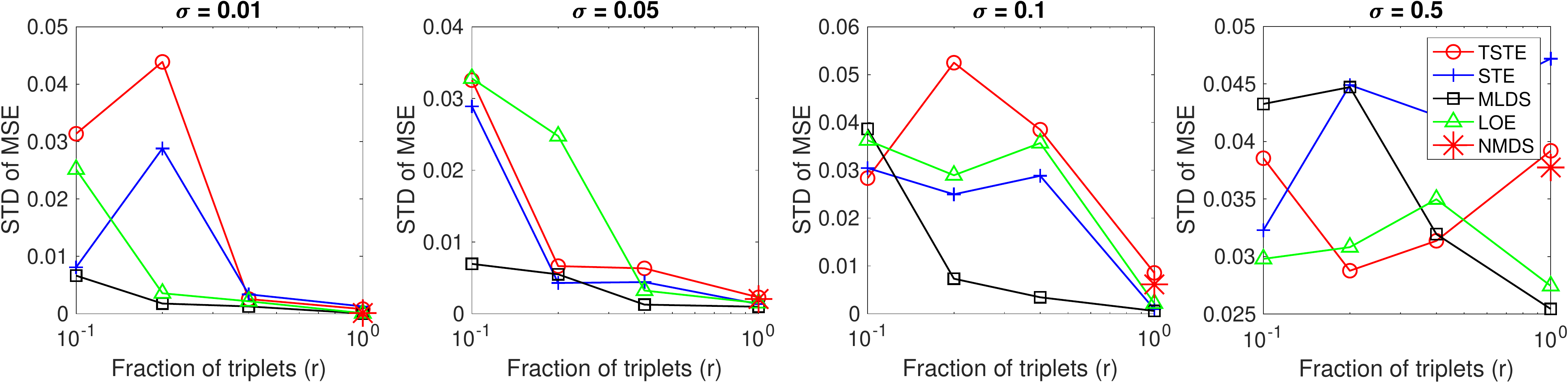}
\end{subfigure}
\begin{subfigure}
    \centering%
    \includegraphics[width = .80\linewidth]{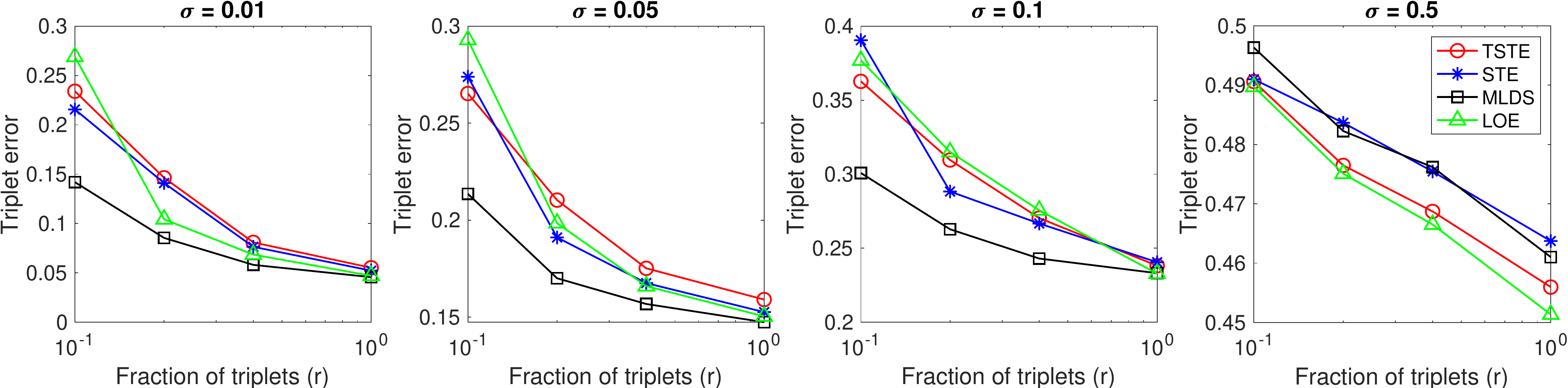}
\end{subfigure}
\begin{subfigure}
    \centering%
    \includegraphics[width = .80\linewidth]{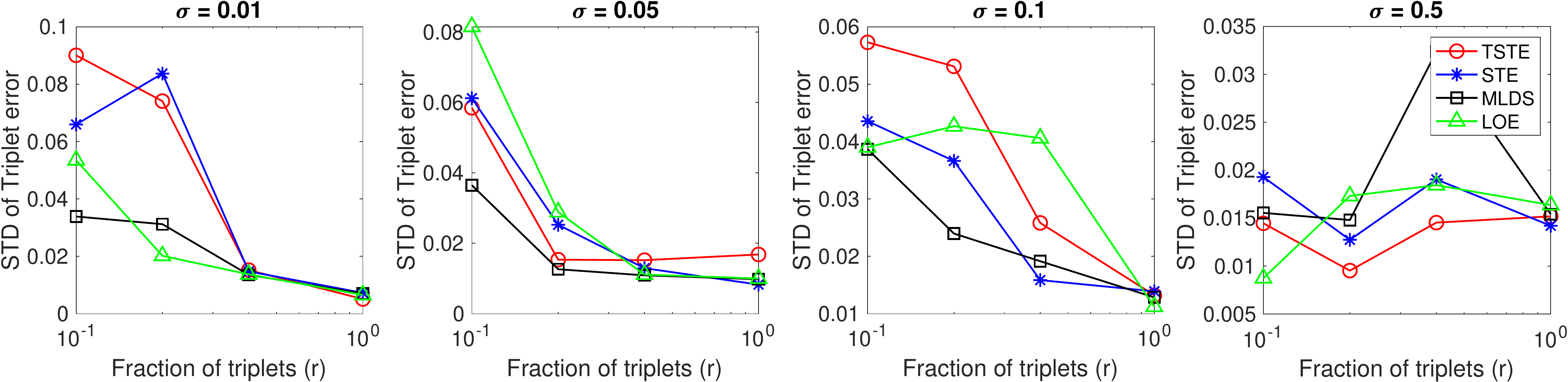}
\end{subfigure}
\caption{\label{fig:simsf1Extra} The comparison of various ordinal embedding methods (LOE, STE, t-STE) against the traditional methods in psychophysics (MLDS and NMDS) for a monotonic one-dimensional perceptual function. The true perceptual function ($y$) is appeared at the top left corner. Ten embedding results ($\hat y$) for a fixed value of noise standard deviation ($\sigma$) and triplet fraction ($r$) is shown on the top right corner. The second and third row depict the average MSE and the standard deviation of MSE for 10 runs of the algorithms. The fourth and fifth row show the average triplet error and the standard deviation of triplet error for 10 runs of the algorithms.}\end{figure}

\begin{figure}
\centering
\begin{subfigure}
    \centering
    \includegraphics[width = .32\linewidth]{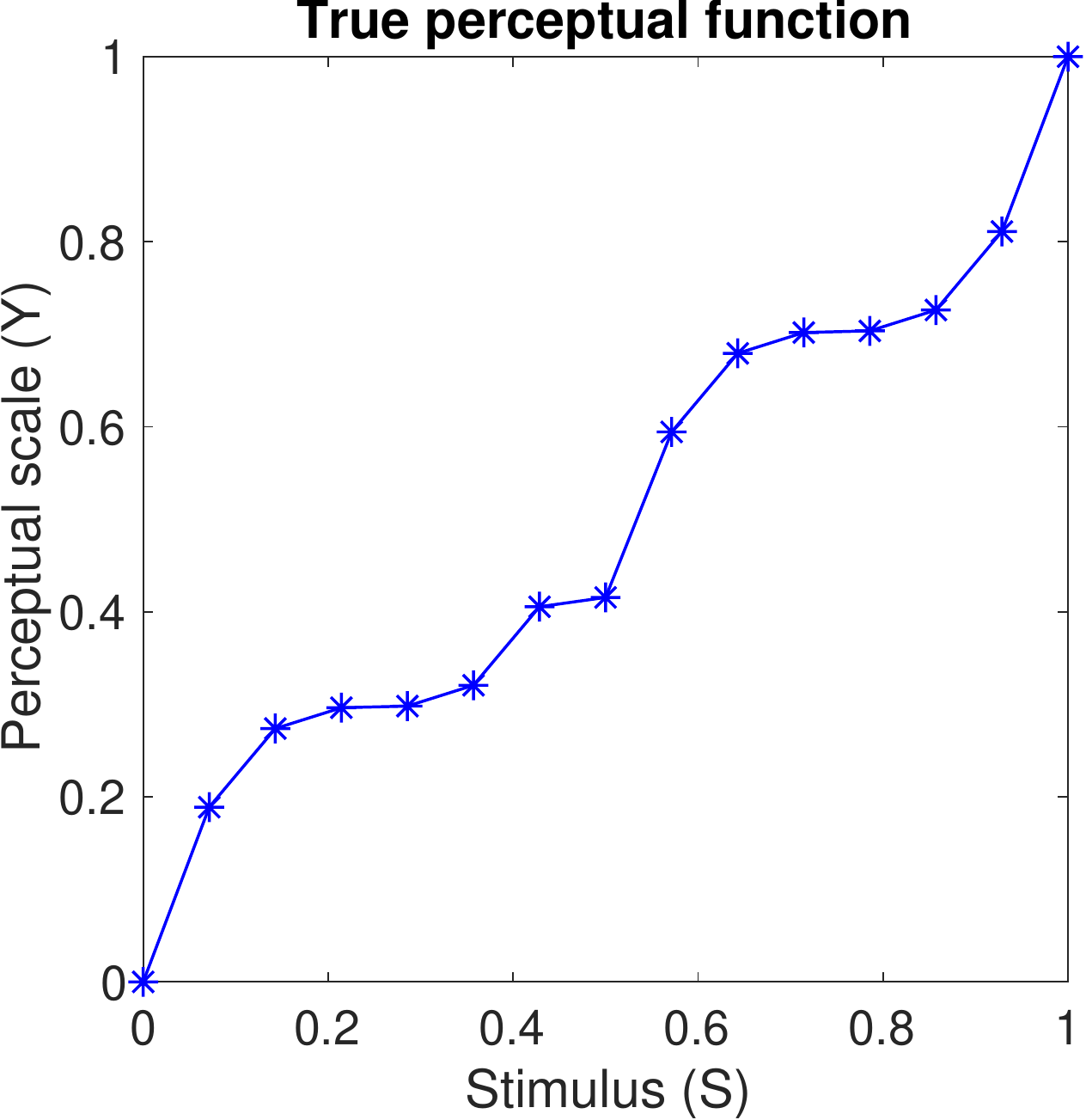}
\end{subfigure}\hspace{5mm}
\begin{subfigure}
    \centering
    \includegraphics[width = .30\linewidth]{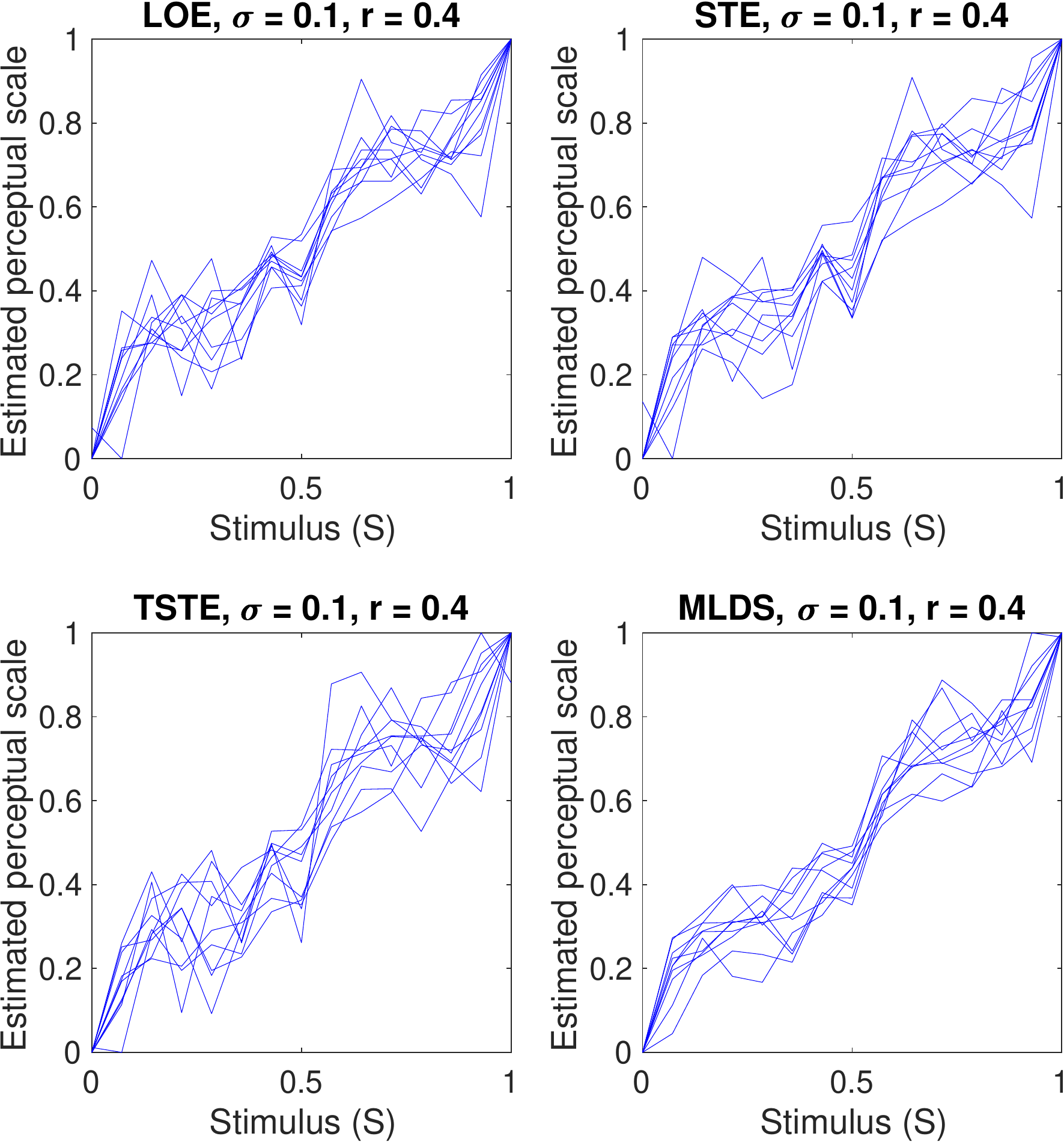}
\end{subfigure}
\begin{subfigure}
    \centering%
    \includegraphics[width = .80\linewidth]{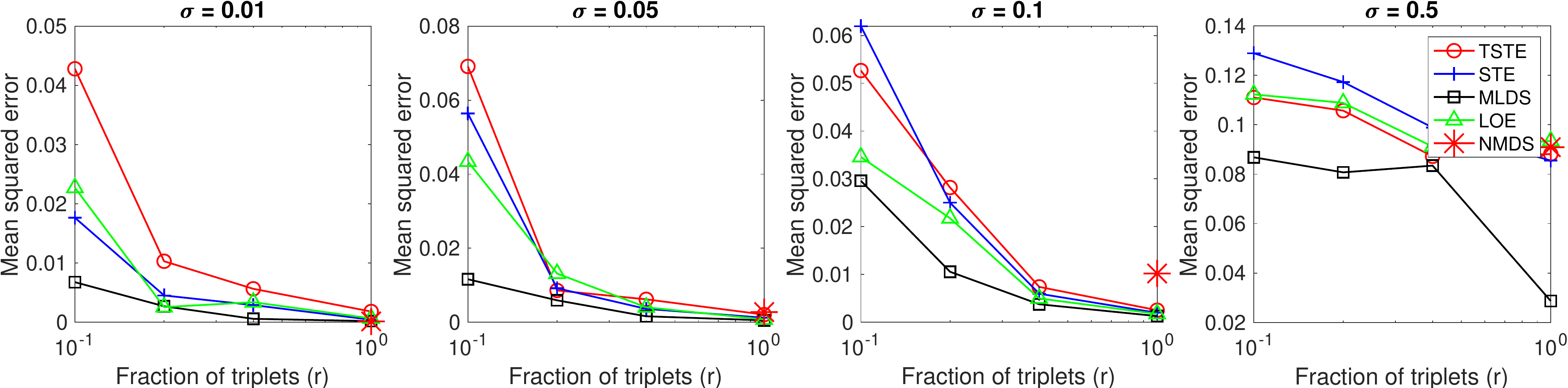}
\end{subfigure}
\begin{subfigure}
    \centering%
    \includegraphics[width = .80\linewidth]{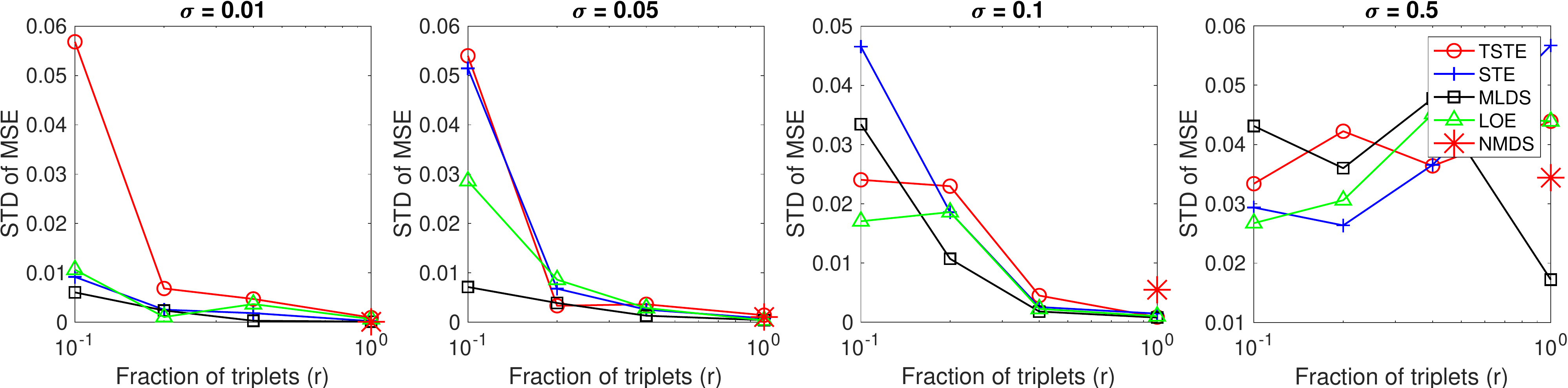}
\end{subfigure}
\begin{subfigure}
    \centering%
    \includegraphics[width = .80\linewidth]{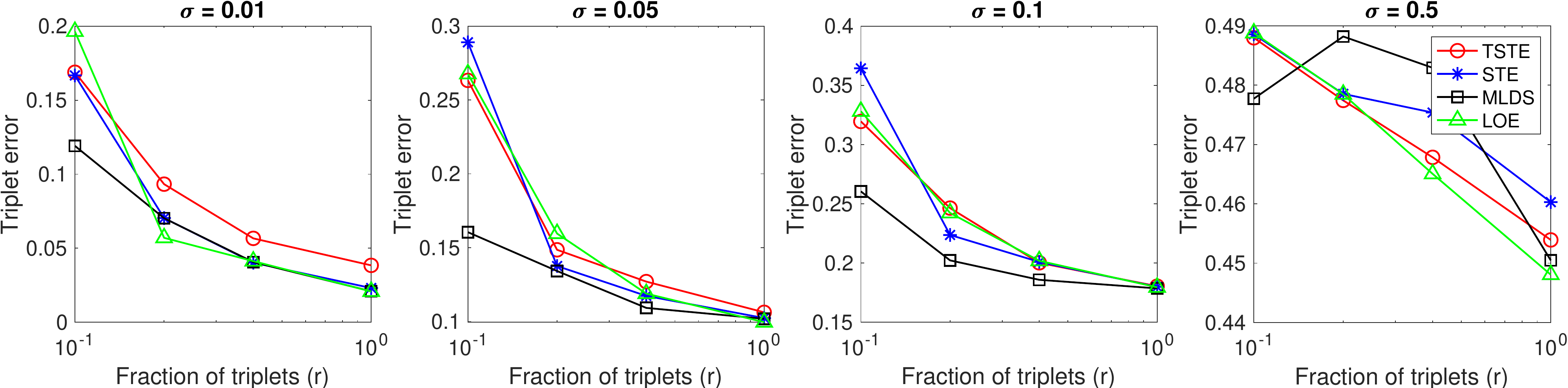}
\end{subfigure}
\begin{subfigure}
    \centering%
    \includegraphics[width = .80\linewidth]{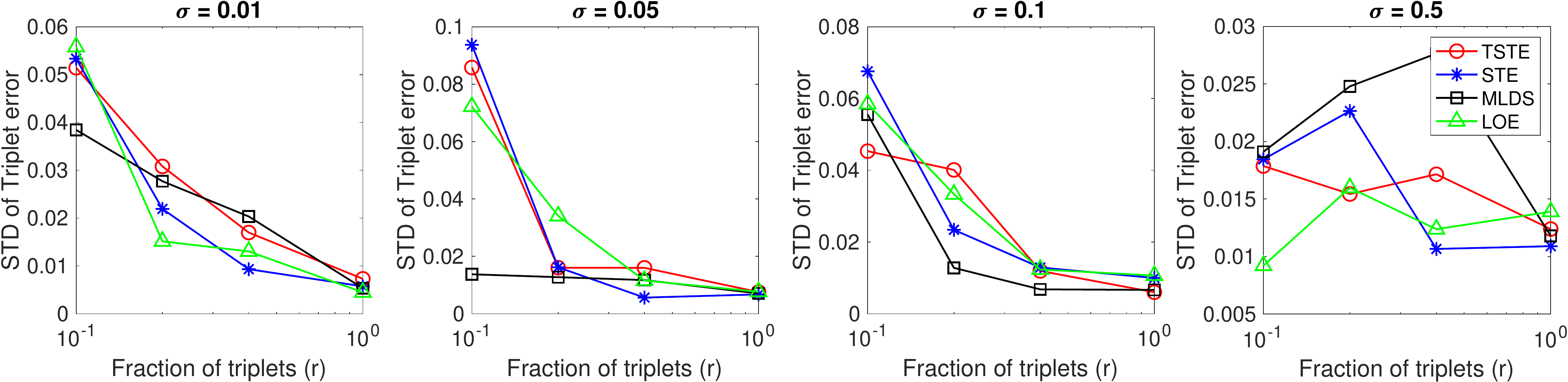}
\end{subfigure}
\caption{\label{fig:simsf2Extra} The comparison of various ordinal embedding methods (LOE, STE, t-STE) against the traditional methods in psychophysics (MLDS and NMDS) for a monotonic one-dimensional perceptual function. The true perceptual function ($y$) is appeared at the top left corner. Ten embedding results ($\hat y$) for a fixed value of noise standard deviation ($\sigma$) and triplet fraction ($r$) is shown on the top right corner. The second and third row depict the average MSE and the standard deviation of MSE for 10 runs of the algorithms. The fourth and fifth row show the average triplet error and the standard deviation of triplet error for 10 runs of the algorithms.}\end{figure}

\textbf{Monotonic scaling functions:} Here we present extensive results of the simulations with the monotonic scaling functions. Figure~\ref{fig:simsf1Extra} shows the extended results of the embeddings for a Sigmoid function. Besides the MSE and triplet error, we also show the standard deviation of these two criteria. In order to keep the plots neat and orderly, we reported the standard deviations in separate plots instead of using error-bars. We again see that MLDS works slightly better than other methods, as the scaling function meets the assumptions of MLDS. We also report the results for the same experiment with a different scaling function, see Figure~\ref{fig:simsf2Extra}. The function is a conditional degree 3 polynomial. Again, the MLDS outperforms all other embeddings.\\

\begin{figure}
\centering
\begin{subfigure}
    \centering
    \includegraphics[width = .32\linewidth]{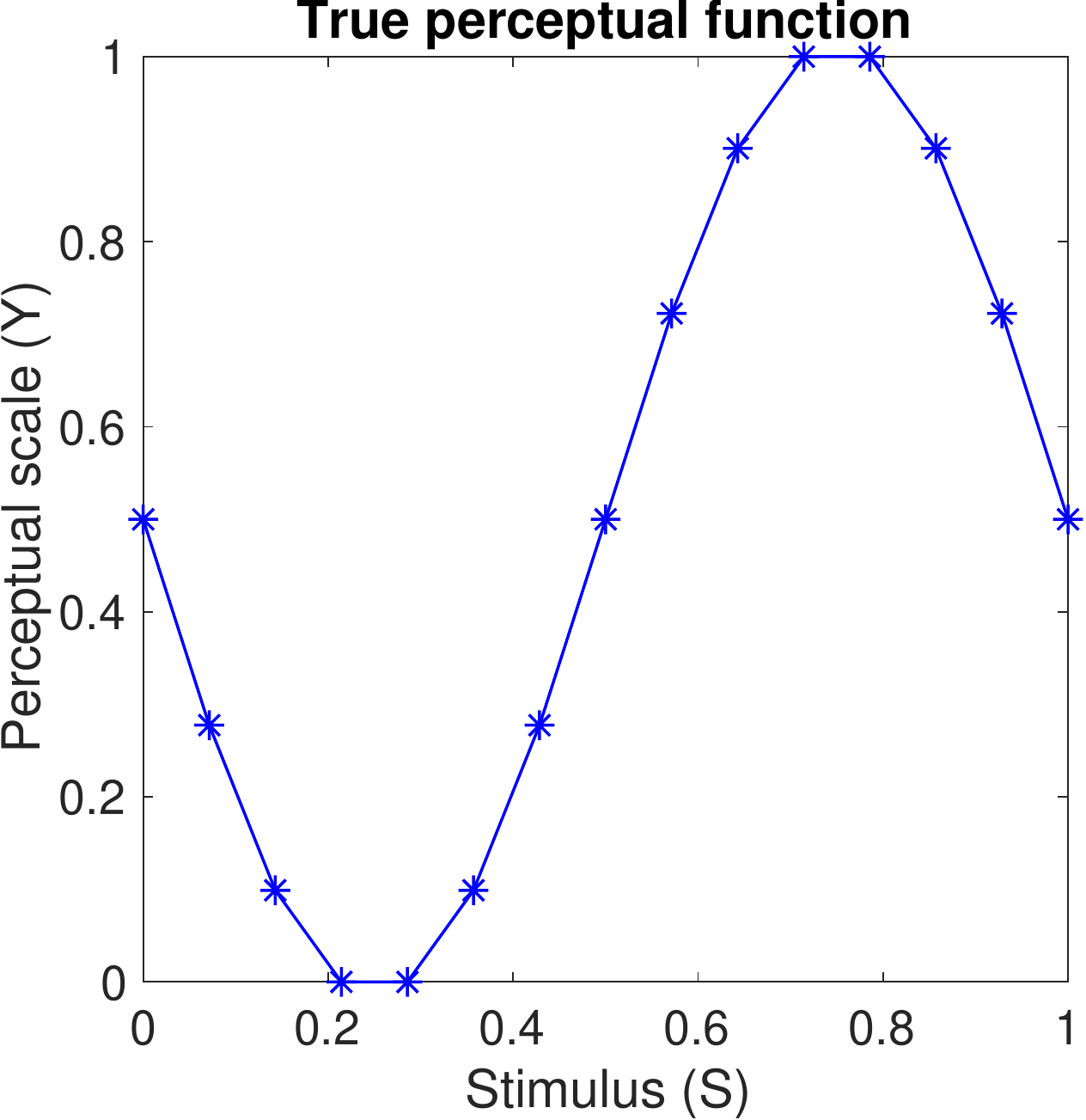}
\end{subfigure}\hspace{5mm}
\begin{subfigure}
    \centering
    \includegraphics[width = .30\linewidth]{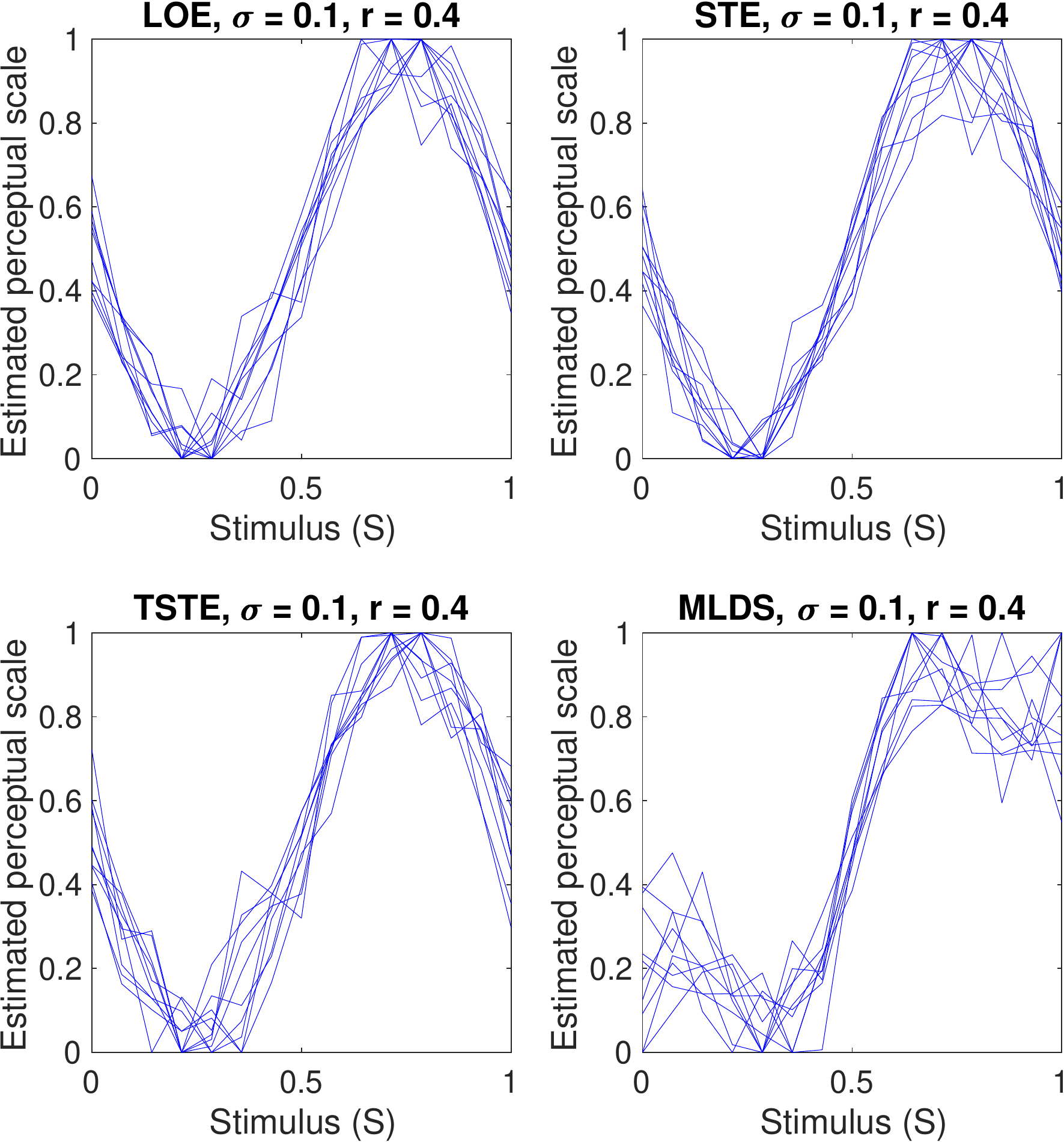}
\end{subfigure}
\begin{subfigure}
    \centering%
    \includegraphics[width = .80\linewidth]{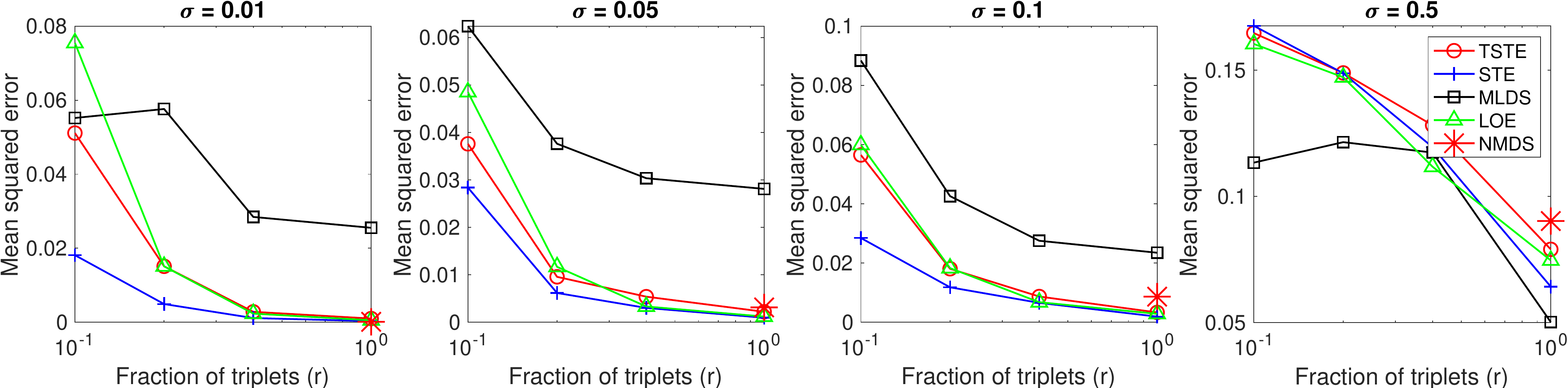}
\end{subfigure}
\begin{subfigure}
    \centering%
    \includegraphics[width = .80\linewidth]{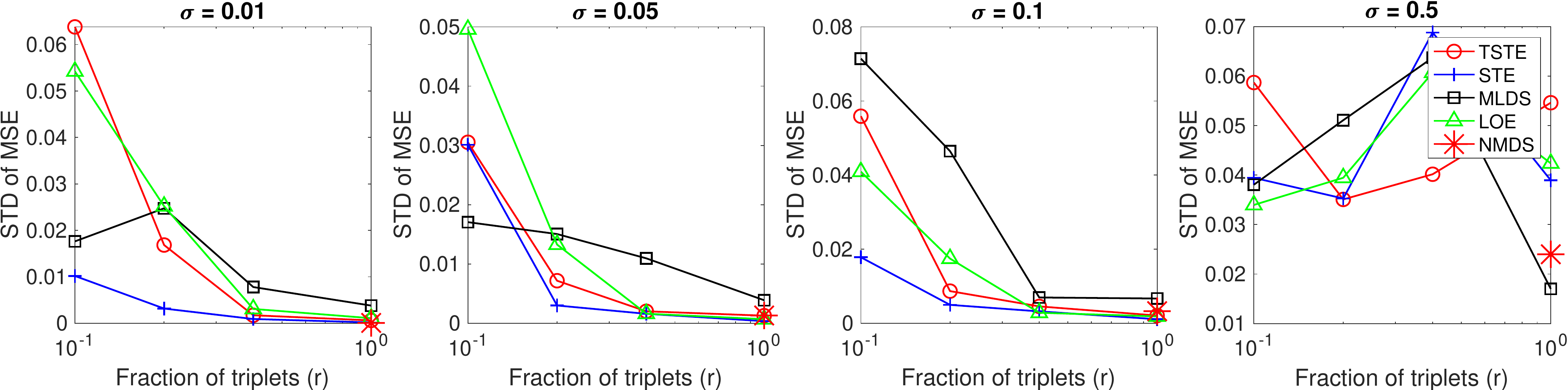}
\end{subfigure}
\begin{subfigure}
    \centering%
    \includegraphics[width = .80\linewidth]{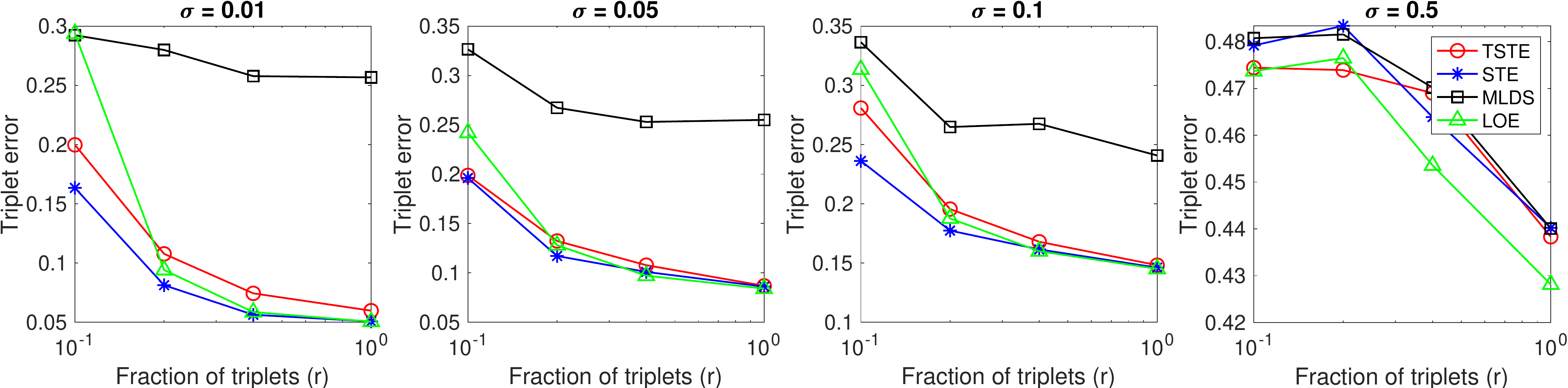}
\end{subfigure}
\begin{subfigure}
    \centering%
    \includegraphics[width = .80\linewidth]{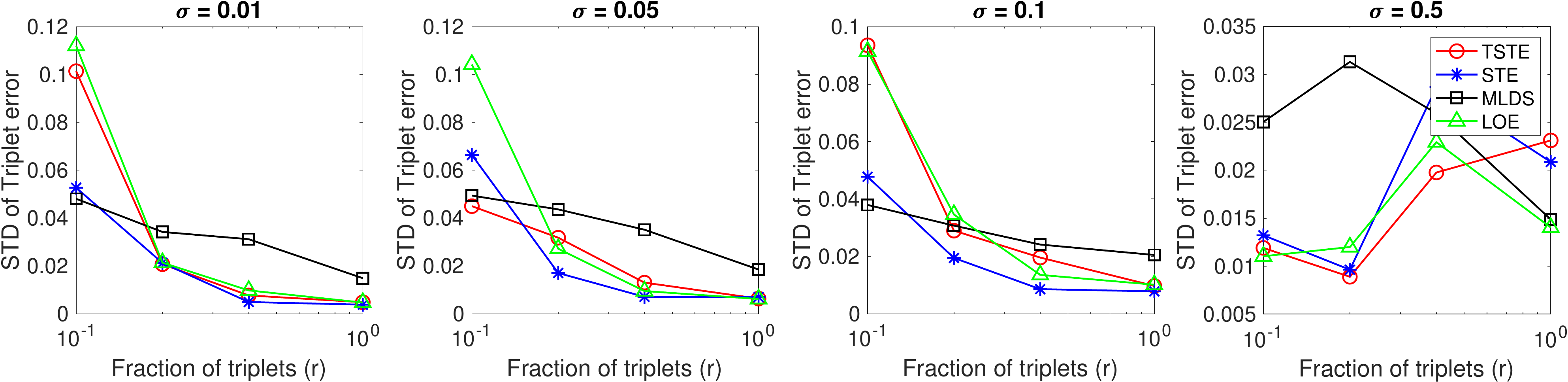}
\end{subfigure}
\caption{\label{fig:simsf3Extra} The comparison of various ordinal embedding methods (LOE, STE, t-STE) against the traditional methods in psychophysics (MLDS and NMDS) for a monotonic one-dimensional perceptual function. The true perceptual function ($y$) is appeared at the top left corner. Ten embedding results ($\hat y$) for a fixed value of noise standard deviation ($\sigma$) and triplet fraction ($r$) is shown on the top right corner. The second and third row depict the average MSE and the standard deviation of MSE for 10 runs of the algorithms. The fourth and fifth row show the average triplet error and the standard deviation of triplet error for 10 runs of the algorithms.}\end{figure}

\begin{figure}
\centering
\begin{subfigure}
    \centering
    \includegraphics[width = .32\linewidth]{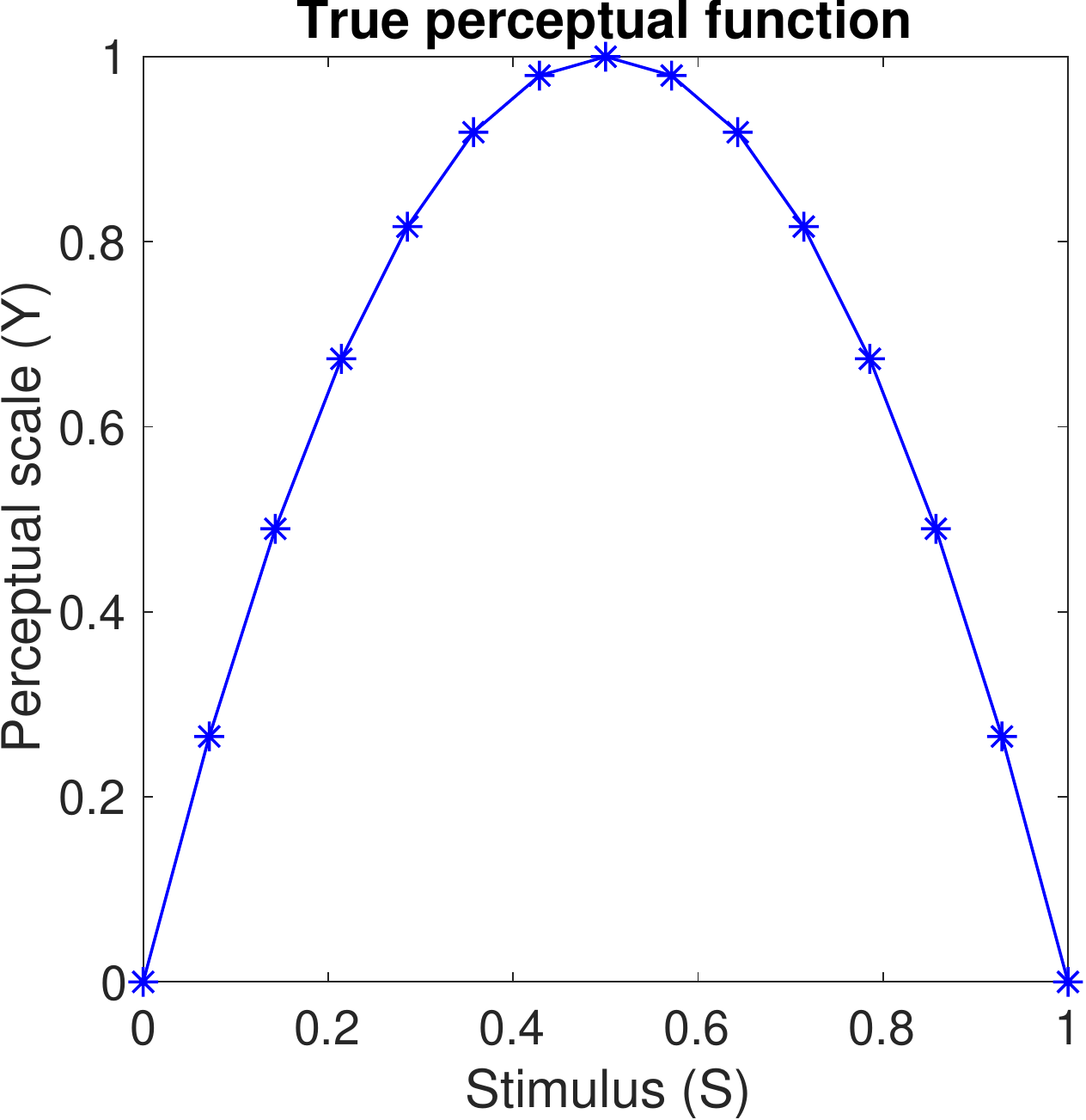}
\end{subfigure}\hspace{5mm}
\begin{subfigure}
    \centering
    \includegraphics[width = .30\linewidth]{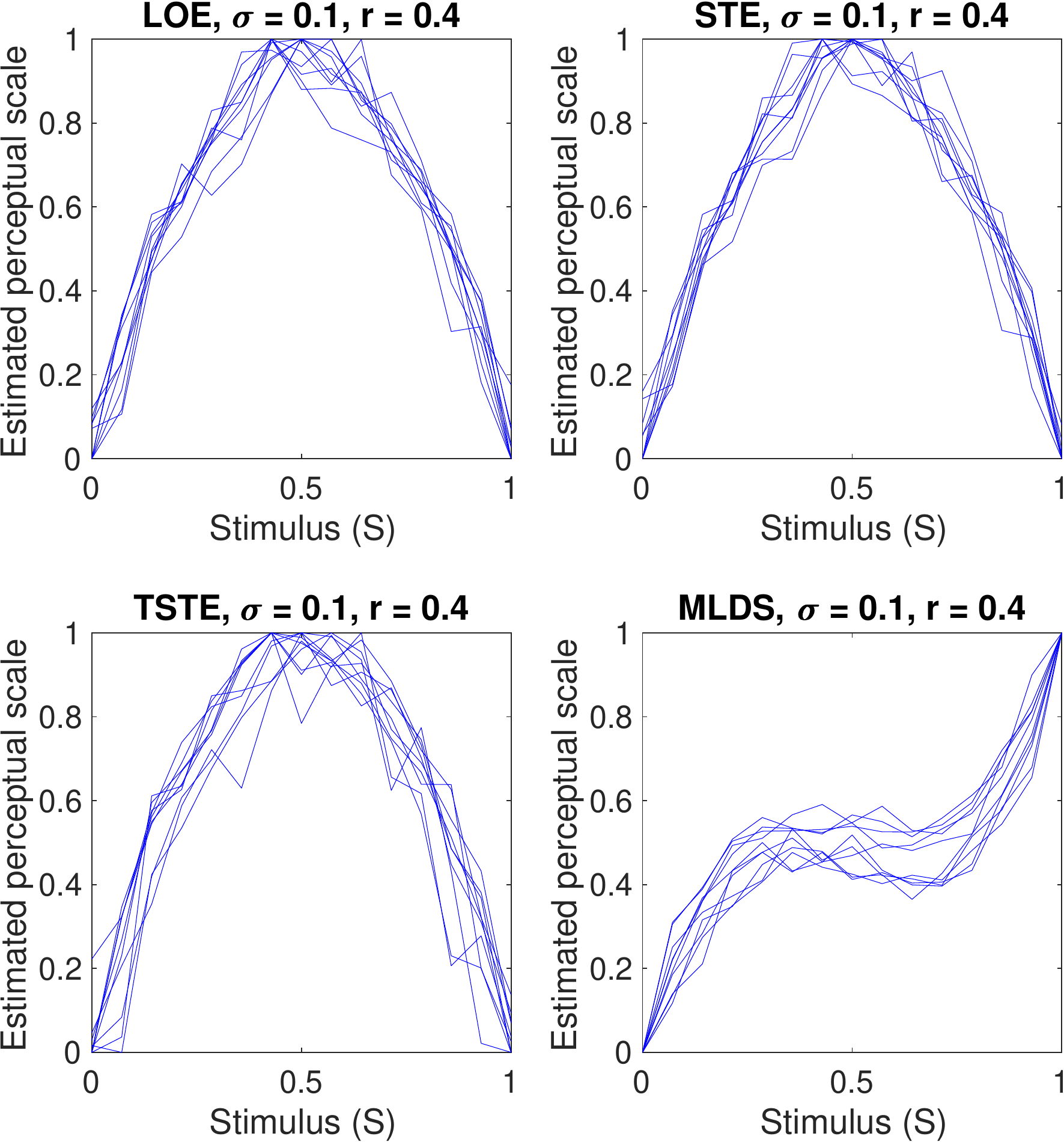}
\end{subfigure}
\begin{subfigure}
    \centering%
    \includegraphics[width = .80\linewidth]{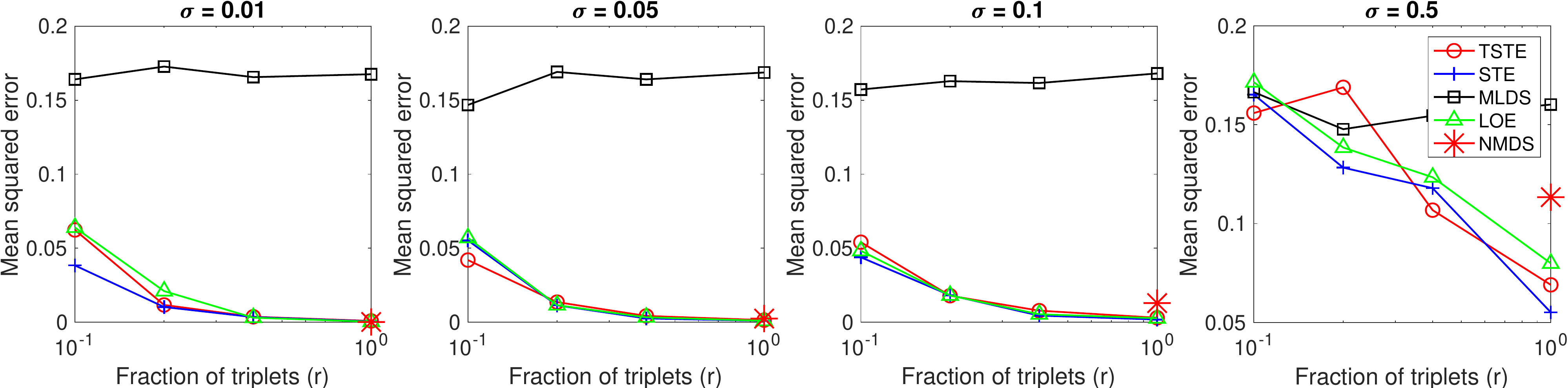}
\end{subfigure}
\begin{subfigure}
    \centering%
    \includegraphics[width = .80\linewidth]{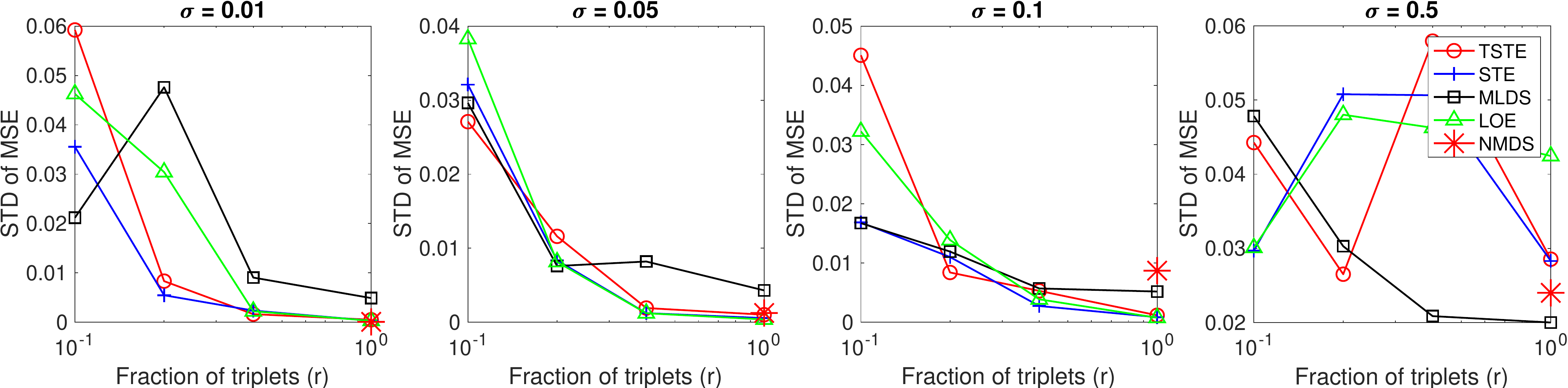}
\end{subfigure}
\begin{subfigure}
    \centering%
    \includegraphics[width = .80\linewidth]{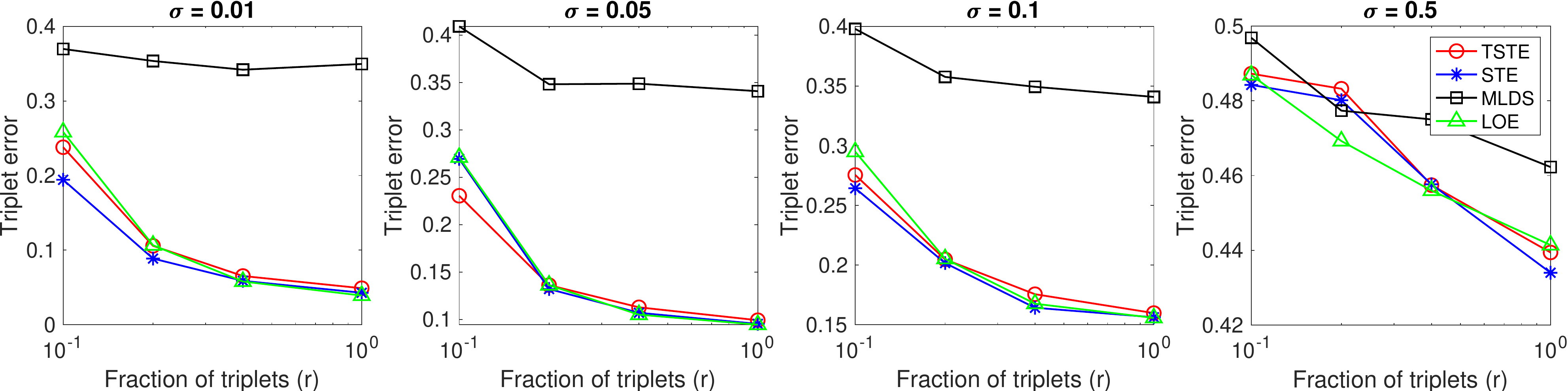}
\end{subfigure}
\begin{subfigure}
    \centering%
    \includegraphics[width = .80\linewidth]{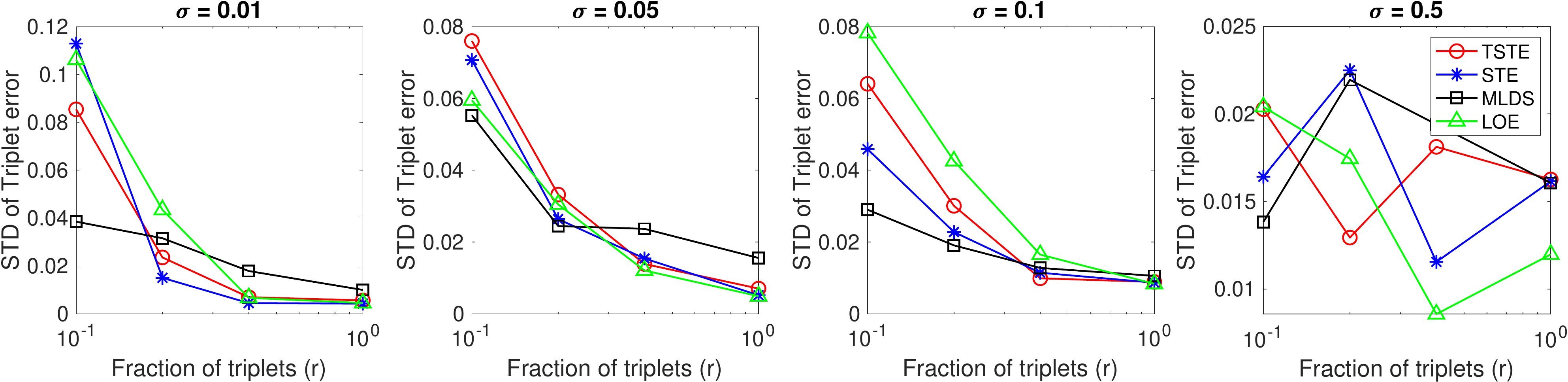}
\end{subfigure}
\caption{\label{fig:simsf4Extra} The comparison of various ordinal embedding methods (LOE, STE, t-STE) against the traditional methods in psychophysics (MLDS and NMDS) for a monotonic one-dimensional perceptual function. The true perceptual function ($y$) is appeared at the top left corner. Ten embedding results ($\hat y$) for a fixed value of noise standard deviation ($\sigma$) and triplet fraction ($r$) is shown on the top right corner. The second and third row depict the average MSE and the standard deviation of MSE for 10 runs of the algorithms. The fourth and fifth row show the average triplet error and the standard deviation of triplet error for 10 runs of the algorithms.}\end{figure}

\textbf{Non-monotonic scaling functions:} Similar to the monotonic functions, we compare the performance of the embedding methods on two non-monotonic function. Figure~\ref{fig:simsf4Extra} shows the extra results for a degree two polynomial scaling function. The MSE and triplet error both show poor performance of MLDS in comparison with the embedding methods. The second scaling function is a Sinusoid. The results of comparison are shown in Figure~\ref{fig:simsf3Extra}. The embedding methods work clearly better than the MLDS, however the difference is not as big as the previous function. The reason might be that the Sinusoid is closer to a monotonic function than the second degree polynomial.

\begin{figure}
\centering
\begin{subfigure}
    \centering
    \includegraphics[width = .32\linewidth]{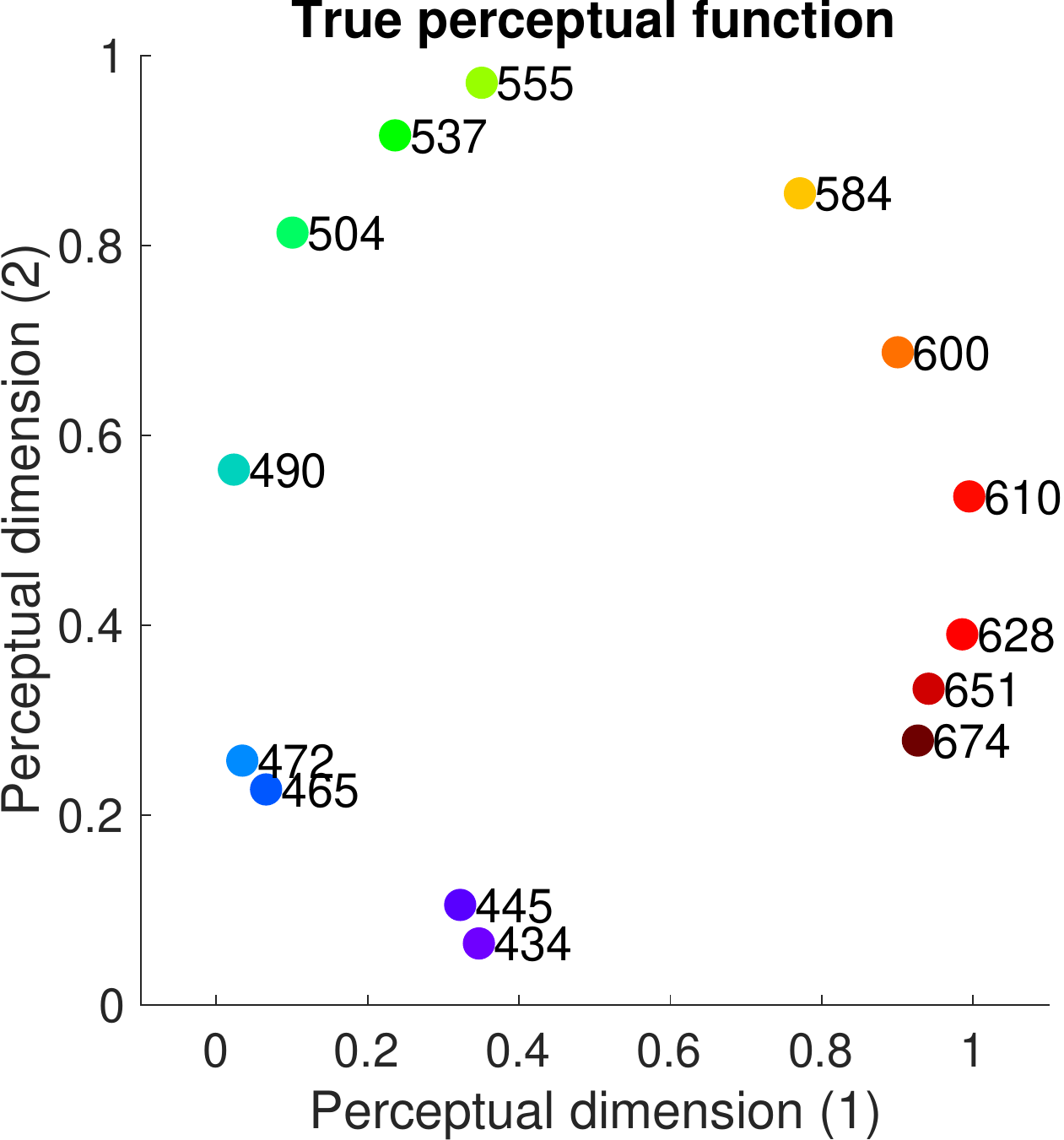}
\end{subfigure}\hspace{5mm}
\begin{subfigure}
    \centering
    \includegraphics[width = .30\linewidth]{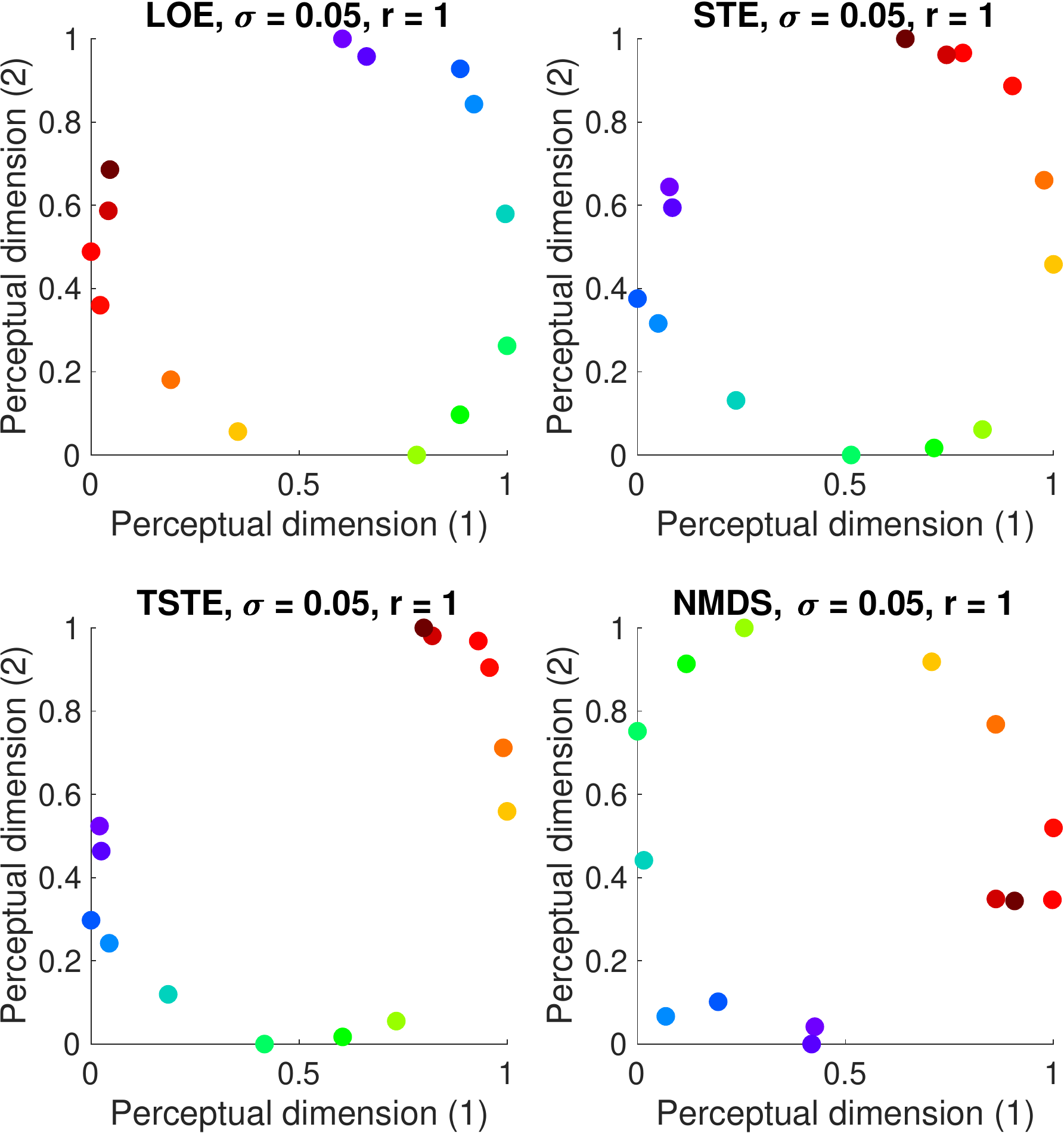}
\end{subfigure}

\begin{subfigure}
    \centering%
    \includegraphics[width = .80\linewidth]{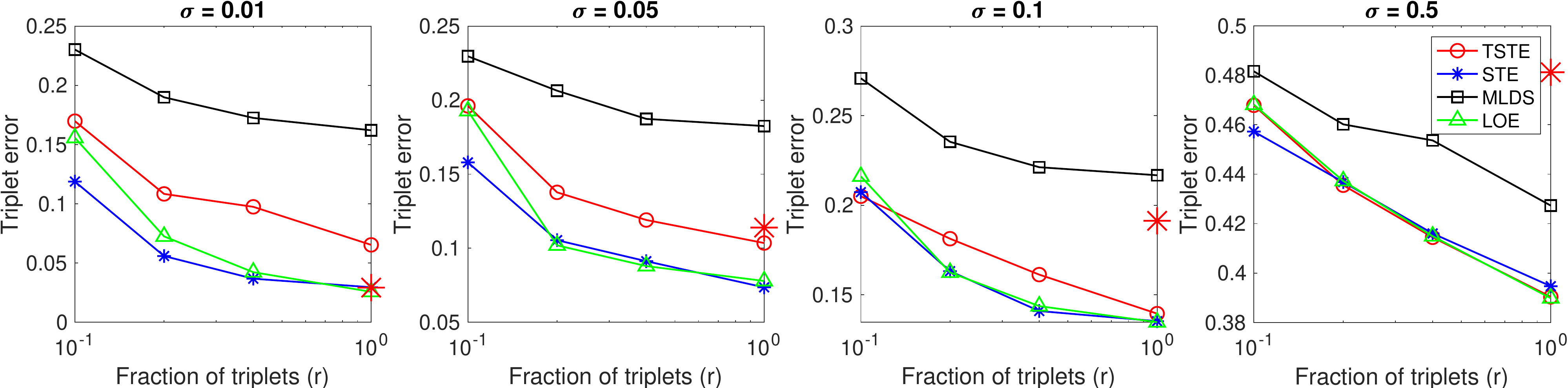}
\end{subfigure}

\caption{\label{fig:simsf5Extra} The comparison of various ordinal embedding methods (LOE, STE, t-STE) against the traditional methods in psychophysics (MLDS and NMDS) for a two-dimensional perceptual function. The true perceptual function ($y$) is appeared at the top left corner. one of embedding results ($\hat y$) for a fixed value of noise standard deviation ($\sigma$) and triplet fraction ($r$) is shown on the top right corner. The second row shows the average triplet error over 10 repetitions of the algorithms.}\end{figure}
\end{document}